\documentclass[11pt,letterpaper,english]{article}
\usepackage{microtype}
\usepackage{float}
\usepackage{amsmath} 
\usepackage{amsthm}
\usepackage{mathrsfs}
\usepackage{bm}
\usepackage{svg}
\usepackage{graphicx}
\usepackage{setspace}
\setdisplayskipstretch{0.9}
\usepackage{amssymb}
\usepackage{multirow}
\usepackage{rotating}
\usepackage{array}
\usepackage{booktabs}
\usepackage{algorithm}
\usepackage{algpseudocode}
\usepackage{color}
\usepackage{pifont}
\usepackage{threeparttable}
\usepackage{mdframed}   
\usepackage{tabularx}  
\usepackage[margin=1in]{geometry}  
\usepackage{mathtools}
\usepackage[small,compact]{titlesec}
\newtheorem{definition}{Definition}
\newcolumntype{C}[1]{>{\centering\arraybackslash}m{#1}}
\newmdenv[
  topline=true,
  bottomline=true,
  rightline=true,
  leftline=true,
  linewidth=1pt,
  roundcorner=5pt,
  skipabove=\baselineskip,
  skipbelow=\baselineskip
]{definitionbox}
\DeclareMathAlphabet{\mathcalligra}{T1}{calligra}{m}{n}
\usepackage[round,sort]{natbib}
\usepackage{times}

\usepackage{mathastext}
\usepackage{abstract}

\usepackage[hang,flushmargin]{footmisc}
\usepackage[hidelinks,hypertexnames=false]{hyperref}
\usepackage[section]{placeins}
\usepackage{footnote}
\usepackage[toc,page]{appendix}
\usepackage{subcaption}
\usepackage{blkarray}
\usepackage{stackengine}
\usepackage{accents}
\usepackage{enumitem}
\usepackage[mathscr]{euscript}
\newcommand{\argmax}{\arg\!\max}
\theoremstyle{definition}

\theoremstyle{definition}

\usepackage[utf8]{inputenc}
\usepackage{longtable}
\usepackage{epstopdf}
\usepackage{comment}
\usepackage{pgfplots}

\usepackage{caption}
\usepackage[normalem]{ulem}

\theoremstyle{plain}

\newcommand{\squishlist}{
   \begin{list}{$\bullet$}
    { \setlength{\itemsep}{0pt} \setlength{\parsep}{1pt}
      \setlength{\topsep}{1pt} \setlength{\partopsep}{1pt}
      \setlength{\leftmargin}{1.5em} \setlength{\labelwidth}{1em}
      \setlength{\labelsep}{0.5em} } }

\newcommand{\squishlisttwo}{
   \begin{list}{$\bullet$}
    { \setlength{\itemsep}{0pt} \setlength{\parsep}{0pt}
      \setlength{\topsep}{0pt} \setlength{\partopsep}{0pt}
      \setlength{\leftmargin}{1em} \setlength{\labelwidth}{1.5em}
      \setlength{\labelsep}{0.5em} } }

\newcommand{\squishend}{
    \end{list}  }

\def\mk#1{{\bf \color{brown}#1}}

\def\hy#1{{\bf \color{blue}#1}}
\begin{document}
\pagestyle{empty}

\title{Boundedly Rational Meta-Learning in Sequential Consumer Choice}

\author{
  Mehrzad Khosravi\thanks{We are grateful to the participants of the 2025 INFORMS Marketing Science Conference, the 2025 Conference on AI, ML, and Business Analytics, the 2026 Haring Symposium, the thirteenth triennial choice symposium 2026, and the University of Washington marketing seminar for helpful feedback and comments, which have significantly improved this paper. We also thank Cheng-Yu Hung for his thoughtful discussion notes at the 2026 Haring Symposium. Please address all correspondence to: \href{mailto:mehrzad@uw.edu}{mehrzad@uw.edu}, \href{mailto:maxkw@uw.edu}{maxkw@uw.edu}, and \href{mailto:hemay@uw.edu}{hemay@uw.edu}.} \\
  \textit{University of Washington}
  \and
  Max Kleiman-Weiner \\ \textit{University of Washington}
  \and
  Hema Yoganarasimhan \\ \textit{University of Washington}
}

\date{\today}

\date{}
\maketitle

\begin{abstract}
\begin{singlespace}

Many consumer decisions involve repeated choices under uncertainty, where experience in one context may inform decisions in another. For example, experience with a brand in one market or usage context may shape beliefs about that brand in a new context. We study whether such cross-context transfer takes the form of meta-learning, in which experience across contexts updates higher-order beliefs that guide learning in a new context. In a hierarchical laboratory task, participants choose among airlines across routes and observe noisy binary outcomes. Participants improve both within and across routes, indicating cross-route knowledge transfer. We compare human choices with no-transfer, fully integrated meta-learning, and boundedly rational meta dynamic programming policies, \(\mathrm{BRMDP}(D)\), where \(D\) is the number of hyper-posterior draws used to approximate integration. Trial-by-trial likelihood comparisons show that low-\(D\) policies, especially \(\mathrm{BRMDP}(1)\), best predict participant choices. The results suggest that consumers transfer information across contexts using coarse representations of higher-order uncertainty.

\end{singlespace}
\end{abstract}

\thispagestyle{empty}
\newpage

\section{Introduction}
\label{sec:introduction}

\subsection{Sequential Decision-Making and Meta-Learning}
\label{ssec:meta_learning_decision_making}

Many consumer decisions are repeated choices under uncertainty rather than isolated one-shot decisions. For example, households choose among brands or providers across purchase occasions, observe stochastic outcomes, and balance exploiting the current best option with learning through exploration. This exploration--exploitation trade-off is central in classic models of consumer learning and dynamic choice \citep{hutchinson2008consumer, ching2017empirical}. 
Yet in many markets, the relevant learning sequence does not restart when a consumer enters a new context. The same brand may reappear in a new category, channel, usage occasion, or location, and the consumer must decide how much prior experience should carry over. In these settings, experience in one context can become informative in another, especially when direct experience in the new context is still limited.

Consider the example of a traveler who chooses among several airlines for repeated trips on a route \citep{meyer1995sequential}. Over time, they learn which airline performs better on that route. When they travel to a new destination, however, they face a related decision problem: the same airlines are available, but their performance may differ on the new route. A natural question is whether they start from scratch---treating the new route as unrelated---or leverage past experience to form a better initial belief about each airline. Similar forms of knowledge-transfer arise when consumers evaluate brand extensions or related products. In such cases, experience with one product can shape beliefs about the brand and influence expectations in a new category \citep{wernerfelt1988umbrella}.

To formalize these ideas, we consider two forms of learning. The first is \emph{within-context learning}, in which consumers update beliefs about alternatives using feedback from the current route, product category, or usage context. The second is \emph{cross-context learning}, in which consumers use experience from prior contexts to inform beliefs and decisions in a new but related context. We focus on a particular form of cross-context learning---\emph{meta-learning}---in which experience across contexts is used to learn higher-order regularities that shape beliefs in a new context. In the airline example, within-context learning corresponds to discovering which airline performs well on the current route, while meta-learning corresponds to using experience from earlier routes to learn about the airline-level distribution that generates route-specific performance.

The central question is therefore not just whether consumers transfer information across contexts, but {\it how} they do so. One possibility is that each new context is treated as independent, with consumers relying solely on within-context learning. Another is that consumers act as fully integrated Bayesian meta-learners, carrying all prior uncertainty into the next context and using it in forward-looking planning. A third possibility lies between these extremes: consumers may transfer brand-level information, but only through coarse or simplified summaries of prior experience. This paper distinguishes among these possibilities.

These distinctions matter for measurement, theory, and managerial decision-making. Yet despite extensive work on consumer learning and sequential decision-making (see $\S$\ref{sec:literature_review}), we still know relatively little about how consumers transfer knowledge across contexts when feedback is noisy, experience is sparse, and horizons are short. In particular, we lack evidence on whether cross-context transfer resembles fully integrated Bayesian meta-learning or a cognitively simpler approximation.

\subsection{Research Agenda and Challenges}

We address these gaps by developing a framework that compares human behavior with algorithmic benchmarks in an environment that creates structured opportunities for meta-learning. Our research agenda is organized around three questions. First, do consumers transfer knowledge acquired in one context to another when learning is based on sparse, stochastic feedback? Second, if consumers do transfer knowledge, how closely does their behavior resemble an optimal meta-learning policy that fully exploits cross-context structure? Third, if behavior departs from this optimum, which boundedly rational model best captures the observed dynamics and the conditions under which transfer is stronger or weaker?

Answering these questions requires overcoming four challenges.
\squishlist
\item The first is {\it design}: we need an experimental environment that is sufficiently realistic to capture key features of consumer choice---repeated decisions, noisy outcomes, and brand regularities across contexts---while remaining controlled enough to isolate cross-context or meta learning. In particular, the environment must allow consumers to learn within a context and then encounter a new context in which the same alternatives reappear.  

\item The second is {\it diagnosis}: before estimating formal models, we need reduced-form evidence that separates improvement within a context from improvement across contexts. Such evidence can establish whether participants carry information forward to later contexts.

\item The third is {\it modeling}: meta-learning requires integrating higher-order uncertainty across contexts, which can make a fully integrated dynamic program computationally demanding even in small problems. Thus, a useful model class must include a no-transfer benchmark as well as meta-learning policies that vary in how finely they integrate prior uncertainty.

\item The fourth is {\it comparison}: we need a formal comparison framework that evaluates competing transfer mechanisms on the same decision histories and information sets, so that cross-context learning can be distinguished from within-context learning, task familiarity, and generic decision noise.
\squishend

\subsection{Our Approach and Findings}

To address these four challenges, we develop a four-part pipeline that mirrors this structure: a hierarchical laboratory environment that creates controlled opportunities for meta-learning; reduced-form behavioral diagnostics that document within-route learning and cross-route transfer; dynamic-programming-based policy classes that formalize no transfer, fully integrated transfer, and boundedly rational transfer; and a trial-by-trial likelihood framework that compares these mechanisms on the same decision histories.

The laboratory environment is an interactive, incentive-aligned airline choice task. Participants choose among three fictitious airlines over a sequence of routes, with multiple flights on each route. Outcomes are binary (\textit{on-time} vs.\ \textit{delayed}), as in a Bernoulli finite-horizon multi-armed bandit. The key design feature is hierarchical structure: each airline has a latent brand-level distribution over route-specific performance. Thus, routes are neither identical nor unrelated. Experience on one route can inform beliefs on later routes, allowing us to study within-route learning and cross-route knowledge transfer in a controlled setting, while the hierarchical structure creates scope for meta-learning.

The reduced-form evidence is consistent with both within-route as well as across-route meta-learning. Participants not only improve within routes as they accumulate direct experience, but they also improve across routes: they start with better airlines in later routes, choose the best-performing airline more often on later routes, and their pseudo-regret declines across routes. These patterns suggest that participants do not treat each new route as a fresh learning problem. They appear to carry forward information from earlier routes and use it to make better initial decisions in later ones.

These reduced-form patterns provide evidence of cross-context transfer, but they do not distinguish among the different explanations for the observed improvement. Similar aggregate patterns could arise because participants become more familiar with the task, because they transfer information across routes in a fully integrated Bayesian manner, or because they transfer information more coarsely using an approximate representation of prior experience. Distinguishing among these explanations requires a common framework that maps each candidate learning process into sequential choices.

We therefore develop three dynamic-programming-based policy classes. The baseline Dynamic Programming (DP) policy is Bayes-optimal \emph{within a route} under our Bernoulli--Beta model, but treats routes as independent and allows no cross-route knowledge transfer. The Meta Dynamic Programming (MetaDP) policy is the fully integrated Bayesian benchmark: it updates beliefs across routes and carries the resulting prior uncertainty into within-route dynamic planning. Finally, we introduce Boundedly Rational Meta Dynamic Programming policies, denoted \(\mathrm{BRMDP}(D)\), which approximate full meta-learning by representing the hyper-posterior with \(D\) draws. Small values of \(D\) correspond to coarse integration of prior uncertainty, while larger values approach the fully integrated MetaDP benchmark. This formulation is motivated by the resource-rationality perspective in cognitive science, which interprets apparent suboptimality as the efficient use of limited computational resources \citep{lieder2020resource,bhui2021resource}.

We then compare these policies with human behavior using a trial-by-trial likelihood framework. Following model comparison methods from sequential choice research \citep{daw2011trial}, we evaluate each policy on the same observed decision histories. We compute participant-level likelihoods for DP and MetaDP directly and develop a Monte Carlo likelihood estimator for \(\mathrm{BRMDP}(D)\), which integrates over the latent internal draws implied by the boundedly rational policy. This common likelihood framework allows us to isolate the incremental explanatory power of cross-route transfer and to ask whether transfer is best characterized as absent, fully integrated, or boundedly rational.

The likelihood results favor boundedly rational meta-learning. Low-\(D\) specifications, especially \(\mathrm{BRMDP}(1)\), fit the choice data better than both the no-transfer DP benchmark and the fully integrated MetaDP benchmark. Thus, the data reject both extremes. Participants do not behave as if they ignore cross-route information, but they also do not behave like fully integrated Bayesian meta-planners. Instead, they appear to transfer knowledge across routes using a coarse representation of prior uncertainty. As \(D\) increases and \(\mathrm{BRMDP}(D)\) approaches MetaDP, model fit deteriorates and the policy must rely on more decision noise to resemble human behavior. We also compare \(\mathrm{BRMDP}(1)\) with simpler knowledge-transfer policies that carry airline-specific information across routes without learning higher-order beliefs; \(\mathrm{BRMDP}(1)\) continues to fit choices better than these benchmarks. Robustness checks across experimental environments, alternative priors, shorter planning horizons, and softmax choice rules confirm the same qualitative conclusion.

\subsection{Contributions and Implications}
\label{ssec:findings_contribution}

The paper makes three main contributions. Substantively, it not only shows that consumers meta-learn, but also provides evidence on \emph{how} they do so. We study the cross-context information transfer process in a dynamic setting where consumers repeatedly face exploration--exploitation trade-offs, receive sparse and stochastic feedback, and must decide how much experience from one context should inform choices in the next. The main substantive finding is that meta-learning is present but approximate: among the policies we consider, consumer choices are better described by a policy that carries forward brand-level regularities across adjacent contexts, but does so through a coarse representation of prior uncertainty rather than full Bayesian integration.

Methodologically, to our knowledge, this is the first paper to introduce and empirically test a MetaDP model and a boundedly rational \(\mathrm{BRMDP}(D)\) model of consumer meta-learning in sequential choice. The MetaDP model provides a fully integrated Bayesian benchmark that combines cross-context belief updating with finite-horizon dynamic programming. The \(\mathrm{BRMDP}(D)\) model provides a tractable boundedly rational alternative that limits the integration of prior uncertainty through a small number of hyper-posterior draws. We also develop a likelihood-based comparison framework that evaluates DP, MetaDP, and \(\mathrm{BRMDP}(D)\) on the same observed histories, allowing us to test whether consumers behave like no-transfer learners, fully integrated meta-learners, or boundedly rational meta-learners.

Our results have implications for both research and practice. For researchers, the implication is that cross-context learning should not be treated as a binary modeling choice between no transfer and full Bayesian integration. Models that match aggregate learning patterns can still imply the wrong underlying mechanism if they ignore how consumers represent and use transferred uncertainty. Our framework provides a way to test whether cross-context transfer is absent and, when it is present, whether it is better described by full or approximate meta-learning. This matters for empirical work on consumer learning, brand spillovers, dynamic demand, and counterfactual policy analysis.

For managers, the results imply that early experiences can shape downstream demand through transferable brand-level beliefs. But managers should not assume that consumers either ignore prior contexts or integrate all prior uncertainty fully. When behavior is closer to \(\mathrm{BRMDP}(1)\), demand in a new context is better viewed as arising from a mixture of consumers with coarse transferred priors rather than from a single representative consumer with one integrated posterior. This matters for pricing, promotion, trial, and route-entry interventions: a no-transfer model can be too conservative after favorable early experience, while a fully integrated model can miss profitable targeted interventions by averaging away coarse heterogeneity in consumer priors.

The remainder of this paper is organized as follows. In~$\S$\ref{sec:literature_review}, we review related literature. In~$\S$\ref{sec:setting_and_experiment}, we define the hierarchical Bernoulli--Beta environment and describe the experimental task. In~$\S$\ref{sec:behavioral_evidence_meta_learning}, we present reduced-form evidence of meta-learning through best-airline selection and pseudo-regret. In~$\S$\ref{sec:dp_models}, we introduce the DP, MetaDP, and BRMDP policy classes. In~$\S$\ref{sec:empirical_likelihood_comparisons}, we present the likelihood-based comparison framework and empirical model comparisons. In~$\S$\ref{sec:managerial_implications}, we discuss managerial implications and stylized counterfactuals, and conclude in $\S$\ref{sec:conclusion}.

\section{Related Literature}
\label{sec:literature_review}

Our research sits at the intersection of several related literatures. Within marketing and economics, it relates most directly to three streams of work: cross-category or cross-context belief transfer, structural models of sequential consumer learning under uncertainty, and boundedly rational dynamic choice. More broadly, it also connects to two strands of cognitive science research: hierarchical Bayesian learning and boundedly rational or resource-rational accounts of decision-making. Below, we discuss each of these literatures and how our work contributes to them. 

Marketing research has long recognized that experience with a brand in one category can shape beliefs and choices in another. In the umbrella-branding and spillover literature, experience with a brand in one category affects perceived quality, perceived risk, preferences, and choice in a second category \citep{wernerfelt1988umbrella, montgomery1992risk,mccoy2022two}. A series of empirical papers studies cross-category spillovers and establishes the presence of cross-category learning  \citep{erdem1998empirical, sridhar2012investigating, coscelli2004empirical}. The vast majority of these papers consider static settings and do not account for sequential decision-making or forward-looking behavior, except for \citep{che2015consumer}. This literature generally assumes that consumers are Bayesian learners and behave in a fully optimal way when transferring information from one context to another. It does not test \emph{how} consumers carry information from one context to another in sequential decision problems with sparse and noisy feedback. Conceptually, this literature establishes that information can transfer across categories or contexts. Our contribution is to examine whether experience across contexts is used to learn higher-order structure governing context-specific performance---a form of transfer we refer to as \emph{meta-learning}---and, if so, how consumers integrate the resulting uncertainty. We model and empirically distinguish among no cross-context transfer, fully integrated meta-learning, and boundedly rational approximations to meta-learning.


A second relevant stream studies sequential consumer learning under uncertainty within a single context. Beginning with \citet{erdem1996decision}, this literature models brand choice as Bayesian learning in dynamic discrete-choice settings, often allowing consumers to be forward-looking and to recognize the information value of experimentation; see \citet{ching2017empirical} for a review. These models correspond to the no-transfer DP benchmark in our framework: they capture within-context learning over time, but treat each new context as starting from the same prior. Relative to this work, our framework adds cross-context belief transfer. We ask whether information accumulated in one context shapes the priors consumers bring to the next and, if so, whether that transfer is fully integrated or approximate.


A third stream asks whether consumers actually solve such sequential problems in the fully optimal manner assumed by standard structural models discussed above.Work in economics, marketing, and decision research has long noted that dynamic optimization can be descriptively demanding, even in relatively simple environments \citep{hutchinson1994dynamic, meyer1995sequential} and \citet{meyer2016we} show that assuming agents behave as if they solve Bellman equations can produce brittle counterfactuals and misleading managerial implications when the underlying behavioral process is misspecified. A growing stream of research proposes and tests boundedly rational approximations to dynamic choice, including biased beliefs and inattention, non-Bayesian belief updating, limited-horizon planning, index-based rules, directed cognition models, and myopic value-of-information heuristics \citep{grubb2015cellular, jindal2021importance, gabaix2006costly, yang2015bounded, lin2015learning, liu2020understanding,  tehrani2024heuristic}.  While this literature also considers bounded-rational approaches, it has largely focused on approximating the DP problem {\it within} a single context. The main distinction in our paper is that the boundedness we study lies in the transfer process across contexts---that is, in how consumers approximate the integration of higher-order priors across routes or categories---rather than in how they approximate DP within a given context.\footnote{We note that approximations of DP proposed in this literature can also be embedded within our formulation; see $\S$\ref{ssec:robustness_and_extensions} for details.}

More broadly, our work builds on two strands of cognitive science research. First, modeling learning as inference over hierarchical Bayesian models \citep{griffiths2024bayesian}. This line of research investigates how humans update beliefs across multiple knowledge categories to learn from sparse and ambiguous data. \cite{kemp2007learning} show that humans (and even young children) form and update {\it overhypotheses}, i.e., priors over priors that enable learning across categories. For instance, if a parent points to the family pet's face and says ``dog'', the child must consider multiple consistent hypotheses. Is a ``dog'' the whole animal? Does it refer to just the face? Or is the name of that specific animal ``dog'' but other similar animals will have different names? Children disambiguate these possibilities by learning higher-order priors (i.e., overhypotheses) such as, ``new words are more likely to refer to the whole object and not just a part of the object'' so that ``dog'' most likely refers to the whole animal \citep{xu2007word}. These hierarchical priors enable humans to learn new concepts from just a few examples \citep{tenenbaum2011grow}. They have also been useful for building human-like computer vision systems that can quickly learn new visual concepts \citep{salakhutdinov2012learning}.

Cognitive scientists have also used these models to explain how people leverage hierarchical priors in spatial decision-making. \cite{schulz2020finding} conducted laboratory experiments on a spatial multi-armed bandit where each arm was tied to a letter on a single row of a keyboard. They show that subjects can quickly exploit the spatial structure in the task and learn linear patterns, e.g., expected rewards increase/decrease from `a' to `;'. These linear patterns helped subjects more efficiently balance exploration and exploitation. Once subjects have learned that the rewards were aligned linearly across the keys (arms), they could quickly leverage this knowledge by just checking the relative values of the endpoints (`a' and `;') to see whether rewards increase or decrease along the row of the keyboard and then exploit from there. This provides experimental support for the idea that humans employ a form of hierarchical learning in decision-making under uncertainty. Here, our goal is to build off these basic insights from cognitive science to systematically test formal models of consumer meta-learning and decision-making that leverage hierarchical priors over brands and products. 

The second related branch of cognitive science to our paper is the study of humans as ``{\it bounded-rational}'' decision-makers. This literature aims to provide a theory that explains how humans make rational decisions in some conditions while being biased in others. In this view, bounded-rational analysis models judgments as the optimal use of limited time, memory, and attention, deriving simple algorithms that trade accuracy for cost, thereby reconciling near-optimal behavior with classic biases \citep{lieder2020resource}. Evidence comes from two complementary strands: within-person “wisdom of the crowd,” where averaging a person’s two independent estimates improves accuracy, revealing probabilistic internal representations; and sampling-based decisions, where people appear to act from only a few samples of a posterior \citep{vul2008measuring, vul2014one}. Formal analyses show that when samples and deliberation time are costly, using one or a few samples can maximize long-run reward, reproducing probability matching, speed–accuracy trade-offs, and Hick-law-like scaling. Complementary information-theoretic accounts treat cognitive effort as an information-capacity constraint, linking efficient coding and rational inattention to softmax (stochastic) choice, perseveration, and context-dependent value representations; under these limits, “noise” and reference dependence emerge as rational adaptations \citep{bhui2021resource}. Our study contributes to this literature by providing a framework for testing bounded rationality in human knowledge transfer across a sequential decision-making context.

Taken together, our contribution is not simply to add another dynamic choice model. Rather, we: (1) present a methodological framework that integrates cross-context spillovers, sequential exploration-exploitation, and bounded rationality in a single framework, and (2) we compare no-transfer, fully integrated, and boundedly rational transfer models on common information sets and trial-by-trial choice data from a real experiment. This lets us speak directly to whether consumers meta-learn across contexts and whether that meta-learning is fully optimal or only approximate.

\section{Model Environment and Experimental Task}
\label{sec:setting_and_experiment}

We now define the decision environment we study and describe how we implement it in the laboratory. Because our focus is meta-learning, we need an environment that can support both learning within a context and knowledge transfer across related contexts, with higher-order structure linking those contexts. To that end, we build on the airline choice problem in the seminal work by \citet{meyer1995sequential}, which considers a setting in which a consumer makes a sequence of choices about which airline brand to fly with on a single route. This setting has two key advantages. First, it has been shown to work well in studies with real participants because it is simple enough for subjects to understand and complete repeatedly. Second, the setup is quite flexible. While their work only considers single-route settings, we can easily extend the environment to multi-route scenarios.

\subsection{Single-Route Decision Problem}
\label{ssec:single_route_problem}

We begin with the consumer's problem on a single route. Consider a consultant based in Seattle who has recently secured a contract with a client in Dallas, requiring repeated trips on the Seattle--Dallas route. There are \(K\) airlines serving this route, and for each trip the consultant must choose one airline. As a busy professional, they care about whether their flights are on time.\footnote{For expositional ease, we treat on-time arrivals as the primary quality metric. However, our framework is agnostic about the quality metric used; one could instead use comfort or a broader utility measure that combines multiple service dimensions.} At the outset, they have no direct experience with the airlines on this route, but they can learn from realized outcomes as they accumulate flights.

\begin{figure}[htp!]
    \centering
    \includegraphics[width=1\textwidth]{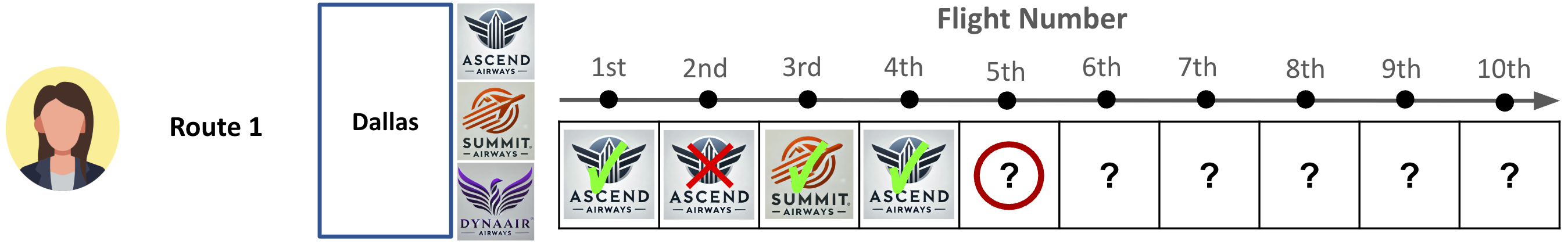}
    \caption{An example of a consultant in a single-route environment, with some experience with Ascend and Summit Airways before the fifth flight. For the fifth flight, they must decide whether to exploit their current information or explore a less familiar airline.}
    \label{fig:simple_example_outcome_new}
\end{figure}

We consider a setting with three fictitious airlines\footnote{We use fictitious airlines to avoid confounding the experiment with participants' preexisting beliefs, brand familiarity, loyalty, or idiosyncratic past experiences with real carriers. Using fictional names helps standardize the initial information set across participants and ensures that learning in the task reflects experience generated within the experiment rather than outside knowledge.}---Ascend, Summit, and DynaAir---so that \(K=3\), and assume that the consultant takes \(T=10\) flights on a given route. Figure~\ref{fig:simple_example_outcome_new} illustrates the sequential decision-making process in an environment in which the consultant initially has no experience with any airline. On each subsequent flight, they face the standard exploration--exploitation trade-off: they can \textit{exploit} their current knowledge by choosing the airline that has performed best so far, or \textit{explore} by selecting a less familiar airline in the hope of discovering a better-performing option.

In this single-route setting, the consultant's objective is to choose a policy \(\pi\) that maximizes the expected number of on-time flights over the finite horizon:
\begin{definition}
\label{def:single_route_objective_new}
We define the consumer's objective in a single-route setting as
\begin{equation}
    \max_{\pi} \mathbb{E} \left [\sum_{t=1}^T \gamma^{t-1} Y_t \,\bigg|\, \pi \right]
    \qquad \text{s.t.} \qquad Y_t \in \{0,1\},
    \label{eq:objective_function_new}
\end{equation}
where \(Y_t \in \{0,1\}\) is the outcome of flight \(t\) (\(1 \equiv\) on-time) and \(\gamma \in (0,1]\) is the discount factor.
\end{definition}

\subsection{Multi-Route Decision Problem}
\label{ssec:multi_route_problem}

\begin{figure}[htp!]
    \centering
    \includegraphics[width=1\textwidth]{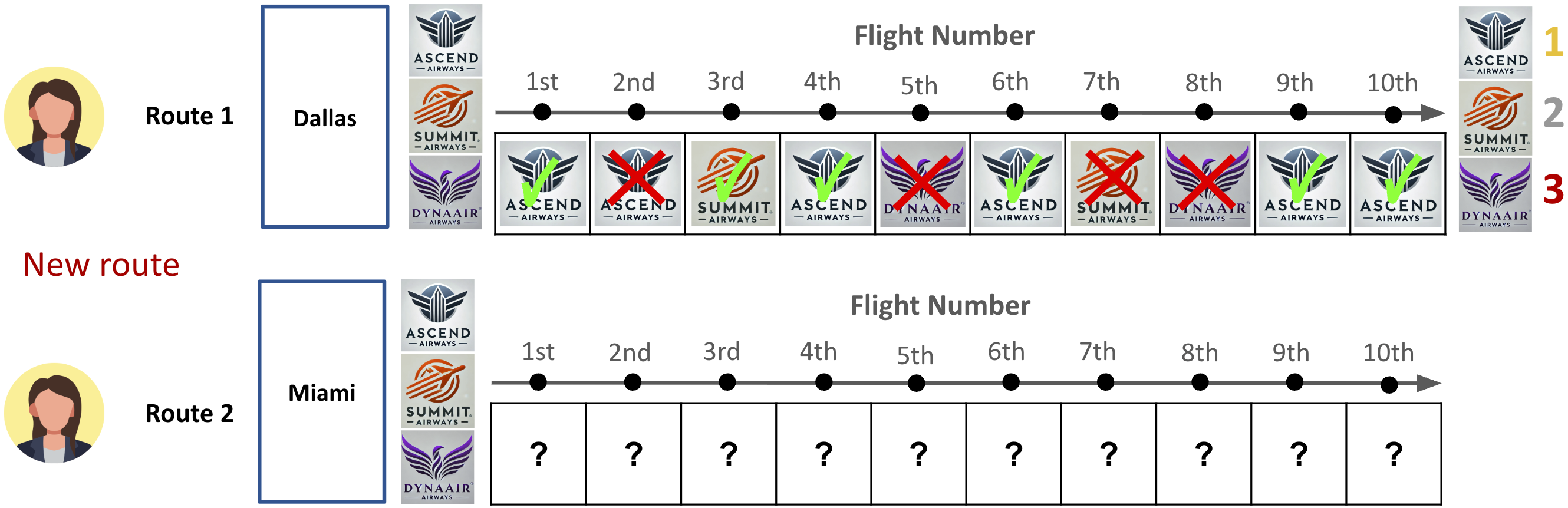}
    \caption{An example of a multi-route environment. The consultant learns about airline performance on the Seattle--Dallas route and may use that knowledge to inform decisions on the Seattle--Miami route. In this example, they conclude that Ascend performed best on the first route but remain uncertain about airline performance on the second.}
    \label{fig:meta_learning_timeline_new}
\end{figure}

We now extend the problem to multiple routes. After finishing one contract, the consultant may sign another one in a different city and therefore face a new route.\footnote{This is a common situation for many professionals, especially contract-based workers such as lawyers, construction contractors, salespeople, or actors.} The same airlines are available, but their performances may differ across routes because of variation in scheduling, crew allocation, operational hubs, gate access, or route-specific conditions. Figure~\ref{fig:meta_learning_timeline_new} illustrates this extension to a multi-route problem.

If there are \(M\) routes and the consultant makes \(T\) flights on each route, then the full horizon contains \(M \times T\) decisions. As in the single-route setting, outcomes are binary. The key difference is that when the consultant begins route \(m\), they have already accumulated information from earlier routes, and this information can shape their beliefs about airline performance on the current route.

Let $\mathcal{H}_m = \bigl((A_{m,1}, y_{m,1}), \ldots, (A_{m,T}, y_{m,T})\bigr)$ be the observed history for route $m$,
We can then define the start-of-route history object as \(\mathcal H_{1:m-1} = (\mathcal{H}_1,\ldots,\mathcal{H}_{m-1})\) for the concatenated history of the previous routes. Then the problem on route \(m\) is the same as in the single-route setting, except that it is conditioned on \(\mathcal H_{1:m-1}\). Thus, the consultant's objective is to choose policy $\pi_m$ on route $m$ as follows.
\begin{definition}
\label{def:multi_route_objective_new}
We define the consultant's objective on route \(m\) in the multi-route setting as
\begin{equation}
    \max_{\pi_m} \mathbb{E} \left [\sum_{t=1}^T \gamma^{t-1} Y_{m,t} \,\bigg|\, \mathcal H_{1:m-1}, \pi_m \right]
    \qquad \text{s.t.} \qquad Y_{m,t} \in \{0,1\},
    \label{eq:multi_route_objective_new}
\end{equation}
where \(Y_{m,t}\) is the outcome of flight \(t\) on route \(m\).
\end{definition}

This formulation creates scope for meta-learning.  Experience on earlier routes may not only improve decisions within those routes, but also shape the beliefs the consultant brings to a new route and thereby improve their choices on later routes. 

\subsection{Hierarchical Environment}
\label{ssec:environment_new}

We now formalize the environment that generates outcomes. In the single-route case, each airline would have one unknown on-time rate for that route. In the multi-route case, however, each airline-route pair has its own route-specific on-time rate, and these route-specific rates are themselves related through a higher-level airline-specific distribution.

We model the true on-time rate of airline \(k\) on route \(m\), denoted by \(\theta^{\ast}_{m,k}\), as a draw from an airline-specific Beta distribution. These airline-level Beta distributions capture the idea that some airlines tend to be better than others across routes, even though route-specific performance still varies stochastically. The consumer does not observe the underlying parameters of these airline-level distributions. They only observe the realized outcomes of the flights they take.

\begin{figure}[htp!]
    \centering
    \includegraphics[width=.6\textwidth]{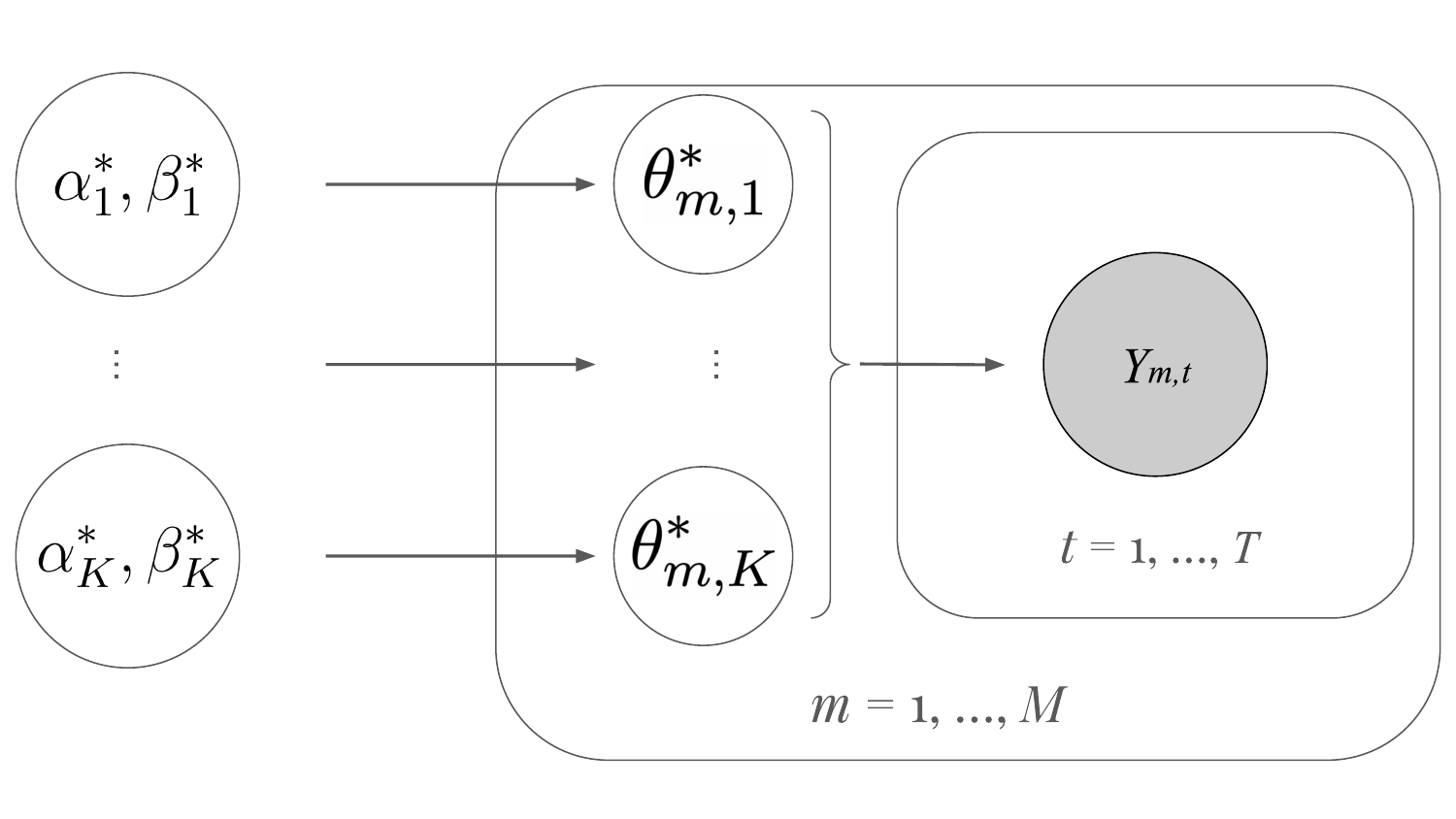}
    \caption{Graphical representation of the hierarchical Bayesian environment. Airline \(k\)'s on-time rate on route \(m\), denoted \(\theta^{\ast}_{m,k}\), is drawn from an airline-specific Beta\((\alpha_k^{\ast},\beta_k^{\ast})\) distribution. These Beta distributions and route-level on-time rates are properties of the environment and are not observable to the consumer. The only observable component is the realized outcome of each chosen flight.}
    \label{fig:bayesian_hier_new}
\end{figure}

Figure~\ref{fig:bayesian_hier_new} provides a graphical representation of this hierarchical structure. Formally, we define the environment through the following hierarchical generative process \citep{koller2009probabilistic}:
\squishlist
    \item Stage 1: For each route \(m\), draw the true on-time rate of airline \(k\) from an airline-specific Beta distribution:
    \begin{equation}
        \theta^{\ast}_{m,k} \sim \text{Beta}(\alpha^{\ast}_k,\beta^{\ast}_k),
        \qquad \forall \; k \in \{1,\ldots,K\}, \quad \forall \; m \in \{1,\ldots,M\},
        \label{eq:mean_draw_new}
    \end{equation}
    where \(M\) is the number of routes.
    \item Stage 2: Let \(A_{m,t}\) denote the airline chosen on route \(m\), flight \(t\). Conditional on the chosen airline, draw the realized outcome from a Bernoulli distribution with mean \(\theta^{\ast}_{m,k}\):
    \begin{equation}
        Y_{m,t} \,\big|\, A_{m,t}=k \sim \mathrm{Bernoulli}(\theta^{\ast}_{m,k}),
        \qquad Y_{m,t}\in\{0,1\}.
        \label{eq:reward_draw_new}
    \end{equation}
\squishend

Throughout the paper, we use uppercase \(Y_t\) and \(Y_{m,t}\) to denote outcome random variables, and lowercase \(y_t\) and \(y_{m,t}\) to denote their realizations.

This hierarchical structure is central to the paper. It ensures that routes are neither identical nor completely unrelated. Because route-specific performances are drawn from stable airline-level distributions, experience on one route can inform future routes. That is exactly what creates scope for meta-learning, where a participant can use experience from earlier routes to infer the airline-level distributions governing route-specific performance. We list all the notation we used so far in Table~\ref{tab:notation_problem_environment} for convenience.

\begin{table}[htp!]
\centering
\small
\caption{Notation for the problem definition, model environment, and experimental task.}
\label{tab:notation_problem_environment}
\resizebox{\textwidth}{!}{%
\begin{tabular}{p{.20\textwidth} p{.80\textwidth}}
\toprule
\textbf{Notation} & \textbf{Definition} \\
\midrule
$k \in \{1,\ldots,K\}$ & Airline index \\
$m \in \{1,\ldots,M\}$ & Route index \\
$t \in \{1,\ldots,T\}$ & Flight index within a route \\
$K$ & Number of airlines \\
$M$ & Number of routes \\
$T$ & Number of flights per route \\
$\gamma$ & Discount factor applied to future payoffs within a route \\
$\theta_{m,k}^{\ast}$ & True on-time rate of airline $k$ on route $m$ \\
$(\alpha_k^{\ast},\beta_k^{\ast})$ & Environment's true Beta parameters for airline $k$ across routes \\
$A_t$ & Airline chosen at flight $t$ within a route \\
$A_{m,t}$ & Airline chosen at flight $t$ on route $m$ \\
$Y_t$ & Outcome random variable at flight $t$ in the single-route setting \\
$Y_{m,t}$ & Outcome random variable at flight $t$ on route $m$ \\
$y_t$ & Realized outcome at flight $t$ in the single-route setting \\
$y_{m,t}$ & Realized outcome at flight $t$ on route $m$ \\
$\pi,\pi_m$ & Decision policy in the single-route problem (\(\pi\)) and on route \(m\) in the multi-route problem (\(\pi_m\)) \\
$\mathcal{H}_{1:m-1}$ & History available at the beginning of route \(m\) consisting of past actions and realized outcomes from routes \(1,\ldots,m-1\) \\
\bottomrule
\end{tabular}
}
\end{table}

\subsection{Experimental Task}
\label{ssec:experimental_task}

\begin{figure}[htp!]
    \centering
    \begin{subfigure}[t]{.45\textwidth}
        \centering
        \includegraphics[width=\textwidth]{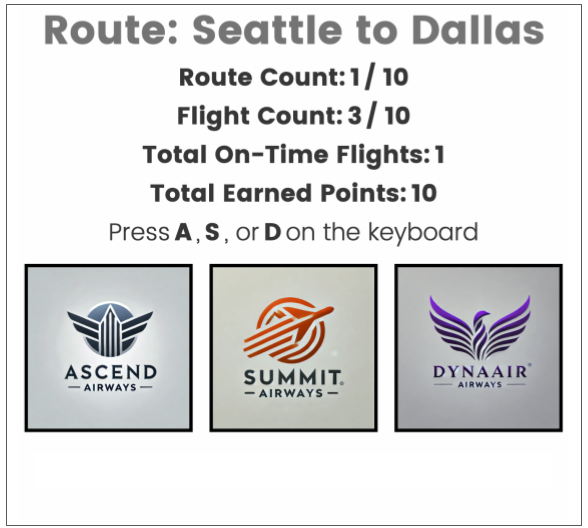}
        \caption{Before the participant chooses an airline.}
    \end{subfigure}
    \hfill
    \begin{subfigure}[t]{.45\textwidth}
        \centering
        \includegraphics[width=\textwidth]{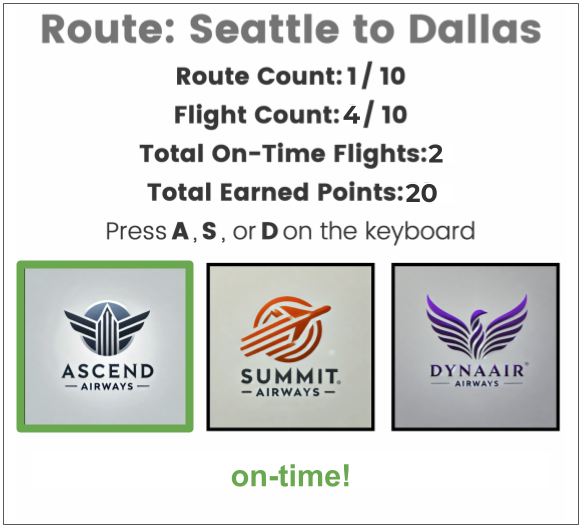}
        \caption{After the participant chooses Ascend and gets an on-time flight.}
    \end{subfigure}

    \caption{Airline choice environment used in the experiments. {\bf Route Count} represents the number of the current route that the participant is flying on, and {\bf Flight Count} represents the flight number on the current route. {\bf Total On-Time Flights} shows the accumulated number of on-time flights so far. {\bf Total Earned Points} shows the points that the participant has gotten so far over all flights. Each on-time flight earns $10$ points, and each $10$ points earns \$.005 bonus payment. After each choice, participants receive immediate binary feedback (on-time vs.\ delayed) before proceeding to the next flight.}
    \label{fig:3_choice_environment_new}
\end{figure}

We implement the hierarchical learning environment in an interactive, web-based sequential decision task framed as repeated airline choice across routes. In each round, participants choose among three fictional airlines---Ascend, Summit, and DynaAir---and then receive immediate binary feedback indicating whether the selected flight was on time or delayed. Figure~\ref{fig:3_choice_environment_new} illustrates the interface. 


Each route constitutes a distinct context. The airline identities remain fixed across routes, but their route-specific on-time rates vary from one route to the next according to the hierarchical environment described in \S\ref{ssec:environment_new}. The interface displays the three airline options and provides immediate feedback after each choice. It also includes a minimal progress summary to keep incentives transparent, while avoiding the presentation of explicit statistical summaries that could substitute for participants' own learning. Participants could respond using keyboard shortcuts mapped to the three airlines, consistent with the on-screen prompts. Before beginning the main task, participants received instructions explaining the choice environment, feedback structure, and bonus scheme, and they completed a short demonstration to familiarize themselves with the interface; see Web Appendix~\ref{web-sec:experiments_instructions} for details.

The experiment consists of \(M=10\) routes, each containing \(T=10\) flights, for a total of \(100\) sequential decisions per participant. We recruited 305 participants through the Prolific platform and compensated them at an effective rate of approximately \$8 per hour. To maintain incentive alignment, we also offered a performance-contingent bonus tied directly to realized outcomes: participants earned \$.005 for each on-time flight during the experiment. Because the task involved \(100\) flight decisions in total, the maximum attainable bonus was \$.50. This payment structure aligned participants' monetary incentives with the objective of the decision problem---maximizing the number of on-time flights---while preserving the simple, immediate-feedback structure of the task. This payment method is common in cognitive science and decision theory literature, used to make the experiments incentive-aligned \citep{schulz2019algorithmic,wu2018generalization,shanks2002re}.

Throughout the experiment, we record the chosen airline \(A_{m,t}\), the realized outcome \(y_{m,t}\), and the response time for each flight \(t\) on route \(m\). In addition, because the task is generated from the hierarchical environment in \S\ref{ssec:environment_new}, we retain the latent route-specific on-time rates \(\{\theta^{\ast}_{m,k}\}_{k=1}^{K}\) used to generate outcomes. Participants do not observe these latent rates, but we use them to compute benchmark performance measures in the subsequent analysis.

\subsection{Experimental Parameters}
\label{ssec:experiments_design_new}

We parameterize each airline-level Beta distribution by its mean \((\mu)\) and variance \((\sigma^2)\), because these quantities are more interpretable than the corresponding \((\alpha,\beta)\) parameters. For a Beta distribution, there is a one-to-one mapping between \((\alpha,\beta)\) and \((\mu,\sigma^2)\), given by $
\mu = \frac{\alpha}{\alpha + \beta}$ and $\sigma^2 = \frac{\alpha \beta}{(\alpha + \beta)^2(\alpha + \beta + 1)}$. The means determine the airlines' typical on-time performance across routes, while the variance determines how much route-specific performance fluctuates around those means. We use subscripts \(L\), \(M\), and \(H\) to denote the airlines with the lowest, middle, and highest overall performance, respectively.

Table~\ref{tab:three_arms_params_new} reports the experimental parameters and the number of participants assigned to each condition. We present the actual empirical mean and variance that the participants experienced in Web Appendix~\ref{web-sec:environment_check}.

\begin{table}[htp!]
    \centering
    \small
    \begin{tabular}{lcc}
    \hline
     & Low Var. & High Var. \\
    \hline
    \noalign{\vskip 2mm}
    Far Means &
      \begin{tabular}{c}
        \(\mu_L = .2, \mu_M = .5, \mu_H = .8, \sigma^2 = .02\) \\
        (N = 76)
      \end{tabular}
    &
      \begin{tabular}{c}
        \(\mu_L = .2, \mu_M = .5, \mu_H = .8, \sigma^2 = .04\) \\
        (N = 76)
      \end{tabular} \\[1ex]
    \noalign{\vskip 2mm}
    Close Means &
      \begin{tabular}{c}
        \(\mu_L = .4, \mu_M = .6, \mu_H = .8, \sigma^2 = .02\) \\
        (N = 76)
      \end{tabular}
    &
      \begin{tabular}{c}
        \(\mu_L = .4, \mu_M = .6, \mu_H = .8, \sigma^2 = .04\) \\
        (N = 77)
      \end{tabular} \\
    \noalign{\vskip 2mm}
    \hline
    \end{tabular}
    \caption{Three-airline experiment parameters. Means \((\mu_L,\mu_M,\mu_H)\) define the airlines' typical across-route on-time rates, and \(\sigma^2\) governs across-route variability in those rates through the airline-level Beta distributions.}
    \label{tab:three_arms_params_new}
\end{table}

The experiment follows a \(2\times 2\) design that varies two features of the environment: the separation between airline means and the amount of across-route variance. This yields four conditions: (1) Far Means with Low Variance, (2) Far Means with High Variance, (3) Close Means with Low Variance, and (4) Close Means with High Variance.

These two dimensions map naturally to different aspects of learning. Mean separation primarily shapes within-route learning. When the airline means are farther apart, the airlines tend to be more distinguishable within a given route, so participants can identify the better-performing airline with fewer observations. When the means are closer together, the route-specific on-time rates of the airlines are more similar, making feedback less diagnostic and within-route learning slower and noisier.

Variance primarily shapes cross-route learning. When variance is low, each airline's route-specific performance remains relatively stable around its overall mean, so experience from earlier routes provides a useful guide for later routes. This makes it easier for participants to infer persistent differences across airlines and transfer that knowledge across contexts. When variance is high, airline performance fluctuates more across routes, reducing the predictive value of prior experience from earlier routes and weakening the scope for cross-route transfer.

Taken together, these manipulations determine how informative the environment is for both learning within a route and generalizing across routes. The Far Means--Low Variance condition provides the cleanest environment for learning, because airlines are easier to distinguish within routes and their relative performance is more stable across routes. By contrast, the Close Means--High Variance condition is the most difficult, because the airlines are both harder to distinguish within a route and less predictable across routes. This design, therefore, allows us to examine not only whether consumers meta-learn but also whether the strength of such learning depends on how informative the environment is.

\section{Behavioral Evidence of Meta-Learning}
\label{sec:behavioral_evidence_meta_learning}

We now examine the main behavioral patterns in the experiment. The central empirical question is whether participants exhibit only within-route learning or whether they also transfer knowledge across routes. As discussed above, the experimental design allows for both types of learning. Within a route, participants can learn about the route-specific performance of the three airlines from realized outcomes. Across routes, participants may infer more general regularities about the airlines and use those regularities to improve decisions on later routes.

To study these patterns, we use best-airline selection and pseudo-regret as our main performance measures. Best-airline selection captures whether participants pick the best airline for any given flight. Pseudo-regret complements this measure by capturing the expected reward loss relative to the best airline on a given route, and is a standard performance metric in the stochastic bandit literature \citep{lai1985asymptotically, auer2002finite, bubeck2012regret, lattimore2020bandit}. In our setting, focusing on the best-airline selection and the {\it expected} reward loss are important because flight outcomes are stochastic. Even when a participant chooses the better airline on a route, the realized flight may still be delayed, and even an inferior airline may occasionally arrive on time. A performance measure based directly on realized outcomes, therefore, combines decision quality with idiosyncratic outcome noise. Instead, the two measures we use place greater emphasis on the quality of the choices. 

\paragraph{Best-airline selection.}
Let \(A_{m,t}\) denote the airline chosen on flight \(t\) of route \(m\), and let \(\theta^{\ast}_{m,k}\) denote the true on-time rate of airline \(k\) on route \(m\). We define the best-airline selection as: 
\begin{equation}
    I^{\text{best}}_{m,t} = \mathbf{1}\!\{A_{m,t} = \argmax_{k} \theta^{\ast}_{m,k}\},
    \label{eq:flight_best_airline_new}
\end{equation}
which equals 1 if the participant chooses the airline with the highest on-time rate on that route, and 0 otherwise. We also denote the total number of times that the participant chooses the best airline on route \(m\) as \(I_m^{best}\):
\begin{equation}
    I_m^{best} = \sum_{t=1}^{T} I_{m,t}^{best}.
\end{equation}
This metric measures the participant's overall success in finding the best airline on each route. 


\paragraph{Pseudo-regret.}
We define flight-wise pseudo-regret, \(r_{m,t}\), as the difference between the highest true on-time rate on route \(m\) and the true on-time rate of the airline chosen at flight \(t\) on that route:
\begin{equation}
    r_{m,t} = \bar{\theta}^{\ast}_{m} - \theta^{\ast}_{m,A_{m,t}}, \quad \textrm{where} \quad 
\bar{\theta}^{\ast}_{m} = \max_{k} \theta^{\ast}_{m,k}
    \label{eq:flight_regret_new}
\end{equation}
is the highest true on-time rate among the airlines on route \(m\). This quantity measures the expected performance loss from the participant's choice relative to the best available airline on that route. It equals zero when the participant selects an optimal airline and is positive otherwise.

We then aggregate flight-wise pseudo-regret within each route to obtain route-level pseudo-regret:
\begin{equation}
    R_{m} = \sum_{t=1}^{T} r_{m,t}
    = \sum_{t=1}^{T}\bigl(\bar{\theta}^{\ast}_{m} - \theta^{\ast}_{m,A_{m,t}}\bigr).
    \label{eq:round_regret_new}
\end{equation}
This measure summarizes how much expected performance is lost over the full sequence of decisions on route \(m\). Because the experimental environment records the latent route-specific on-time rates for all airlines, both \(r_{m,t}\) and \(R_m\) are directly computable for every participant and route.

These metrics provide a useful behavioral benchmark for distinguishing within-route learning from meta-learning across routes. If participants learn only within a route, then best-airline selection should increase, and pseudo-regret should decline over flights within that route, but there should be little systematic improvement from earlier routes to later ones. By contrast, if participants transfer knowledge across routes, then later routes should begin with higher best-airline selection and lower pseudo-regret. Combined with within-route learning, this should result in better overall visible performance on later routes. Thus, within-route changes in best-airline selection and pseudo-regret reflect route-specific learning, whereas across-route shifts are diagnostic of meta-learning.

\subsection{Best-Airline Selection and Regret Patterns}
\label{ssec:qualitative_regret_patterns}

\begin{figure}[htp!]
  \centering
  \includegraphics[width=.8\textwidth]{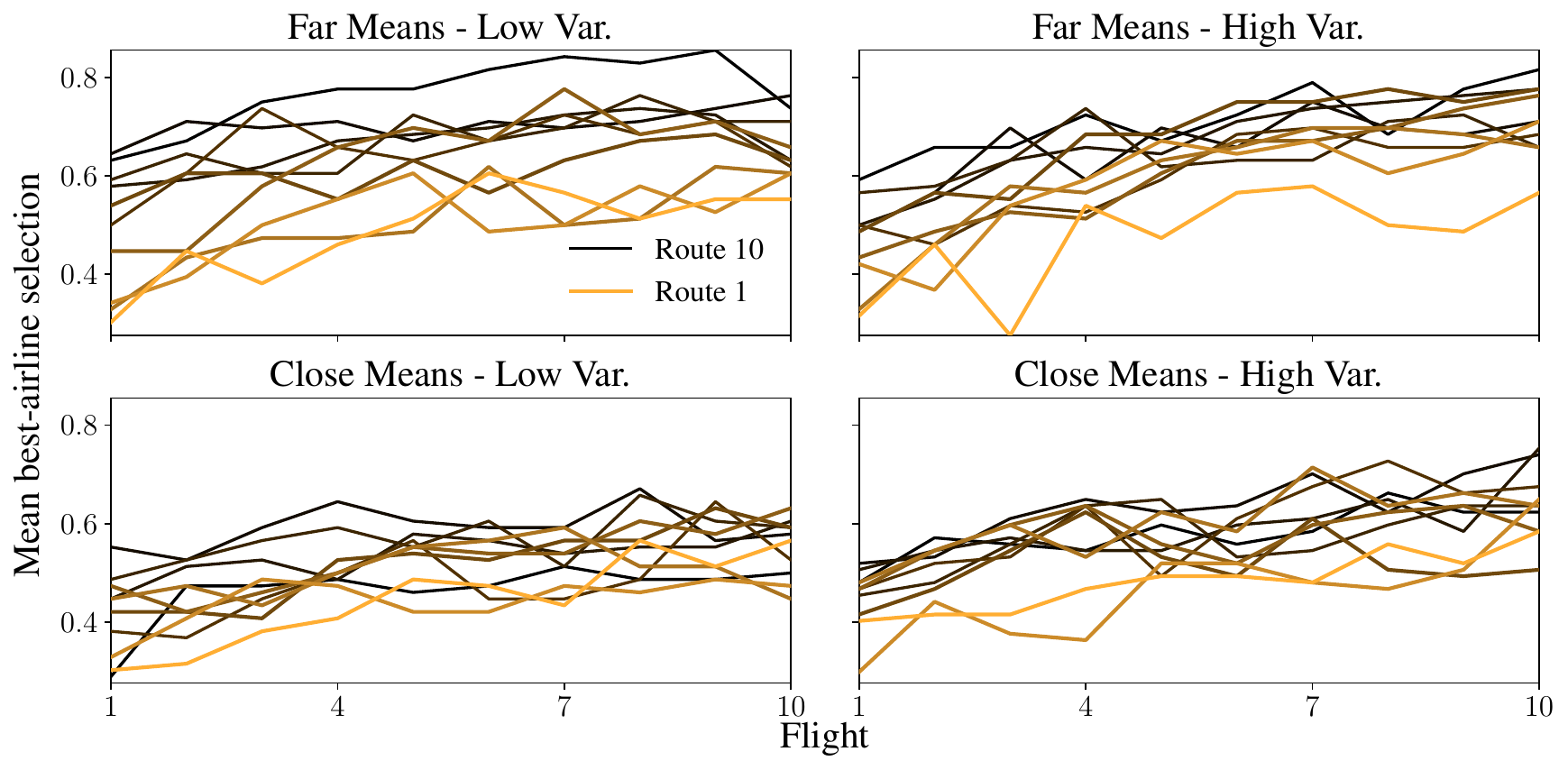}
    \caption{Participants' mean best-airline selection over flights for different routes.  Later routes generally lie above earlier ones, indicating improvement across routes in addition to the within-route improvement in best-airline selection.}
    \label{fig:3_qualitative_best_arm_selection}
\end{figure}

\begin{figure}[htp!]
  \centering
  \includegraphics[width=.8\textwidth]{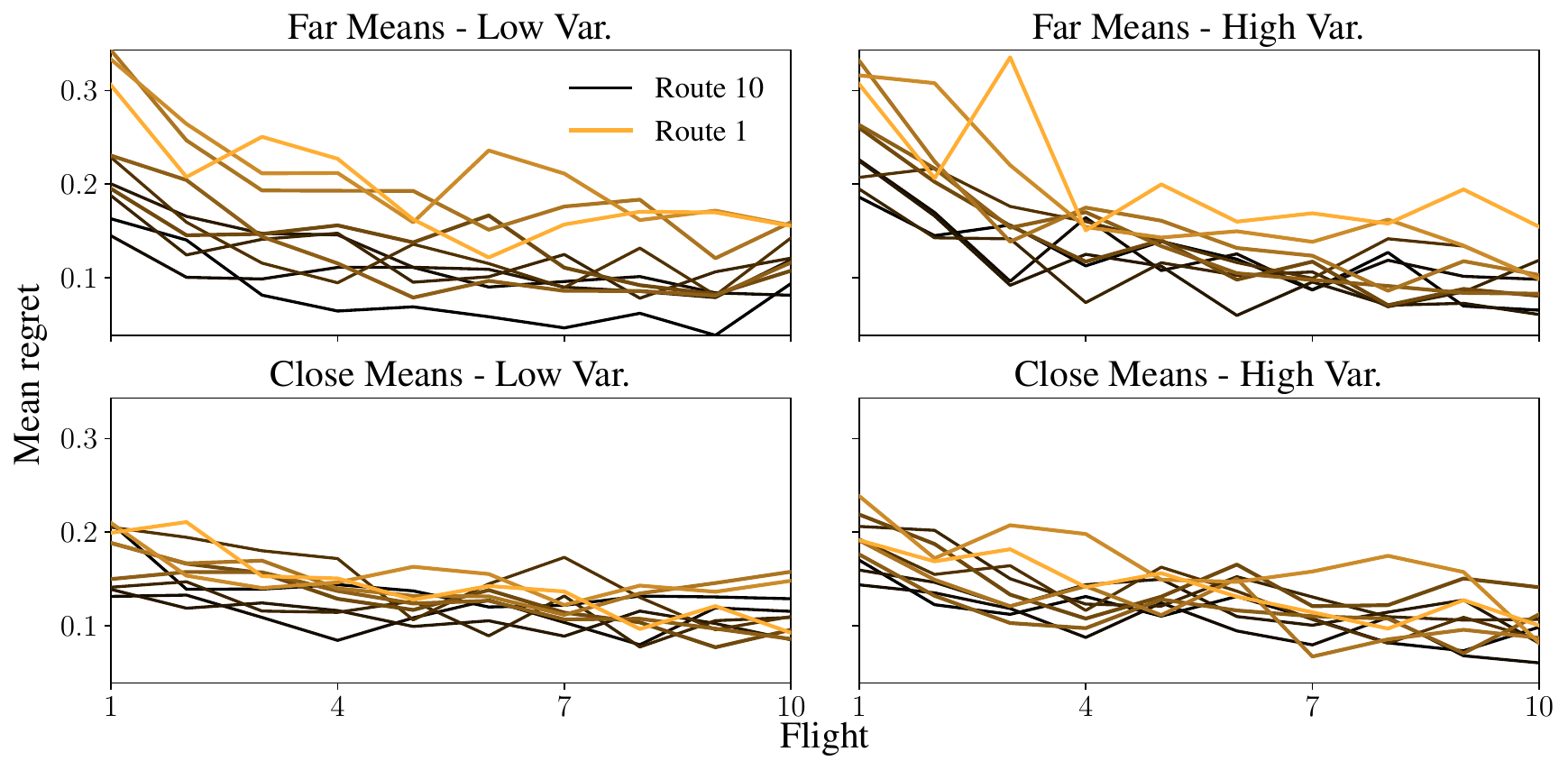}
    \caption{Participants' mean pseudo-regret over flights for different routes. Each line represents a distinct route, and earlier routes are in lighter shades. Later routes generally lie below earlier ones, indicating improvement across routes in addition to the within-route decline in regret.}
    \label{fig:3_qualitative_regrets}
\end{figure}

We begin with the most direct descriptive evidence. Figure~\ref{fig:3_qualitative_best_arm_selection} plots the mean best-airline selection per flight across all routes for the four experimental conditions. Each line corresponds to a distinct route, with earlier routes shown in lighter shades. If participants improve only within a route, the lines should slope upward within each route but remain broadly similar across routes. If participants meta-learn across routes, later-route lines should shift upward relative to earlier-route lines. Similarly, Figure~\ref{fig:3_qualitative_regrets} plots the mean pseudo-regret per flight across routes for all the experimental conditions. In the pseudo-regret plots, we consider downward-sloping lines as evidence of within-route learning, and later routes lying below earlier ones as evidence of meta-learning.

Figures~\ref{fig:3_qualitative_best_arm_selection} and ~\ref{fig:3_qualitative_regrets} reveal two clear patterns. First, the best-airline selection tends to increase, and pseudo-regret tends to decline within routes. This indicates that participants learn from realized outcomes and improve their choices over the ten flights of a given route. Second, both metrics also improve across routes. Accordingly, later routes generally start from a better baseline and remain above the earlier ones in the best-airline selection plots, and below the initial routes in the pseudo-regret plots, especially in the Far Means environments. This improvement suggests that participants do not treat each new route as an entirely fresh problem. Instead, they appear to carry information from previous routes into subsequent ones.

This pattern provides reduced-form evidence of meta-learning in our setting. Participants are not merely learning which airline performs best on the current route; they are also learning something more general that helps them make better choices when a new route begins. At the same time, the across-route improvement is not identical across experimental conditions. It is visually strongest in environments where airline differences are easier to distinguish and weaker in more difficult environments where means are closer together, and across-route variance is higher.

\subsection{Route-Level Evidence on Best-Airline Selection and Pseudo-Regret}
\label{ssec:regret_reward_regression_new}


We now formalize the descriptive evidence from the previous section using route-level regressions of best-airline selection and pseudo-regret on route number. This route-level analysis complements the flight-level plots by testing whether participants perform better on later routes than on earlier ones, which is the central reduced-form implication of meta-learning.

Let \(Z_{i,m}\) denote the route-level outcome of interest ($I^{best}_{i,m}$ or $R_{i,m}$) for participant \(i\) on route \(m\). We estimate
\begin{equation}
    Z_{i,m} = \alpha + \lambda m + \boldsymbol{\omega}^{\top}\!\left(m \times \mathbf{1}_{\text{experiment},i}\right) + \xi_i + u_{i,m},
    \label{eq:regret_reward_regression_new}
\end{equation}
where \(m\) is the route number, \(\mathbf{1}_{\text{experiment}, i}\) is a 3-vector of condition dummies for the non-baseline environments, \(\xi_i\) denotes participant fixed effects, and \(u_{i,m}\) is an idiosyncratic error term. The omitted condition is Far Means--Low Var., so the coefficient on \(m\) captures the across-route performance trend in that baseline condition, while the interaction terms capture how that trend differs in the other three environments. Participant fixed effects absorb time-invariant heterogeneity across participants, so identification comes from within-participant changes in performance over routes.

\begin{table}[htp!]
\centering
\small
\begingroup
\setlength{\tabcolsep}{3pt}
\begin{tabular}{lcccccc}
\toprule
& \multicolumn{3}{c}{\textbf{Best-Airline Selection}}
& \multicolumn{3}{c}{\textbf{Pseudo-Regret}}\\
\cmidrule(lr){2-4}\cmidrule(lr){5-7}
& Coefficient & Std. Error & \(\mathit{p}\)-value
& Coefficient & Std. Error & \(\mathit{p}\)-value\\
\midrule
Route number (\(m\))
& \(.294\) & \((.040)\) & \(<.001\)
& \(-.129\) & \((.017)\) & \(<.001\)\\
Far Means--High Var. \(\times m\)
& \(-.113\) & \((.059)\) & \(.055\)
& \(.043\) & \((.024)\) & \(.082\)\\
Close Means--Low Var. \(\times m\)
& \(-.205\) & \((.058)\) & \(<.001\)
& \(.099\) & \((.021)\) & \(<.001\)\\
Close Means--High Var. \(\times m\)
& \(-.167\) & \((.054)\) & \(.002\)
& \(.092\) & \((.023)\) & \(<.001\)\\
Constant
& \(5.002\) & \((.198)\) & \(<.001\)
& \(1.403\) & \((.083)\) & \(<.001\)\\
Participant fixed effects
& \multicolumn{3}{c}{\checkmark}
& \multicolumn{3}{c}{\checkmark}\\
\midrule
\(R^{2}\) (within)
& \multicolumn{3}{c}{.041}
& \multicolumn{3}{c}{.052}\\
Observations
& \multicolumn{3}{c}{3,050}
& \multicolumn{3}{c}{3,050}\\
Participants (clusters)
& \multicolumn{3}{c}{305}
& \multicolumn{3}{c}{305}\\
\bottomrule
\end{tabular}
\endgroup
\caption{Regression results for route-level best-airline selection and
pseudo-regret over route number with condition-specific slopes (baseline
condition: Far Means--Low Var.). Participant-clustered standard errors
are in parentheses.
Reported \(\mathit{p}\)-values are from two-sided cluster-robust Wald tests.}
\label{tab:3_regret_reward_new}
\end{table}

Table~\ref{tab:3_regret_reward_new} reports the results. Two findings stand out. First, performance improves across routes in the baseline condition. The coefficient on route number is positive for best-airline selection (\(.294\), \(\mathrm{SE}=.040\), \(\mathit{p}<.001\)) and negative for
pseudo-regret (\(-.129\), \(\mathrm{SE}=.017\), \(\mathit{p}<.001\)). This indicates that participants perform better on later routes than on earlier ones. Second, the rate of improvement depends on the environment. Relative to the Far Means--Low Var.\ condition, improvement is significantly weaker in the Close Means--Low Var.\ and Close Means--High Var.\ conditions, as indicated by the negative and positive interaction coefficients in the best-airline-selection and pseudo-regret specification, respectively. By contrast, the Far Means--High Var.\ interaction is smaller and not statistically distinguishable from the baseline slope.

Taken together, these results show that participants do not treat each route as an independent learning problem. Performance improves not only within routes but also across routes, consistent with participants carrying information from earlier to later routes. The strength of this cross-route improvement depends on the informativeness of the environment: it is strongest when airlines are easier to distinguish on average and weaker when the underlying differences are smaller. 

So far, we have analyzed best-airline selection and pseudo-regret, both of which rely on the latent on-time rates \(\theta_{m,k}^{\ast}\). One concern is that these measures evaluate improvement relative to environmental properties that participants never observe, rather than using information available to the participants themselves. In Web Appendix~\ref{web-ssec:choice_path_transfer}, we address this concern using only each participant's observed choices and outcomes, focusing on the first choice on each route. We show that participants are more likely to choose an airline whose posterior mean, constructed from their own earlier-route experience, is higher. This pattern provides additional evidence of cross-route knowledge transfer and is consistent with meta-learning.

Overall, these reduced-form patterns are consistent with cross-route belief transfer, but they do not distinguish among the mechanisms that could generate it. In particular, they do not tell us whether participants behave like fully optimal meta-learners or according to a boundedly rational approximation. Indeed, similar choices can be generated by very different cognitive processes. Structural work shows that distinguishing among these processes requires a fully specified model of how consumers form, update, and use beliefs over time \citep{erdem1996decision, ching2013learning, crawford2005uncertainty, hutchinson2008consumer}. Furthermore, recent work in consumer behavior argues for explicit Bayesian formulations of belief dynamics in experimental and survey data \citep{hofacker2024bayesian}. Aggregate statistics are useful as a diagnostic, but they are not sufficient to identify the underlying computations that govern transfer learning in choice.

Second, the choice of behavioral model has direct consequences for the managerial counterfactuals one wants to draw. \citet{meyer2016we} argue that small differences in how learning is modeled can produce qualitatively different policy implications. When a firm carries a misspecified learning model, the resulting beliefs and the policies derived from them can be systematically biased \citep{bohren2025misspecified}. In our setting, the gap between full integration of cross-route uncertainty (MetaDP) and a coarse, boundedly rational version of the same logic (BRMDP) maps to different choices in a managerial decision-making problem (see \S\ref{sec:managerial_implications}). A model that matches the reduced-form but misses the underlying mechanism can make the wrong decisions even if it captures the broad direction of learning.  We therefore turn to the formal policy classes---DP, MetaDP, and BRMDP---and compare them to the observed choice sequences in a likelihood-based framework.


\section{Dynamic Programming Models of Learning and Meta-Learning}
\label{sec:dp_models}

We model meta-learning as cross-route knowledge transfer in which experience across routes updates beliefs about the higher-order distributions that generate route-specific performance. To characterize this process, we introduce three dynamic-programming-based policy classes that differ in whether they transfer information across routes and, among the meta-learning policies, in how fully they integrate higher-order uncertainty. The baseline dynamic programming policy (\textbf{DP}) is Bayes-optimal within a route but treats routes independently, so each route begins from the same uninformative priors. The meta dynamic programming policy (\textbf{MetaDP}) updates higher-order beliefs across routes and fully integrates the resulting uncertainty into within-route dynamic planning. Finally, the boundedly rational meta dynamic programming policy (\textbf{BRMDP}) approximates MetaDP by limiting how finely the consumer integrates uncertainty across prior hypotheses.\footnote{We introduce a suite of non-meta learning policies in Web Appendices~\ref{web-ssec:improving_execution_robustness} and~\ref{web-ssec:non_meta_transfer_robustness} and compare them with our best-performing meta-learning policy.}

We build these policies in steps. We begin with Bayesian updating within a route, which is common to all three policy classes. We then present the no-transfer DP benchmark. Next, we develop Bayesian updating across routes and use it to define MetaDP. Finally, we introduce BRMDP as a boundedly rational approximation to MetaDP. Framing the model family in this way makes clear that the three policies differ not in the basic information structure of the environment, but in how they use available information and how much computation they allocate to planning.

For convenience, Table~\ref{tab:notation_dp_learning} summarizes the core notation for the within-route learning and the baseline DP.

\begin{table}[htp!]
\centering
\small
\caption{Notation for within-route Bayesian updating and baseline dynamic programming}
\label{tab:notation_dp_learning}
\resizebox{\textwidth}{!}{%
\begin{tabular}{p{.24\textwidth} p{.76\textwidth}}
\toprule
\textbf{Notation} & \textbf{Definition} \\
\midrule
$(\alpha_k,\beta_k)$ & Prior Beta parameters for airline $k$ in the baseline DP policy \\
$\boldsymbol{\alpha},\boldsymbol{\beta}$ & Vectors of prior Beta parameters across airlines \\
$\hat{\alpha}_k,\hat{\beta}_k$ & Posterior Beta parameters for airline $k$ within a route \\
$\hat{\boldsymbol{\alpha}},\hat{\boldsymbol{\beta}}$ & Vectors of posterior Beta parameters across airlines \\
$N_k^{+},N_k^{-}$ & Cumulative numbers of on-time and delayed flights for airline $k$ within a route \\
$\mathbf{N}^{+},\mathbf{N}^{-}$ & Vectors of on-time and delayed counts across airlines within a route \\
$N_{k,m}^{+},N_{k,m}^{-}$ & Numbers of on-time and delayed flights for airline $k$ on route $m$ \\
$\mathbf{N}_{m}^{+},\mathbf{N}_{m}^{-}$ & Vectors of on-time and delayed counts across airlines on route $m$ \\
$T_{k,m}$ & Number of times airline $k$ is chosen on route $m$, equal to $N_{k,m}^{+}+N_{k,m}^{-}$ \\
$\hat{\theta}_{k,t}$ & Posterior mean on-time rate of airline $k$ at flight $t$ in the baseline DP policy \\
$\mathbf{e}_k$ & Unit vector with a $1$ in position $k$ and zeros elsewhere \\
$s_t^{DP}$ & State in the baseline DP policy: $(\mathbf{N}^{+},\mathbf{N}^{-};\boldsymbol{\alpha},\boldsymbol{\beta})$ \\
$V_{\epsilon}(k \mid s_t^{DP})$ & Baseline DP action value from choosing airline $k$ at state $s_t^{DP}$ under policy noise $\epsilon$ \\
$W_{\epsilon}(\cdot)$ & Baseline DP state value under policy noise $\epsilon$ \\
$\pi_{\epsilon}^{DP}(\cdot \mid \cdot)$ & $\epsilon$-greedy baseline DP choice rule \\
\bottomrule
\end{tabular}
}
\end{table}

\subsection{Bayesian Updating within a Route}
\label{ssec:within_bayesian_update_reordered}

We begin with Bayesian updating within a route. This is the common building block for all three policy classes. Because each flight outcome is binary (\textit{on-time} versus \textit{delayed}), we model outcomes as Bernoulli and place a Beta prior on each airline's on-time rate. The Beta prior is conjugate to the Bernoulli likelihood and therefore yields closed-form posterior updates \citep{gelman1995bayesian}.

Let \((\hat{\boldsymbol{\alpha}},\hat{\boldsymbol{\beta}})\) denote the current posterior parameter vectors, where $
\hat{\boldsymbol{\alpha}} = \bigl(\hat{\alpha}_1,\ldots,\hat{\alpha}_K\bigr),
\hat{\boldsymbol{\beta}} = \bigl(\hat{\beta}_1,\ldots,\hat{\beta}_K\bigr)$, 
and \((\hat{\alpha}_k,\hat{\beta}_k)\) are the posterior parameters for airline \(k\). These posterior parameters summarize the information accumulated from past realized outcomes within the route and define the consumer's current beliefs about each airline's on-time rate.

Let \(N_k^+\) and \(N_k^-\) denote the cumulative numbers of on-time and delayed flights experienced with airline \(k\) up to flight \(t\). Then the posterior mean on-time rate of airline \(k\) at time \(t\) is
\begin{equation}
    \hat{\theta}_{k,t}
    =
    \frac{\hat{\alpha}_k}{\hat{\alpha}_k+\hat{\beta}_k}
    =
    \frac{\alpha_k + N_k^+}{\alpha_k + \beta_k + N_k^+ + N_k^-},
    \label{eq:posterior_mean_reordered}
\end{equation}
where \((\alpha_k,\beta_k)\) are the prior Beta parameters for airline \(k\). We suppress time subscripts on \(N_k^+\), \(N_k^-\), \(\hat{\alpha}_k\), and \(\hat{\beta}_k\) because the posterior for airline \(k\) depends only on the cumulative number of on-time and delayed outcomes observed for that airline up to flight \(t\), not on the calendar position of those observations.

To simplify notation, let $\mathbf{N}^+ = (N_1^+,\ldots,N_K^+),
\qquad \mathbf{N}^- = (N_1^-,\ldots,N_K^-)$. For example, if the consumer chooses airline \(k\) and the realized outcome is on time, then \(N_k^+ \leftarrow N_k^+ + 1\), or equivalently \(\mathbf{N}^+ \leftarrow \mathbf{N}^+ + \mathbf{e}_k\), where \(\mathbf{e}_k\) is the unit vector with a \(1\) in the \(k\)th position and zeros elsewhere.

At the end of each flight \(t \in \{1,\ldots,T\}\), suppose the consumer chose airline \(k\). After observing the realized outcome \(y_t \in \{0,1\}\), they update the posterior accordingly. Because of conjugacy, the posterior remains in the Beta family after each update. Algorithm~\ref{alg:bayesian_update_reordered} summarizes these updating steps. This within-route updating rule is shared by all the policies considered in the paper. 

\begin{algorithm}[htp!]
\small
\caption{Bayesian Updating within a Route (Bernoulli--Beta)}
\label{alg:bayesian_update_reordered}
\begin{algorithmic}[1]
\State \textbf{Input:} current counts $(\mathbf{N}^+,\mathbf{N}^-)$ and posterior parameters $(\hat{\boldsymbol{\alpha}},\hat{\boldsymbol{\beta}})$; action $A_t = k$; realized outcome $y_t \in \{0,1\}$

\If{$y_t=1$} \Comment{flight is \textit{on-time}}
  \State $\mathbf{N}^+ \gets \mathbf{N}^+ + \mathbf{e}_k$, \quad $\mathbf{N}^- \gets \mathbf{N}^-$
  \State \hspace{1.5em} (equivalently, $\hat{\alpha}_k \gets \hat{\alpha}_k + 1$, \quad $\hat{\beta}_k \gets \hat{\beta}_k$)
\Else \Comment{flight is \textit{delayed}}
  \State $\mathbf{N}^+ \gets \mathbf{N}^+$, \quad $\mathbf{N}^- \gets \mathbf{N}^- + \mathbf{e}_k$
  \State \hspace{1.5em} (equivalently, $\hat{\alpha}_k \gets \hat{\alpha}_k$, \quad $\hat{\beta}_k \gets \hat{\beta}_k + 1$)
\EndIf
\State Updated posterior mean: $\displaystyle \hat{\theta}_{k,t} \gets \frac{\hat{\alpha}_k}{\hat{\alpha}_k+\hat{\beta}_k}$
\end{algorithmic}
\end{algorithm}

\subsection{Baseline Dynamic Programming Policies (DP)}
\label{ssec:DP_reordered}

We now introduce the baseline dynamic programming policies. These policies provide the no-transfer benchmark for the paper. They use the Bayesian updating rule from \S\ref{ssec:within_bayesian_update_reordered}, but they treat each route as an independent learning problem. As a result, every new route begins from the same uninformative priors, and no information is carried over from earlier routes. This benchmark is useful because it isolates what can be learned from repeated outcomes \emph{within} a route alone.

We consider two variants of the baseline DP policy. The first is a deterministic benchmark that always chooses a value-maximizing airline. The second is a stochastic \(\epsilon\)-greedy version that assigns most of its probability to value-maximizing airlines while allowing for uniform choice noise. Because neither policy transfers information across routes, we suppress the route subscript \(m\) throughout this subsection.

\paragraph{State representation.}
At flight \(t\), the state is $
s_t^{DP} = (\mathbf{N}^+, \mathbf{N}^-; \boldsymbol{\alpha}, \boldsymbol{\beta})$.
 Given this state, the posterior parameters are given by
$
\hat{\boldsymbol{\alpha}}=\boldsymbol{\alpha}+\mathbf{N}^+,
\hat{\boldsymbol{\beta}}=\boldsymbol{\beta}+\mathbf{N}^-$,
and the flight index is pinned down by $
\|\mathbf{N}^+\|_1+\|\mathbf{N}^-\|_1=t-1$. Because the horizon is finite, the problem is non-stationary through the number of flights remaining.

\paragraph{State transitions.}
The within-route Bayesian updating rule from \S\ref{ssec:within_bayesian_update_reordered} governs transitions from \(s_t^{DP}\) to \(s_{t+1}^{DP}\). After choosing airline \(k\) at flight \(t\),
\begin{align*}
\text{On-time with probability } \hat{\theta}_{k,t}&:& 
(\mathbf{N}^{\prime +},\mathbf{N}^{\prime -}) 
= 
(\mathbf{N}^+ + \mathbf{e}_k,\mathbf{N}^-),  \\
\text{Delayed with probability } 1 - \hat{\theta}_{k,t}&:& 
(\mathbf{N}^{\prime +},\mathbf{N}^{\prime -}) 
= 
(\mathbf{N}^+,\mathbf{N}^- + \mathbf{e}_k). 
\end{align*}
These transition probabilities apply to all DP variants.

\paragraph{Deterministic DP benchmark.}
Under the deterministic benchmark, the consumer chooses a value-maximizing airline at every state. This corresponds to the special case \(\epsilon=0\). The action value from choosing airline \(k\) at state \(s_t^{DP}\) is
\begin{equation}
V_{0}(k \mid s_t^{DP})
=
\hat{\theta}_{k,t}
+
\gamma \,
\mathbb{E}\!\left[
  W_{0}(s_{t+1}^{DP})
  \,\middle|\,
  s_t^{DP},\, A_t = k
\right],
\label{eq:choice_value_function_reordered}
\end{equation}
where \(\hat{\theta}_{k,t}\) is the posterior mean on-time rate of airline \(k\). Because outcomes are binary, \(\hat{\theta}_{k,t}\) is also the expected current-period payoff.

Substituting the state-transition probabilities into Equation~\eqref{eq:choice_value_function_reordered} yields
\begin{equation}
    V_{0}(k \mid \mathbf{N}^+,\mathbf{N}^-;\boldsymbol{\alpha},\boldsymbol{\beta})
    =
    \hat{\theta}_{k,t}
    +
    \gamma \Bigl[
      \hat{\theta}_{k,t}\, W_{0}(\mathbf{N}^+ + \mathbf{e}_k,\mathbf{N}^-;\boldsymbol{\alpha},\boldsymbol{\beta})
      +
      \bigl(1-\hat{\theta}_{k,t}\bigr)\, W_{0}(\mathbf{N}^+,\mathbf{N}^- + \mathbf{e}_k;\boldsymbol{\alpha},\boldsymbol{\beta})
    \Bigr],
\label{eq:Vt_reordered}
\end{equation}
where
\begin{equation}
    W_{0}(\mathbf{N}^+,\mathbf{N}^-;\boldsymbol{\alpha},\boldsymbol{\beta})
    =
    \max_{k} V_{0}(k \mid \mathbf{N}^+,\mathbf{N}^-;\boldsymbol{\alpha},\boldsymbol{\beta})
\label{eq:W0_reordered}
\end{equation}
is the expected value of being in state \((\mathbf{N}^+,\mathbf{N}^-;\boldsymbol{\alpha},\boldsymbol{\beta})\). Thus, the deterministic DP benchmark chooses an airline in \(\arg\max_k V_0(k \mid s_t^{DP})\), breaking ties uniformly when needed. At the terminal flight, there is no continuation value, so
$
V_{0}(k \mid s_T^{DP}) = \hat{\theta}_{k,T}$.

This policy can be solved using backward induction from \(t=T\) to \(t=1\) and then used to compute the action values at every feasible state; see Web Appendix~\ref{web-ssec:backward_induction_algo} for details of the backward induction algorithm.

\paragraph{Stochastic \(\epsilon\)-greedy DP policy.}
Although the deterministic benchmark is a useful normative reference, it imposes a sharp behavioral implication: at every state, the consumer always selects a value-maximizing airline, exactly following the policy. Prior work shows that consumers often deviate from exact dynamic optimization in sequential choice settings. As such, a more natural policy is the \(\epsilon\)-greedy policy, which provides a parsimonious stochastic relaxation of the deterministic benchmark \citep{sutton1998reinforcement}. In our behavioral interpretation, \(\epsilon\) captures value-independent choice noise rather than information-seeking exploration, which is already accounted for by the continuation value.

Under the \(\epsilon\)-greedy policy, the consumer assigns total probability \(1-\epsilon\) to the set of value-maximizing airlines and distributes the remaining probability \(\epsilon\) uniformly across all \(K\) airlines. At state \(s_t^{DP} =(\mathbf{N}^+,\mathbf{N}^-;\boldsymbol{\alpha},\boldsymbol{\beta})\), the policy is
\begin{equation}
\label{eq:e_greedy_probability_func_reordered}
\pi^{DP}_{\epsilon}\!\bigl(\tilde{k} \mid \mathbf{N}^+,\mathbf{N}^-;\boldsymbol{\alpha},\boldsymbol{\beta}\bigr)
=
(1-\epsilon)\,
\frac{
\mathbf{1}\!\Bigl\{
\tilde{k} \in \arg\max\limits_{k}
V_{\epsilon}\!\bigl(k \mid \mathbf{N}^+,\mathbf{N}^-;\boldsymbol{\alpha},\boldsymbol{\beta}\bigr)
\Bigr\}
}{
\left|
\arg\max\limits_{k}
V_{\epsilon}\!\bigl(k \mid \mathbf{N}^+,\mathbf{N}^-;\boldsymbol{\alpha},\boldsymbol{\beta}\bigr)
\right|
}
+
\frac{\epsilon}{K}.
\end{equation}
The first term assigns total probability \(1-\epsilon\) to the set of value-maximizing airlines and splits that mass evenly in case of ties. The second term distributes the choice-noise probability uniformly across all airlines. When \(\epsilon=0\), the policy collapses to the deterministic DP benchmark. An alternative way to induce stochasticity is to use a softmax policy \citep{bhui2021resource}. We consider this formulation in an extension in Web Appendix~\ref{web-ssec:softmax_spec} and focus on the $\epsilon-$greedy formulation in the main text. 


Because future selves follow the same \(\epsilon\)-greedy rule, the continuation value must average over the action probabilities induced by that rule. The corresponding state value is
\begin{equation}
\label{eq:We_generic_reordered}
W_{\epsilon}\!\bigl(\mathbf{N}^+,\mathbf{N}^-;\boldsymbol{\alpha},\boldsymbol{\beta}\bigr)
=
\mathbb{E}_{k \sim \pi^{DP}_{\epsilon}(\cdot \mid \mathbf{N}^+,\mathbf{N}^-;\boldsymbol{\alpha},\boldsymbol{\beta})}
\left[
V_{\epsilon}\!\bigl(k \mid \mathbf{N}^+,\mathbf{N}^-;\boldsymbol{\alpha},\boldsymbol{\beta}\bigr)
\right].
\end{equation}
Substituting Equation~\eqref{eq:e_greedy_probability_func_reordered} gives
\begin{equation}
\label{eq:We_uniform_reordered}
W_{\epsilon}\!\bigl(\mathbf{N}^+,\mathbf{N}^-;\boldsymbol{\alpha},\boldsymbol{\beta}\bigr)
=
(1-\epsilon)\,\max_{k}
V_{\epsilon}\!\bigl(k \mid \mathbf{N}^+,\mathbf{N}^-;\boldsymbol{\alpha},\boldsymbol{\beta}\bigr)
+
\frac{\epsilon}{K}\sum_{k=1}^{K}
V_{\epsilon}\!\bigl(k \mid \mathbf{N}^+,\mathbf{N}^-;\boldsymbol{\alpha},\boldsymbol{\beta}\bigr).
\end{equation}
The recursive action value under the \(\epsilon\)-greedy policy is therefore
\begin{equation}
\label{eq:Ve_eps_single_reordered}
V_{\epsilon}\!\bigl(k \mid \mathbf{N}^+,\mathbf{N}^-;\boldsymbol{\alpha},\boldsymbol{\beta}\bigr)
=
\hat{\theta}_{k,t}
+
\gamma \Bigl[
\hat{\theta}_{k,t}\,
W_{\epsilon}\!\bigl(\mathbf{N}^+ + \mathbf{e}_k,\mathbf{N}^-;\boldsymbol{\alpha},\boldsymbol{\beta}\bigr)
+
\bigl(1-\hat{\theta}_{k,t}\bigr)\,
W_{\epsilon}\!\bigl(\mathbf{N}^+,\mathbf{N}^- + \mathbf{e}_k;\boldsymbol{\alpha},\boldsymbol{\beta}\bigr)
\Bigr].
\end{equation}
At the terminal flight, $
V_{\epsilon}\!\bigl(k \mid \mathbf{N}^+,\mathbf{N}^-;\boldsymbol{\alpha},\boldsymbol{\beta}\bigr) = \hat{\theta}_{k,T}$, 
and
$W_{\epsilon}\!\bigl(\mathbf{N}^+,\mathbf{N}^-;\boldsymbol{\alpha},\boldsymbol{\beta}\bigr) = (1-\epsilon)\,\max_{k} \hat{\theta}_{k,T}
+ \frac{\epsilon}{K}\sum_{k=1}^{K} \hat{\theta}_{k,T}$.
Backward induction over Equation~\eqref{eq:Ve_eps_single_reordered} then yields \(V_{\epsilon}\) and \(W_{\epsilon}\) at all feasible states. Setting \(\epsilon=0\) yields the deterministic benchmark, while \(\epsilon>0\) gives the stochastic \(\epsilon\)-greedy policy. See Web Appendix~\ref{web-ssec:baseline_dp_algo} for the DP pseudocode.

\subsection{Bayesian Updating across Routes}
\label{ssec:cross_bayesian_update_reordered}

The baseline DP benchmark treats every route as a fresh learning problem. We now relax that restriction and allow the consumer to transfer information across routes. To model meta-learning, we introduce a hyper-posterior over candidate airline-level prior specifications. Updating this hyper-posterior allows experience on earlier routes to change beliefs about the distributions that generate route-specific performance.

Two features of our formulation are new relative to prior bandit meta-learning work of \cite{kveton2021meta}. First, we keep the posterior update in closed form via Beta--Binomial marginal likelihoods, which makes it possible to discretize the hyper-posterior coarsely. This discretization is what allows us to define a one-parameter family of boundedly rational models in \S\ref{ssec:BRMDP_reordered}. Second, we use the resulting hyper-posterior as an input to a finite-horizon dynamic program within each route. We formalize that extension in \S\ref{ssec:metaDP_reordered}. For now, we focus only on the updating rule itself.


For convenience, Table~\ref{tab:notation_meta_reordered} summarizes the notation used for cross-route learning and the MetaDP policy.

\begin{table}[htp!]
\centering
\small
\caption{Notation for cross-route learning and MetaDP}
\label{tab:notation_meta_reordered}
\begin{tabular}{p{.24\textwidth} p{.68\textwidth}}
\toprule
\textbf{Notation} & \textbf{Definition} \\
\midrule
$(\alpha_k^{(j)},\beta_k^{(j)})$ & Beta prior parameters for airline $k$ under hypothesis $P^{(j)}$ \\
$(\boldsymbol{\alpha}^{(j)},\boldsymbol{\beta}^{(j)})$ & Vectors of prior Beta parameters across airlines under hypothesis $P^{(j)}$ \\
$\mathcal{P}=\{P^{(j)}\}_{j=1}^{J}$ & Set of candidate prior hypotheses used for cross-route learning \\
$P^{\ast}$ & The element of $\mathcal P$ that is the true prior in the environment\\
$Q$ & Initial hyper-prior over hypotheses in $\mathcal{P}$ \\
$Q_m(j)$ & Hyper-posterior weight before route $m$ on hypothesis $P^{(j)}$ \\
$f_m(j)$ & Marginal likelihood of route-$m$ outcomes under hypothesis $P^{(j)}$ \\
$\hat{\theta}_{k,t,m}^{(j)}$ & Posterior mean on-time rate of airline $k$ at flight $t$ on route $m$ under hypothesis $P^{(j)}$ \\
$s_{t,m}^{\mathrm{MetaDP}}$ & State in MetaDP at flight $t$ on route $m$: $(\mathbf{N}_{m}^{+},\mathbf{N}_{m}^{-};Q_m)$ \\
$V_{\epsilon}^{(j)}(k \mid \cdot)$ & Hypothesis-conditioned MetaDP action value under prior hypothesis $P^{(j)}$ \\
$\bar{V}_{\epsilon}(k \mid \cdot)$ & Integrated MetaDP action value, averaged over hypotheses using $Q_m$ \\
$\bar{W}_{\epsilon}(\cdot)$ & Integrated MetaDP state value under the $\epsilon$-greedy rule \\
$\pi_{\epsilon}^{MetaDP}(\cdot \mid \cdot)$ & $\epsilon$-greedy MetaDP choice rule \\
\bottomrule
\end{tabular}
\end{table}

\paragraph{Hypothesis space and hyper-prior.}
Recall from Equation~\eqref{eq:mean_draw_new} that at the beginning of each route \(m\), the environment draws the true on-time rate of airline \(k\), denoted \(\theta^{\ast}_{m,k}\), from an airline-specific Beta distribution:
\[
\theta^{\ast}_{m,k} \sim \text{Beta}(\alpha_k^{\ast},\beta_k^{\ast}),
\qquad
\forall k \in \{1,\ldots,K\}, \quad \forall m \in \{1,\ldots,M\}.
\]
These route-specific on-time rates are unobserved. However, by the time the consumer reaches route \(m\), they have accumulated experience from the previous \(m-1\) routes. To capture cross-route learning, we define two objects: a finite set of prior hypotheses and a hyper-prior over them.

Let \(\mathcal{P}=\{P^{(j)}\}_{j=1}^J\) denote the set of prior hypotheses. Each hypothesis \(P^{(j)}\) specifies airline-specific Beta prior parameters \(\{(\alpha_k^{(j)},\beta_k^{(j)})\}_{k=1}^K\). Intuitively, \(\mathcal{P}\) collects the candidate prior specifications that the consumer regards as plausible across routes. We assume that the environment's true prior  is one element of this class, denoted
\(
P^{\ast} = P^{(j^{\ast})},
~
j^{\ast} \in \{1,\ldots,J\}.
\)
Because the consumer does not know which hypothesis is \(P^{\ast}\), they begin with a hyper-prior \(Q_1\).\footnote{A reasonable choice for this hyper-prior is an uninformative one, which gives equal weights to each prior hypothesis--i.e. $Q_1(j) = 1/J, \; \forall~j \in \{1,\ldots,J\}$.} More generally, before route \(m\), their beliefs are summarized by the hyper-posterior \(Q_m\), where
\[
Q_m(j) = \Pr\!\left(P^{\ast}=P^{(j)} \mid \{(N_{k,r}^{+},N_{k,r}^{-})_{k=1}^{K}\}_{r=1}^{m-1}\right),
\qquad
\sum_{j=1}^J Q_m(j) = 1.
\]
Thus, cross-route learning amounts to updating \(Q_m\) after each route so that the weight shifts toward hypotheses that better explain the observed route-level outcomes.

\paragraph{Hyper-posterior update.}
After route \(m\), the consumer updates the hyper-posterior by Bayes' rule:
\begin{equation}
    Q_{m+1}(j)
    =
    \dfrac{f_m(j)\cdot Q_m(j)}{\sum_{\ell=1}^{J} f_m(\ell)\cdot Q_m(\ell)},
    \label{eq:meta_posterior_update_reordered}
\end{equation}
where
\begin{equation}
    f_m(j)
    =
    \prod_{k=1}^{K}
    \frac{\Gamma(\alpha_{k}^{(j)}+\beta_{k}^{(j)})}{\Gamma(\alpha_{k}^{(j)})\,\Gamma(\beta_{k}^{(j)})}
    \cdot
    \frac{\Gamma(\alpha_{k}^{(j)}+N_{k,m}^{+})\,\Gamma(\beta_{k}^{(j)}+N_{k,m}^{-})}{\Gamma(\alpha_{k}^{(j)}+\beta_{k}^{(j)}+T_{k,m})}.
    \label{eq:meta_update_factor_reordered}
\end{equation}
For route \(m\), \(N_{k,m}^{+}\) and \(N_{k,m}^{-}\) denote, respectively, the numbers of on-time and delayed flights experienced with airline \(k\), and \(T_{k,m}=N_{k,m}^{+}+N_{k,m}^{-}\) is the number of times airline \(k\) is chosen on that route. This route-level evidence term is the Beta--Binomial marginal likelihood induced by Beta--Bernoulli conjugacy.

Equation~\eqref{eq:meta_posterior_update_reordered} is Bayes' rule over the finite hypothesis class \(\mathcal{P}\): the next hyper-posterior \(Q_{m+1}(j)\) equals the current weight \(Q_m(j)\) times how well hypothesis \(j\) explains the observed on-time/delay counts. With Beta--Bernoulli conjugacy, this ``fit'' reduces to the Beta--Binomial terms (Gamma ratios) in Equation~\eqref{eq:meta_update_factor_reordered}. Hypotheses whose priors place more mass on the observed \(N_{k,m}^{+}\) and \(N_{k,m}^{-}\) gain weight, and mismatched hypotheses lose weight. Over routes, \(Q_m\) shifts posterior mass toward the true prior \(P^{\ast}\). For convenience, let $
\mathbf{N}_m^{+}=(N_{1,m}^{+},\ldots,N_{K,m}^{+})$, $
\mathbf{N}_m^{-}=(N_{1,m}^{-},\ldots,N_{K,m}^{-})$
denote the vectors of on-time and delayed counts across all airlines on route \(m\). Algorithm~\ref{alg:cross_route_update_reordered} summarizes the resulting cross-route Bayesian updating rule.

\begin{algorithm}[htp!]
\small
\caption{Cross-Route Hyper-Posterior Update}
\label{alg:cross_route_update_reordered}
\begin{algorithmic}[1]
\State \textbf{Inputs:} $\mathcal{P}=\{(\boldsymbol{\alpha}^{(j)},\boldsymbol{\beta}^{(j)})\}_{j=1}^{J}$, initial hyper-prior $Q_1$
\For{$m=1,\ldots,M$}
  \State Play route $m$ with a within-route policy and collect counts $\{\mathbf{N}_m^+,\mathbf{N}_m^-\}$
  \For{$j=1,\ldots,J$}
    \State Compute evidence $f_m(j)$ using Equation~\eqref{eq:meta_update_factor_reordered}
    \State $Q_{m+1}(j)\gets \text{normalized } f_m(j)\,Q_m(j)$ \hfill (Equation~\eqref{eq:meta_posterior_update_reordered})
  \EndFor
\EndFor
\end{algorithmic}
\end{algorithm}

\paragraph{Illustrative example.}
A simple two-airline example clarifies these objects. Suppose the consumer is choosing between Ascend and Summit Airways and considers two hypotheses: $j = 1$, under which Ascend is better than Summit, and $j = 2$, under which Summit is better than Ascend. One possible parameterization is shown in Table~\ref{tab:prior_categories_example_reordered}. The consumer starts with hyper-prior $Q_1 = (.5, .5)$, reflecting equal initial weight on the two hypotheses. Further, suppose that Ascend is actually better than Summit, and the true airline-level Beta parameters are $
(\alpha_A^*, \beta_A^*) = (3, 1)$ and $(\alpha_S^*, \beta_S^*) = (2, 1)$. Consequently, $P^{\ast} = P^{(1)}$, since $P^{(1)}$ contains the true priors in Table~\ref{tab:prior_categories_example_reordered}. The consumer does not observe this directly. Instead, after each route, they use the realized flight outcomes to update $Q_m$. As more routes are observed, the hypothesis that better explains the accumulated outcomes receives increasing weight.

\begin{table}[ht]
\centering
\small
\singlespacing
\begin{tabular}{c C{2.2em} C{2.2em} C{2.2em} C{2.2em}}
\toprule
& \multicolumn{2}{c}{\shortstack{Ascend is Better\\(\(j=1\))}} &
  \multicolumn{2}{c}{\shortstack{Summit is Better\\(\(j=2\))}} \\
\cmidrule(lr){2-3}\cmidrule(lr){4-5}
& $\alpha$ & $\beta$ & $\alpha$ & $\beta$ \\
\midrule
Ascend & 3 & 1 & 2 & 1 \\
Summit & 2 & 1 & 3 & 1 \\
\midrule
$Q_1(j)$ & \multicolumn{2}{c}{.5} & \multicolumn{2}{c}{.5} \\
\bottomrule
\end{tabular}
\caption{Example of two prior hypotheses. Here, \(P^{(1)}=\bigl(\text{Beta}(3,1),\text{Beta}(2,1)\bigr)\) encodes the belief that Ascend is better than Summit, while \(P^{(2)}=\bigl(\text{Beta}(2,1),\text{Beta}(3,1)\bigr)\) encodes the reverse ranking. The consumer starts with hyper-prior \(Q_1=(.5,.5)\).}
\label{tab:prior_categories_example_reordered}
\end{table}

\subsection{Meta Dynamic Programming Policy (MetaDP)}
\label{ssec:metaDP_reordered}

We now introduce the first policy class that allows cross-route learning: Meta Dynamic Programming (MetaDP). This policy combines the hyper-posterior from \S\ref{ssec:cross_bayesian_update_reordered} with finite-horizon dynamic programming within each route. The consumer learns across routes by updating the distribution over prior hypotheses, and then uses that learned distribution directly in route-level planning.


The construction combines two ideas from separate literatures. Within machine learning, the recent meta-bandit literature shows how an algorithm can learn priors by maintaining a hyper-posterior over candidate task priors and using it across tasks \citep{kveton2021meta, basu2021no, hong2022hierarchical,  nabi2022bayesian, guan2024improved}. However, this literature focuses on algorithmic development and pure performance, not building models of human consumers. From the dynamic decision-making literature in marketing (and more broadly), we adopt the Bellman recursion that values current actions partly for the information they generate and the effect of that information on future choices \citep{bellman1966dynamic,degroot2005optimal,bellman1956problem}. These models typically take a single fixed prior as given and do not formalize how that prior itself is learned from prior experience in related contexts. 

MetaDP draws on these two literatures. Its key feature is that it propagates the full hyper-posterior through the within-route Bellman recursion. Rather than selecting one prior hypothesis and solving the route conditional on that choice, the consumer evaluates each airline by integrating over all candidate priors, weighted by their probabilities in the current hyper-posterior. This integrated belief is used to plan over the remaining horizon. First, prior uncertainty enters both as the route's initial beliefs and also as the within-route continuation values that govern exploration and exploitation. Second, MetaDP serves as a rational benchmark. It is the policy a Bayesian consumer would adopt if they could integrate exactly over the hyper-posterior at every decision. We will use this benchmark in \S\ref{ssec:BRMDP_reordered} as the reference point against which to define a one-parameter family of boundedly rational approximations.

\paragraph{Route-level state.}
At flight \(t\) on route \(m\), the MetaDP state is
$s_{t,m}^{\mathrm{MetaDP}} = (\mathbf{N}_m^+,\mathbf{N}_m^-;Q_m)$, where \(\mathbf{N}_m^+ = (N_{1,m}^+,\ldots,N_{K,m}^+)\) and \(\mathbf{N}_m^- = (N_{1,m}^-,\ldots,N_{K,m}^-)\) are the within-route counts of on-time and delayed flights across airlines. As before, the flight index is pinned down by $
\|\mathbf{N}_m^+\|_1+\|\mathbf{N}_m^-\|_1=t-1$.

The hypothesis set \(\mathcal{P}=\{P^{(j)}\}_{j=1}^{J}\) is fixed across routes and is therefore treated as part of the model primitive rather than a changing state variable. Because cross-route learning occurs only after the route ends, the hyper-posterior \(Q_m\) remains fixed throughout route \(m\); within the route, only the count vectors evolve.

\paragraph{Hypothesis-conditioned action values.}
For each hypothesis \(P^{(j)}\in\mathcal{P}\), let
\[
\hat{\theta}_{k,t,m}^{(j)}
=
\frac{\alpha_k^{(j)}+N_{k,m}^{+}}{\alpha_k^{(j)}+\beta_k^{(j)}+N_{k,m}^{+}+N_{k,m}^{-}},
\]
denote the posterior mean on-time rate of airline \(k\) at flight \(t\) on route \(m\) under hypothesis \(j\). Conditional on \(P^{(j)}\), the action value from choosing airline \(k\) is
\begin{equation}
\label{eq:Ve_eps_category_reordered}
\begin{aligned}
V^{(j)}_{\epsilon}\!\bigl(k \mid \mathbf{N}_m^+,\mathbf{N}_m^-,Q_m;\mathcal{P}\bigr)
&= \hat{\theta}^{(j)}_{k,t,m} \\
&\quad + \gamma \Bigl[
\hat{\theta}^{(j)}_{k,t,m}\,
\bar{W}_{\epsilon}\!\bigl(\mathbf{N}_m^{+}+\mathbf{e}_k,\mathbf{N}_m^{-},Q_m;\mathcal{P}\bigr) \\
&\qquad + \bigl(1-\hat{\theta}^{(j)}_{k,t,m}\bigr)\,
\bar{W}_{\epsilon}\!\bigl(\mathbf{N}_m^{+},\mathbf{N}_m^{-}+\mathbf{e}_k,Q_m;\mathcal{P}\bigr)
\Bigr].
\end{aligned}
\end{equation}
The immediate payoff depends on hypothesis \(j\) through \(\hat{\theta}^{(j)}_{k,t,m}\). The continuation term, however, uses the integrated state value \(\bar{W}_{\epsilon}\), because future choices are made using the full hyper-posterior \(Q_m\) rather than any single hypothesis.

\paragraph{Integrated action and state values.}
Because \(Q_m\) is a discrete distribution over hypotheses and remains fixed during route \(m\), the consumer evaluates each airline by averaging the hypothesis-conditioned action values using the weights \(Q_m(j)\):
\begin{equation}
\label{eq:Vbar_meta_reordered}
\bar{V}_{\epsilon}\!\bigl(k \mid \mathbf{N}_m^+,\mathbf{N}_m^-,Q_m;\mathcal{P}\bigr)
=
\sum_{j=1}^{J} Q_m(j)\,
V^{(j)}_{\epsilon}\!\bigl(k \mid \mathbf{N}_m^+,\mathbf{N}_m^-,Q_m;\mathcal{P}\bigr).
\end{equation}
This integrated action value is the defining feature of MetaDP. Unlike a procedure that would sample a single prior hypothesis and then optimize conditionally on that draw, MetaDP carries prior uncertainty directly into the Bellman recursion by averaging across all hypotheses at every state.

At the terminal flight, there is no continuation value, so Equation~\eqref{eq:Vbar_meta_reordered} reduces to
\begin{equation}
\label{eq:Vbar_meta_T_reordered}
\bar{V}_{\epsilon}\!\bigl(k \mid \mathbf{N}_m^+,\mathbf{N}_m^-,Q_m;\mathcal{P}\bigr)
=
\sum_{j=1}^{J} Q_m(j)\,\hat{\theta}^{(j)}_{k,T,m},
\qquad
\|\mathbf{N}_m^+\|_1+\|\mathbf{N}_m^-\|_1 = T-1.
\end{equation}
Applying the \(\epsilon\)-greedy rule to these integrated action values gives the policy-induced state value:
\begin{equation}
\label{eq:Wbar_meta_reordered}
\bar{W}_{\epsilon}\!\bigl(\mathbf{N}_m^+,\mathbf{N}_m^-,Q_m;\mathcal{P}\bigr)
=
(1-\epsilon)\,\max_{k}\,
\bar{V}_{\epsilon}\!\bigl(k \mid \mathbf{N}_m^+,\mathbf{N}_m^-,Q_m;\mathcal{P}\bigr)
+
\frac{\epsilon}{K}\sum_{k=1}^{K}
\bar{V}_{\epsilon}\!\bigl(k \mid \mathbf{N}_m^+,\mathbf{N}_m^-,Q_m;\mathcal{P}\bigr).
\end{equation}
The policy assigns total probability \(1-\epsilon\) to the airlines with the highest integrated action value, while it distributes the choice-noise probability \(\epsilon\) uniformly across all airlines. Setting \(\epsilon=0\) yields the deterministic MetaDP benchmark. Equations~\eqref{eq:Ve_eps_category_reordered}--\eqref{eq:Wbar_meta_reordered} are solved by backward induction over \(t=T,T-1,\ldots,1\) within route \(m\).

\paragraph{Policy execution across routes.}
At each within-route state, the consumer chooses an airline using the \(\epsilon\)-greedy rule applied to the integrated action values:
\begin{equation}
\label{eq:mdp_policy_reordered}
\tilde{k}
\sim
\pi^{MetaDP}_{\epsilon}\!\bigl(\cdot \mid \mathbf{N}_m^+,\mathbf{N}_m^-,Q_m;\mathcal{P}\bigr),
\end{equation}
where
\begin{equation}
\label{eq:e_greedy_mixed_probability_func_reordered}
\pi^{MetaDP}_{\epsilon}\!\bigl(\tilde{k} \mid \mathbf{N}_m^+,\mathbf{N}_m^-,Q_m;\mathcal{P}\bigr)
=
(1-\epsilon)\,
\frac{
\mathbf{1}\!\Bigl\{
\tilde{k}\in \arg\max\limits_{k}
\bar{V}_{\epsilon}\!\bigl(k \mid \mathbf{N}_m^+,\mathbf{N}_m^-,Q_m;\mathcal{P}\bigr)
\Bigr\}
}{
\left|
\arg\max\limits_{k}
\bar{V}_{\epsilon}\!\bigl(k \mid \mathbf{N}_m^+,\mathbf{N}_m^-,Q_m;\mathcal{P}\bigr)
\right|
}
+
\frac{\epsilon}{K}.
\end{equation}
Thus, within-route learning updates the counts \((\mathbf{N}_m^+,\mathbf{N}_m^-)\), while cross-route learning updates the hyper-posterior only after the route is completed. After finishing route \(m\), the consumer uses the realized route-level counts to update \(Q_m\) to \(Q_{m+1}\) via Algorithm~\ref{alg:cross_route_update_reordered}, and then solves the next route using the updated hyper-posterior. See Web Appendix~\ref{web-ssec:metadp_algo} for the MetaDP pseudocode.

\paragraph{Interpretation.} MetaDP is our benchmark for fully integrated meta-learning. The consumer updates a hyper-posterior across routes and carries that uncertainty into the within-route Bellman recursion. As a result, experience from earlier routes affects later behavior both through the beliefs with which a new route begins and through the continuation values that shape forward-looking exploration and exploitation within that route. This full integration is computationally demanding. Solving MetaDP requires the consumer to maintain and update the hyper-posterior over prior hypotheses and to propagate that uncertainty through a finite-horizon dynamic programming problem within each route. This computational burden motivates the next model, BRMDP, which preserves the logic of cross-route learning while approximating the integration step.

\subsection{Boundedly Rational Meta Dynamic Programming (BRMDP)}
\label{ssec:BRMDP_reordered}

We now introduce Boundedly Rational Meta Dynamic Programming (BRMDP), a tractable approximation to MetaDP. BRMDP simplifies the representation of prior uncertainty within the route-level dynamic program using an approximation of the hyper-prior. Instead of integrating exactly over the full hyper-posterior \(Q_m\), the consumer represents that distribution using a finite number of draws. This formulation is motivated by two converging lines of cognitive-science evidence. First, in single-context judgment and choice tasks, people behave as if they use only a small number of samples (often just a single one) from the relevant posterior rather than the full distribution \citep{vul2014one}. Second, recent work on human planning shows that people adapt the representational and computational effort they devote to forward planning to the demands of the task. For example, during decision-making, people prune deep search trees, simplify state representations, and select planning depths \citep{callaway2022rational, ho2023rational, eluchans2025adaptive}. In both cases, departures from full Bayesian or DP-optimal computation reflect a cost--benefit trade-off in cognitive resources.


Drawing on these ideas, BRMDP generates a family of policies indexed by the number of hyper-posterior draws, \(D\). Intuitively, \(D\) governs how well the consumer integrates uncertainty over prior hypotheses across routes. Smaller values of \(D\) reduce computational cost but move behavior farther from MetaDP across routes. Larger values of \(D\) yield a closer approximation to MetaDP, but at a higher computational cost. One clear divergence from the prior cognitive work is that the bounded-resource approximation operates at the \emph{meta}-Bayesian level, i.e., over the hyper-posterior rather than the within-task posterior. The empirical question we take up in \S\ref{sec:empirical_likelihood_comparisons} is which value of \(D\) best describes the way participants actually integrate prior uncertainty across routes.
%
%
Table~\ref{tab:notation_brmdp_reordered} summarizes the BRMDP-specific notation introduced below.

\begin{table}[htp!]
\centering
\small
\caption{Notation for BRMDP}
\label{tab:notation_brmdp_reordered}
\begin{tabular}{p{.24\textwidth} p{.68\textwidth}}
\toprule
\textbf{Notation} & \textbf{Definition} \\
\midrule
$\mathbf{x}_m=(x_{1,m},\ldots,x_{J,m})$ & Vector of sampled hypothesis counts before route $m$ in BRMDP \\
$D$ & Number of hyper-posterior draws used in BRMDP \\
$w_j(\mathbf{x}_m)$ & BRMDP weight on hypothesis $j$ at route $m$, equal to $x_{j,m}/D$ \\
$s_{t,m}^{\mathrm{BRMDP}}$ & State in BRMDP at flight $t$ on route $m$: $(\mathbf{N}_{m}^{+},\mathbf{N}_{m}^{-};\mathbf{x}_m)$ \\
$V_{\epsilon}^{(j)}(k \mid \cdot)$ & BRMDP action value under hypothesis $P^{(j)}$ and sampled mixture $\mathbf{x}_m$ \\
$\tilde{V}_{\epsilon}(k \mid \cdot)$ & BRMDP integrated action value under sampled mixture $\mathbf{x}_m$ \\
$\tilde{W}_{\epsilon}(\cdot)$ & BRMDP state value under sampled mixture $\mathbf{x}_m$ \\
$\pi_{\epsilon,D}^{\mathrm{BRMDP}}(\cdot \mid \cdot)$ & BRMDP choice rule with policy noise $\epsilon$ and $D$ hyper-posterior draws \\
\bottomrule
\end{tabular}
\end{table}

\paragraph{Approximation through hyper-posterior draws.}
Under MetaDP, the consumer evaluates each action by integrating over all prior hypotheses using the full hyper-posterior \(Q_m\). Equivalently, one can view MetaDP as the limiting case of averaging over infinitely many draws from \(Q_m\). BRMDP replaces that exact integration with a finite-sample approximation. 

At the start of route \(m\), the consumer draws \(D<\infty\) hypothesis indices independently with replacement from \(Q_m\). This yields the count vector
\[
\mathbf{x}_m=(x_{1,m},\ldots,x_{J,m})\sim\mathrm{Multinomial}(D,Q_m),
\]
where \(x_{j,m}\) is the number of times hypothesis \(j\) appears among the \(D\) draws. These counts define route-specific mixture weights
\begin{equation}
\label{eq:weights_reordered}
w_j(\mathbf{x}_m)=\frac{x_{j,m}}{\|\mathbf{x}_m\|_1}=\frac{x_{j,m}}{D},
\qquad
\sum_{j=1}^{J} w_j(\mathbf{x}_m)=1.
\end{equation}

The vector \(\mathbf{x}_m\) therefore provides a sampled approximation to the hyper-posterior. Rather than evaluating each hypothesis in proportion to its exact probability \(Q_m(j)\), the consumer evaluates hypotheses in proportion to their empirical frequencies in the \(D\) draws. When \(D<J\), some hypotheses may receive zero weight, which reduces computation by allowing the consumer to ignore sufficiently unlikely possibilities. As \(D\to\infty\), the empirical weights converge almost surely to the exact hyper-posterior weights, \(w_j(\mathbf{x}_m)\to Q_m(j)\), and the BRMDP mixture converges to the fully integrated MetaDP mixture.

Using these sampled weights, we denote the BRMDP state during route \(m\) as
$s_{t,m}^{\mathrm{BRMDP}} =(\mathbf{N}_m^+, \mathbf{N}_m^-; \mathbf{x}_m)$, where \(\mathbf{x}_m\) is fixed throughout route \(m\).

\paragraph{Hypothesis-conditioned action values under sampled integration.}
Conditional on hypothesis \(P^{(j)}\), the action value from choosing airline \(k\) is
\begin{equation}
\label{eq:Ve_resource_category_reordered}
\begin{aligned}
V^{(j)}_{\epsilon}\!\bigl(k \mid \mathbf{N}_m^+,\mathbf{N}_m^-,\mathbf{x}_m;\mathcal{P}\bigr)
&= \hat{\theta}^{(j)}_{k,t,m} \\
&\quad + \gamma \Bigl[
\hat{\theta}^{(j)}_{k,t,m}\,
\tilde{W}_{\epsilon}\!\bigl(\mathbf{N}_m^{+}+\mathbf{e}_k,\mathbf{N}_m^{-},\mathbf{x}_m;\mathcal{P}\bigr) \\
&\qquad + \bigl(1-\hat{\theta}^{(j)}_{k,t,m}\bigr)\,
\tilde{W}_{\epsilon}\!\bigl(\mathbf{N}_m^{+},\mathbf{N}_m^{-}+\mathbf{e}_k,\mathbf{x}_m;\mathcal{P}\bigr)
\Bigr].
\end{aligned}
\end{equation}
This recursion is the same forward-looking Bellman logic used in MetaDP, except that future values are computed under the sampled mixture \(\mathbf{x}_m\) rather than the exact hyper-posterior \(Q_m\).

\paragraph{Integrated action and state values under sampled integration.}
Replacing \(Q_m(j)\) in Equation~\eqref{eq:Vbar_meta_reordered} with \(w_j(\mathbf{x}_m)\) gives the sampled integrated action value:
\begin{equation}
\label{eq:Vbar_meta_resource_reordered}
\tilde{V}_{\epsilon}\!\bigl(k \mid \mathbf{N}_m^+,\mathbf{N}_m^-,\mathbf{x}_m;\mathcal{P}\bigr)
=
\sum_{j=1}^{J} w_j(\mathbf{x}_m)\,
V^{(j)}_{\epsilon}\!\bigl(k \mid \mathbf{N}_m^+,\mathbf{N}_m^-,\mathbf{x}_m;\mathcal{P}\bigr).
\end{equation}
At the terminal flight, there is no continuation value, so
\begin{equation}
\label{eq:Vbar_meta_resource_T_reordered}
\tilde{V}_{\epsilon}\!\bigl(k \mid \mathbf{N}_m^+,\mathbf{N}_m^-,\mathbf{x}_m;\mathcal{P}\bigr)
=
\sum_{j=1}^{J} w_j(\mathbf{x}_m)\,\hat{\theta}^{(j)}_{k,T,m},
\qquad
\|\mathbf{N}_m^+\|_1+\|\mathbf{N}_m^-\|_1 = T-1.
\end{equation}
Applying the \(\epsilon\)-greedy rule to these sampled integrated action values gives the policy-induced state value:
\begin{equation}
\label{eq:Wbar_meta_resource_reordered}
\tilde{W}_{\epsilon}\!\bigl(\mathbf{N}_m^+,\mathbf{N}_m^-,\mathbf{x}_m;\mathcal{P}\bigr)
=
(1-\epsilon)\,\max_{k}\,
\tilde{V}_{\epsilon}\!\bigl(k \mid \mathbf{N}_m^+,\mathbf{N}_m^-,\mathbf{x}_m;\mathcal{P}\bigr)
+
\frac{\epsilon}{K}\sum_{k=1}^{K}
\tilde{V}_{\epsilon}\!\bigl(k \mid \mathbf{N}_m^+,\mathbf{N}_m^-,\mathbf{x}_m;\mathcal{P}\bigr).
\end{equation}
Equations~\eqref{eq:Ve_resource_category_reordered}--\eqref{eq:Wbar_meta_resource_reordered} are solved by backward induction over \(t=T,T-1,\ldots,1\) within route \(m\).

\paragraph{BRMDP choice rule.}
Action selection applies the same \(\epsilon\)-greedy logic as in MetaDP, but now to the sampled integrated action values \(\tilde{V}_{\epsilon}\). Thus, the BRMDP policy is
\begin{equation}
\label{eq:BRMDP_probability_func_reordered}
\pi_{\epsilon,D}^{\mathrm{BRMDP}}\!\bigl(\tilde{k} \mid \mathbf{N}_m^+,\mathbf{N}_m^-,\mathbf{x}_m;\mathcal{P}\bigr)
=
(1-\epsilon)\,
\frac{
\mathbf{1}\!\Bigl\{
\tilde{k}\in\arg\max\limits_{k}
\tilde{V}_{\epsilon}\!\bigl(k \mid \mathbf{N}_m^+,\mathbf{N}_m^-,\mathbf{x}_m;\mathcal{P}\bigr)
\Bigr\}
}{
\left|
\arg\max\limits_{k}
\tilde{V}_{\epsilon}\!\bigl(k \mid \mathbf{N}_m^+,\mathbf{N}_m^-,\mathbf{x}_m;\mathcal{P}\bigr)
\right|
}
+
\frac{\epsilon}{K}.
\end{equation}

Equations~\eqref{eq:Ve_resource_category_reordered}--\eqref{eq:BRMDP_probability_func_reordered} define the BRMDP policy. See Web Appendix~\ref{web-ssec:brmdp_algo} for the BRMDP pseudocode.

\paragraph{Interpretation and limiting cases.}
BRMDP provides a tractable way to model a boundedly rational consumer's inherent reward--computation trade-off. The parameter \(D\) governs how accurately the consumer integrates cross-route prior uncertainty. As \(D\to\infty\), the empirical weights converge to the exact hyper-posterior weights, and BRMDP converges to MetaDP. At the other extreme, small values of \(D\) produce simpler and more approximate policies in which the consumer plans using a coarse representation of prior uncertainty. In this sense, BRMDP nests a family of boundedly rational meta-learning policies between the fully rational MetaDP benchmark and simpler heuristics.

Taken together, the three policy classes span the main behavioral possibilities of interest in this paper. The baseline DP policy captures a consumer who learns only within a route and treats each new route as a fresh problem. MetaDP captures a fully integrated meta-learner who transfers information across routes and incorporates that information into full-horizon dynamic planning. BRMDP spans intermediate cases in which the consumer also transfers information across routes, but does so approximately, using a limited number of hypothesis draws. 

\section{Empirical Comparison of the Policies}
\label{sec:empirical_likelihood_comparisons}
We now compare the behavior of humans in our experiment with that of the three classes of algorithms discussed above. We first present descriptive comparisons from the data and simulations in $\S$\ref{ssec:initial_qual_results} and then present likelihood-based estimation in $\S$\ref{ssec:dgp_identification_reordered}.

\subsection{Qualitative Comparison of Human and Simulated Policy Dynamics}
\label{ssec:initial_qual_results}

\begin{figure}[htp!]
  \includegraphics[width=1\textwidth]{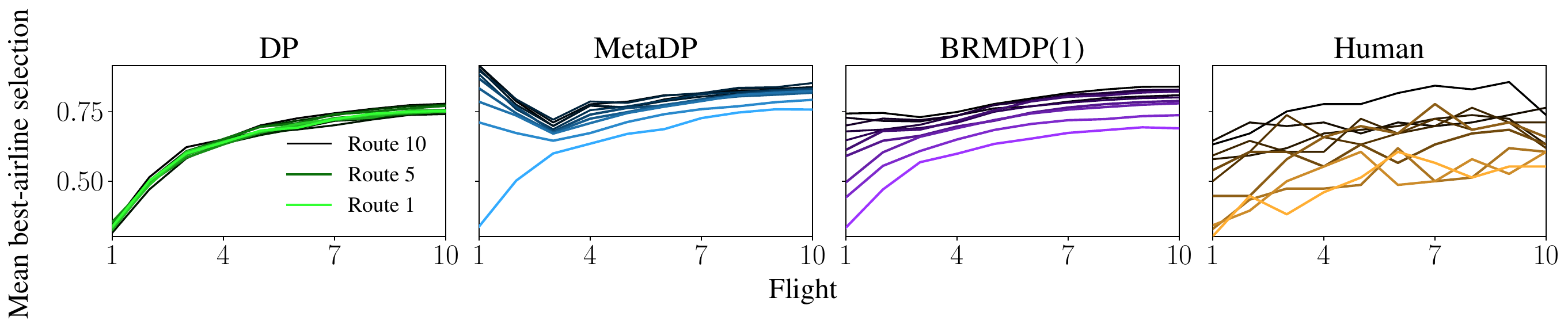}
    \caption{Mean best-airline selection over flights for different routes in the Far Means--Low Var.\ condition. BRMDP(1) more closely resembles participant behavior. Policy curves are averages over $2{,}000$ simulated runs.}
    \label{fig:3_qualitative_best_arm_FarMeans}
\end{figure}

\begin{figure}[htp!]
  \includegraphics[width=1\textwidth]{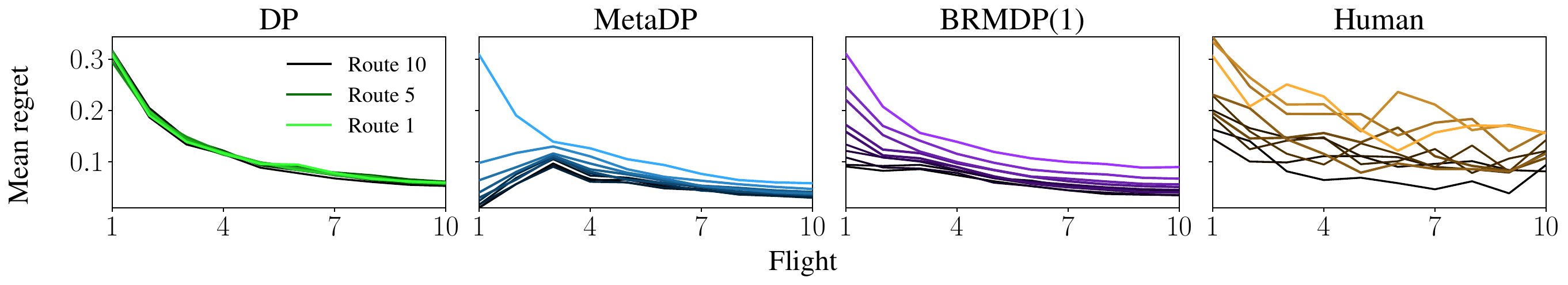}
    \caption{Mean pseudo-regret over flights for different routes in the Far Means--Low Var.\ condition. BRMDP(1) represents behavior closer to participants. Policy curves are averages over $2{,}000$ simulated runs.}
    \label{fig:3_qualitative_regrets_FarMeans}
\end{figure}

Before turning to formal likelihood comparisons, we begin with a descriptive comparison between human behavior and simulated policy paths. Figures~\ref{fig:3_qualitative_best_arm_FarMeans} and ~\ref{fig:3_qualitative_regrets_FarMeans} plot mean best-airline selection and mean pseudo-regret over flights and routes for the Far Means--Low Var.\ condition (see Web Appendix~\ref{web-ssec:qual_human_vs_algo_comparison} for the corresponding plots for the other experimental conditions). For the policy curves, we simulate each policy $2{,}000$ times in the same calibrated environment and compute the same route-by-flight performance measures used for the human data. The figures, therefore, provide a common descriptive benchmark for comparing the qualitative dynamics of humans and policies, rather than a formal goodness-of-fit exercise. As an initial comparison, we set $\epsilon = 0$ for all policies to obtain their noise-free performance. Later, we explain how we include $\epsilon$ in our likelihood-based analysis.

For the baseline DP policy, we use uninformative priors, \((\boldsymbol{\alpha},\boldsymbol{\beta})=(\mathbf{1},\mathbf{1})\). For MetaDP and BRMDP, we construct the hypothesis class from the experimental mean grid \(\{.2,.4,.5,.6,.8\}\), which yields \(5^3=125\) candidate prior hypotheses for the three airlines. We then map each candidate mean \(\mu_k^{(j)}\) into a Beta prior by setting \(\alpha_k^{(j)}=\mu_k^{(j)}\) and \(\beta_k^{(j)}=1-\mu_k^{(j)}\). As an illustrative member of the boundedly rational policy class, we plot BRMDP(1), where BRMDP(1) denotes a policy with one hyper-posterior draw (\(D=1\)).

The figures suggest that the qualitative trajectory of human behavior is closer to BRMDP(1) than to either DP or MetaDP. Because DP does not transfer information across routes, its best-airline-selection and pseudo-regret profiles are nearly the same from one route to the next. MetaDP lies at the opposite extreme: it improves sharply at the beginning of later routes, reflecting aggressive and fully integrated cross-route learning. Human behavior does not closely match either pattern.

By contrast, BRMDP(1) produces trajectories that more closely resemble the human data. In both cases, performance at the beginning of a route improves gradually across routes: best-airline selection starts lower and rises over flights, while pseudo-regret starts higher and declines, with later routes beginning from better baselines than earlier ones. Thus, this descriptive comparison already points to an intermediate form of cross-route learning---stronger than DP, but less sharp than fully integrated MetaDP. The likelihood-based analysis that follows formalizes this comparison and evaluates that conclusion more rigorously.

\subsection{Likelihood-Based Model Comparison}
\label{ssec:dgp_identification_reordered}

Having introduced the three policy classes---DP, MetaDP, and BRMDP---we now compare them to the experimental data. The central question is straightforward: which of these policies most closely matches the sequence of choices participants actually make? To answer this question, we use likelihood-based comparison to evaluate competing hypotheses about the decision-making policies consumers adopt \citep{kass1995bayes}. Following the trial-by-trial approach of \citet{daw2011trial}, we evaluate each model on the full observed decision path rather than on aggregate outcomes alone, and we extend that logic to our multi-route, hierarchical setting.

This framework is especially useful in our context because the candidate policies differ not only in how they learn \emph{within} a route, but also in whether and how they transfer information \emph{across} routes. As a result, the likelihood comparison speaks directly to the paper's main substantive question: do consumers behave like no-transfer learners, fully rational meta-learners, or boundedly rational meta-learners?

Throughout, we condition on the realized flight outcomes when evaluating choice probabilities. This is standard in dynamic choice applications: the policy determines the probability of each action, while realized outcomes affect future choices only through belief updating. Accordingly, a model fits the data well if, after observing the same history that the participant observed, it assigns high probability to the participant's next choice. See Web Appendix~\ref{web-sec:likelihood_estimation_details} for a detailed explanation of the likelihood estimation procedure.

\paragraph{Estimation Results.}

We now turn from likelihood construction to likelihood comparison. Throughout this subsection, the baseline DP policy uses uninformative priors, \((\boldsymbol{\alpha},\boldsymbol{\beta})=(\mathbf{1},\mathbf{1})\). For MetaDP and BRMDP, we construct the hypothesis class from the discrete mean grid used in the experiments,$\{.2,.4,.5,.6,.8\}$.
Taking all ordered triples with replacement yields \(5^3=125\) candidate prior hypotheses for the three airlines.\footnote{In this calibration, the true environment's prior means lie in the hypothesis class. However, participants are not aware of these means, so their prior means could be misspecified. In Web Appendix~\ref{web-ssec:prior_misspecification}, we relax this assumption and show that the likelihood estimations for a case with misspecified prior means are directionally the same as the main results.} We then map each candidate mean \(\mu_k^{(j)}\) into a Beta prior as follows: $\alpha_k^{(j)} = \mu_k^{(j)}$ and $\beta_k^{(j)} = (1-\mu_k^{(j)})$.

We also set the discount factor to \(\gamma=1\). This is a practical and economically sensible choice in our short-horizon task, where rewards are realized at the end of the experiment, and discounting is therefore less salient \citep{liu2020understanding}. More importantly, \(\gamma\) is not separately identified in our setting without an instrument that shifts continuation values while leaving current payoffs unchanged \citep{rust1987optimal, MagnacThesmar2002}. Accordingly, we treat \(\gamma=1\) as a maintained assumption throughout.

For each policy and experimental condition, we aggregate participant-level log-likelihoods. A higher log-likelihood means the policy assigns a higher probability to the observed sequence of choices, conditioned on the same histories participants observed. This comparison is informative because the three policy families embody different behavioral mechanisms. DP captures within-route learning without cross-route transfer. MetaDP captures fully integrated cross-route learning. BRMDP spans an intermediate class in which consumers transfer knowledge across routes, but approximately, through a limited number of prior draws.

\begin{figure}[htp!]
\centering
\includegraphics[width=.6\textwidth]{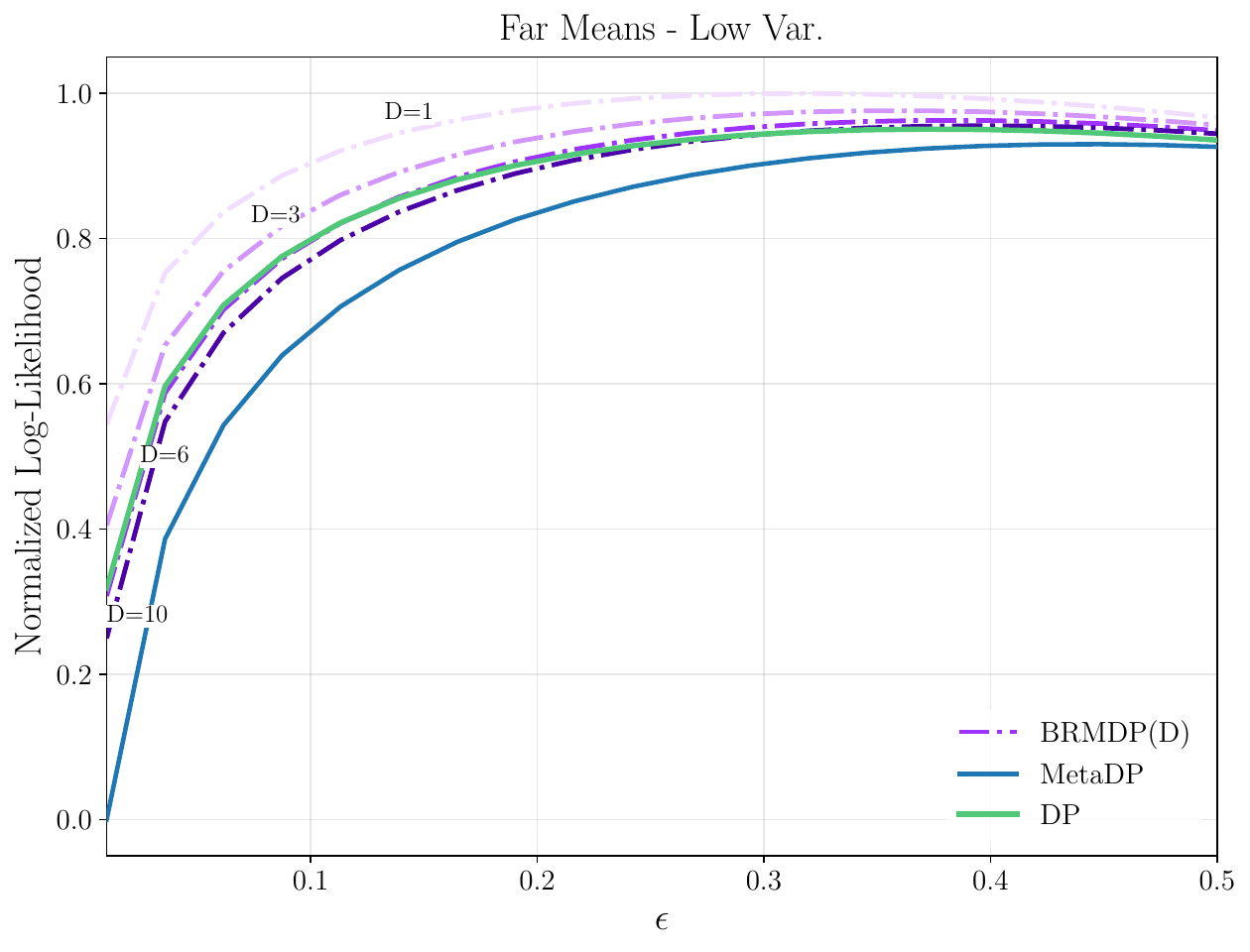}
\caption{Log-likelihood comparison across policies for \(20\) values of \(\epsilon\). Monte Carlo likelihood calculations use \(B=2{,}000\) samples. We normalized the log-likelihood values via $(l_. - l_{min})/(l_{max} - l_{min})$ for better visualizations and unified plots. BRMDP(1) represents the highest values, indicating a better fit.}
\label{fig:dp_meta_log_like_reordered}
\end{figure}

We evaluate the total log-likelihood over a grid of \(20\) noise values, spanning \(\epsilon\in[.01,.5]\). This allows us to compare policies without tying the conclusion to one particular choice of noise parameter. We vary \(D\) over \(\{1,3,6,10\}\). Figure~\ref{fig:dp_meta_log_like_reordered} and Table~\ref{tab:ll_by_algorithm_epsilon_last10} report the resulting fit profiles for DP, MetaDP, and BRMDP for the Far Means--Low Variance experiment. See Web Appendix~\ref{web-ssec:all_experiments_loglike} for the results for the other three conditions.  Here, we assume the policies solve the dynamic programming problem, taking the full ten flights of each route into account. Nevertheless, we present similar results for shorter planning horizons in Web Appendix~\ref{web-ssec:planning_horizon_robustness}. 

\begin{table}[!htbp]
\centering
\caption{Log-likelihood by policy and $\varepsilon$}
\label{tab:ll_by_algorithm_epsilon_last10}
\small
\setlength{\tabcolsep}{4pt}
\renewcommand{\arraystretch}{1.1}
\resizebox{\textwidth}{!}{%
\begin{tabular}{lcccccccccc}
\toprule
Policy & $.268$ & $.294$ & $.319$ & $.345$ & $.371$ & $.397$ & $.423$ & $.448$ & $.474$ & $.500$ \\
\midrule
DP & -6161.5 & -6116.6 & -6086.3 & -6068.6 & \textbf{-6062.2} & -6066.0 & -6079.1 & -6100.8 & -6130.6 & -6168.1 \\ \hline
BRMDP(1) & -5737.1 & -5721.2 & \textbf{-5717.6} & -5724.9 & -5741.9 & -5768.0 & -5802.2 & -5844.2 & -5893.5 & -5949.8 \\
BRMDP(3) & -5884.9 & -5857.4 & -5846.3 & \textbf{-5844.5} & -5849.3 & -5861.0 & -5880.0 & -5942.4 & -5978.7 & -6022.9 \\
BRMDP(6) & -6078.9 & -6024.3 & -5988.4 & \textbf{-5970.1} & -5978.1 & -6006.2 & -5991.6 & -6009.2 & -6037.9 & -6074.8 \\
BRMDP(10) & -6181.9 & -6122.4 & -6079.2 & -6050.1 & -6033.8 & \textbf{-6028.8} & -6034.3 & -6049.5 & -6073.8 & -6106.7 \\
MetaDP & -6506.1 & -6415.5 & -6344.3 & -6290.0 & -6250.5 & -6224.5 & -6210.6 & \textbf{-6207.9} & -6215.6 & -6233.1 \\
\bottomrule
\end{tabular}
}
\vspace{.5em}

\begin{minipage}{.95\textwidth}
\footnotesize
\textit{Notes:} Only the last 10 (largest) $\varepsilon$ values are shown. Boldfaced values represent the largest log-likelihood within each policy row.
\end{minipage}
\end{table}

There are two main takeaways from this analysis. First, for a range of smaller \(D\) values, the BRMDP policy has a higher log-likelihood than the baseline DP benchmark. \footnote{Since the finite-simulation Jensen bias is downward, the exact BRMDP log-likelihoods would be higher in expectation than the reported simulated values. Thus, the strong fit of $\mathrm{BRMDP}(1)$ is conservative.
} This superiority shrinks and eventually reverses as \(D\) becomes larger, and the policy moves closer to the fully integrated MetaDP benchmark. This pattern suggests that participants do transfer information across routes, but they do so in a boundedly rational way rather than as fully optimal meta-learners. Second, for larger values of \(D\), and consequently more integrated meta-learning, the maximum log-likelihood tends to occur at larger noise values. Under the \(\epsilon\)-greedy specification, a larger noise parameter means that the policy places less weight on the value-maximizing airline and more weight on uniform exploration across all airlines. As a result, choices become less tightly tied to the policy's computed action values and therefore more stochastic. This indicates that as the policy becomes more optimal, it becomes more distant from observed human behavior unless paired with substantial decision noise.

The likelihood results provide further support for the paper's substantive conclusion. They reject both behavioral extremes. On the one hand, participants are not pure within-route learners: DP fits systematically worse than a range of bounded meta-learning models. On the other hand, participants are not fully optimal meta-planners: the highest-fitting models use relatively small \(D\), not the full MetaDP benchmark. The data, therefore, point to boundedly rational meta-learning: among the policies we consider, consumers generalize from prior routes using simplified representations of prior uncertainty. This aggregate result also holds broadly across participants. In Web Appendix~\ref{web-ssec:out_of_sample_predictive_validation}, participant-level out-of-sample comparisons show that, in every condition, the median predictive-likelihood share favors BRMDP(1) over both DP and MetaDP, although there remains meaningful heterogeneity across participants.

This pattern aligns with theories from cognitive science that view human behavior as resource-rational rather than unbounded \citep{simon1955behavioral, lieder2020resource, bhui2021resource, gershman2018deconstructing, griffiths2019doing}. Among these theories are those that show evidence for approximate Bayesian inference done by drawing a small number of posterior samples. Under these models, one- or few-sample decision-making has been shown to account for behavior in single-context judgment and choice tasks \citep{vul2014one,  chater2020probabilistic}. Our results extend this picture. First, in our model, the logic of approximation through sampling operates at the \emph{meta-level} of prior uncertainty across tasks, not only at the within-task level: BRMDP$(1)$, which represents the cross-route hyper-posterior with a single draw, fits human choice data better than BRMDP$(D>1)$ and MetaDP variants. Second, our findings emerge in a setting that requires forward-looking planning, not one-shot judgment. This suggests that coarse internal sampling may control how consumers integrate prior uncertainty into sequential, finite-horizon decisions. These new findings complement recent work that examines human cognition as the output of a meta-learning process over distributions of related tasks \citep{binz2024meta, lai2024human}. Once such a meta-level prior is in place, our model characterizes how a boundedly rational agent can use it for efficient decision-making.

\subsection{Robustness Checks and Extensions}
\label{ssec:robustness_and_extensions}

We subject both the reduced-form evidence of knowledge transfer and the policy-comparison results to a series of robustness checks. These analyses address several alternative explanations for our main findings: that the behavioral evidence is an artifact of researcher-observed outcome measures; that BRMDP(1)'s advantage is specific to one experimental environment or prior specification; that it reflects assumptions about planning horizon or stochastic choice; that cross-route improvement is driven by task familiarity rather than knowledge transfer; or that any simple form of cross-route transfer would fit equally well. We also evaluate whether BRMDP(1)'s advantage persists out of sample and across participants. Across these exercises, the main conclusion remains unchanged. We summarize the results below and provide the full analyses in Web Appendix~\ref{web-ssec:checks_extensions}.

\squishlist

\item \textbf{Choice-path evidence of knowledge transfer.}
Our main reduced-form measures in $\S$\ref{sec:behavioral_evidence_meta_learning} (best-airline selection and pseudo-regret) use the true route-specific on-time rates, which participants do not observe. Web Appendix~\ref{web-ssec:choice_path_transfer} therefore examines transfer using only participants' own observed choices and outcomes. Before each new route, we construct airline-specific posterior means from either all earlier routes or only the immediately preceding route and relate these beliefs to the participant's first choice on the new route. Under both history definitions, participants are significantly more likely to begin a new route with an airline whose posterior mean is higher. Thus, the evidence for cross-route knowledge transfer is visible directly in the choice path and does not rely on latent environmental parameters.

\item \textbf{Robustness across experimental environments.}
The main-text qualitative and likelihood comparisons focus on the Far Means--Low Var.\ condition, where airlines are relatively easy to distinguish, and their performance is comparatively stable across routes. In Web Appendices~\ref{web-ssec:qual_human_vs_algo_comparison} and~\ref{web-ssec:all_experiments_loglike}, we repeat the comparisons in all four experimental conditions. BRMDP(1) continues to outperform both DP and MetaDP in predicting participants' choices across these environments. Its relative advantage is therefore not confined to the setting most favorable to cross-route learning.

\item \textbf{Prior-mean misspecification.}
Our main specification constructs the candidate prior class from the pooled mean grid across the experimental conditions, so the true environmental means are contained in the hypothesis class. Web Appendix~\ref{web-ssec:prior_misspecification} deliberately misspecifies this class by swapping the mean sets across experiments: we use ${.2,.5,.8}$ when the true means are ${.4,.6,.8}$, and vice versa. BRMDP(1) retains the highest likelihood and the policy ordering remains directionally consistent with the main results. Thus, the advantage of coarse integration is not driven by including the true prior means in the candidate class.

\item \textbf{Planning-horizon robustness.}
The main analysis assumes that consumers plan over the full ten-flight route horizon. Because prior work suggests that shorter planning horizons can better describe human sequential choice \citep{meyer1995sequential,camerer2004cognitive}, in Web Appendix~\ref{web-ssec:planning_horizon_robustness} we re-solve all policies using rolling two- and five-flight planning horizons. BRMDP(1) remains the best-fitting policy under both specifications. The preference for coarse integration of higher-order uncertainty therefore persists even when consumers are allowed to be substantially more myopic within a route.

\item \textbf{Alternative stochastic choice rule.}
Our main specification uses an $\epsilon$-greedy rule to map action values into stochastic choices. Because the policy ranking could in principle reflect this mapping rather than the learning process itself, Web Appendix~\ref{web-ssec:softmax_spec} replaces $\epsilon$-greedy choice with a smooth softmax rule \citep{bhui2021resource}. BRMDP(1) remains the best-fitting policy. Its relative advantage therefore does not depend on the particular form of stochasticity used to translate action values into choice probabilities.

\item \textbf{Out-of-sample predictive validation.}
Next, we ask whether BRMDP(1)'s in-sample advantage generalizes to held-out choices. Web Appendix~\ref{web-ssec:out_of_sample_predictive_validation} conducts two validation exercises using the exact BRMDP(1) likelihood. First, we estimate the choice-noise parameter using all other participants in the same condition and predict the held-out participant. Second, we select the noise parameter using routes $1$--$5$ and evaluate predictive fit on routes $6$--$10$. In every experimental condition and in both exercises, the mean participant-level predictive log-likelihood difference favors BRMDP(1) over both DP and MetaDP, with 95\% participant-bootstrap confidence intervals entirely above zero. The median participant-level predictive-likelihood share also exceeds $.5$ against both benchmarks in every condition. Thus, BRMDP(1)'s advantage persists for held-out participants and future choices and reflects a broad pattern across participants rather than a small number of influential observations, although meaningful individual heterogeneity remains.

\item \textbf{No-transfer learning with improving execution.}
Cross-route improvement could arise even without knowledge transfer if participants simply become more familiar with the task and execute the same within-route strategy more consistently over time. To examine this possibility, in Web Appendix~\ref{web-ssec:improving_execution_robustness}, we introduce Dynamic Programming with Improving Execution (DPIE), a no-transfer DP policy whose choice noise is allowed to decline across routes. We estimate the declining noise profile using leave-one-participant-out validation within each experimental condition. BRMDP(1) achieves a higher mean predictive log-likelihood than DPIE in all four conditions. Thus, improvements in task execution alone do not account for the predictive advantage of cross-route learning.


\item \textbf{Non-meta-learning knowledge-transfer benchmarks.}
Finally, our main comparison focuses on meta-learning policies, but BRMDP(1) could fit well simply because it carries airline-specific information across routes rather than because it learns and represents higher-order uncertainty. To distinguish these explanations, in Web Appendix~\ref{web-ssec:non_meta_transfer_robustness}, we consider three knowledge-transfer policies that do not maintain a hyper-posterior. The first, Knowledge Transfer Mean-Updating (KTMU), pools observations from earlier and current routes and chooses myopically based on the resulting posterior means. This provides a simple form of knowledge transfer that effectively treats past and current experience as coming from the same context. The second, $\nu$-Knowledge Transfer Mean-Updating ($\nu$-KTMU), instead summarizes earlier experience as a transferred prior with a fixed effective sample size $\nu$. This allows past experience to shape beliefs on a new route without treating all earlier observations as if they were generated in the current context, while remaining myopic in its choices. Finally, $\nu$-Knowledge Transfer Dynamic Programming ($\nu$-KTDP) uses the same transferred prior but incorporates it into a finite-horizon dynamic program, allowing us to separate the role of forward-looking planning from the higher-order learning embodied in BRMDP. Across all four experimental conditions, BRMDP(1) has a higher likelihood than these non-meta-learning benchmarks throughout the displayed $\epsilon$ grid. Its advantage therefore cannot be explained simply by pooling past outcomes, transferring a smoothed brand-level mean across routes, or combining such transfer with forward-looking planning.

\squishend

\section{Managerial Implications and Takeaways}
\label{sec:managerial_implications}

The results above have implications not only for model fit, but also for the managerial counterfactuals one would draw from the model. In many repeated-choice settings, consumers do not appear to treat each new route, product category, or usage context as an entirely fresh problem. Instead, they carry forward brand-level beliefs from prior experience. This means that quality inferences can spill over across contexts, so that experiences in one setting may shape demand in another.

At the same time, our likelihood results indicate that this transfer is not best described by either extreme benchmark. Consumers do not behave as if they ignore cross-context information altogether, but neither do they appear to fully integrate all prior uncertainty in the manner implied by MetaDP. Rather, the best-fitting low-\(D\) BRMDP specifications suggest that consumers transfer knowledge using coarse summaries of prior experience. This distinction matters because pricing, promotion, trial, and launch decisions are often nonlinear in consumer beliefs. As a result, the wrong behavioral model can generate the wrong managerial recommendation even when it captures the broad direction of learning.

The results yield three main managerial takeaways. Most broadly, they imply that brand-level learning can carry across adjacent contexts and therefore affect downstream demand beyond the initial setting. They also imply that the behavioral model used for counterfactual analysis matters: a firm that assumes either no transfer or full integration may mischaracterize how consumers respond to new offerings. Finally, they suggest that when consumers behave more like BRMDP(1), effective managerial policies should be designed for a market composed of consumers with coarse transferred priors rather than for a single fully integrated representative consumer. We discuss these implications below. Detailed numerical illustrations are provided in Web Appendix~\ref{web-sec:web_managerial_counterfactuals}.

\squishlist

\item \textbf{Early experiences matter beyond the initial context.}
Because consumers appear to carry brand-level priors across routes, categories, or adjacent usage contexts, a firm's early performance can affect later demand even when the consumer encounters the brand in a new setting. This increases the long-run value of high-signal early experiences and increases the downstream cost of salient early failures. The pricing example in Web Appendix~\ref{web-ssec:web_pricing_promotion_counterfactuals} illustrates this point: after favorable prior experience, a no-transfer DP forecast can be too conservative because it ignores the demand created by transferable brand-level beliefs.

\item \textbf{The choice of learning model matters for managerial counterfactuals.}
The likelihood results in \S\ref{sec:empirical_likelihood_comparisons} favor BRMDP(1) over either DP or fully integrated MetaDP. Accordingly, a manager who forecasts demand using either a no-transfer model or a fully integrated meta-learning model may choose the wrong price, promotion, or trial policy even when the model captures the broad direction of learning. In the spirit of \citet{meyer2016we}, the issue is not only model fit; it is that the wrong behavioral model can imply the wrong policy. The counterfactuals in Web Appendix~\ref{web-ssec:web_pricing_promotion_counterfactuals} illustrate both directions of error: DP can underprice or overinvest in support because it ignores transfer, while MetaDP can miss profitable targeted interventions because it collapses heterogeneous coarse priors into a single representative posterior.

\item \textbf{Coarse integration implies mixture demand.}
When consumers behave more like BRMDP(1), the market looks less like a representative consumer with one integrated posterior and more like a mixture of consumers using coarse sampled priors. This matters whenever firm decisions are nonlinear in beliefs, as in pricing, promotions, launch trials, or assortment visibility. In such settings, DP can under-react because it ignores transfer, whereas MetaDP can misstate threshold responses because it averages away coarse heterogeneity in consumers' priors. The route-entry examples in Web Appendix~\ref{web-ssec:web_route_entry_counterfactuals} show the same logic for featured placement and trial credits.

\squishend

Taken together, these implications suggest that managers should treat early experiences as investments in transferable brand beliefs, but should avoid assuming that consumers integrate such beliefs fully and continuously. When behavior is closer to BRMDP(1), launch demand and intervention response are better viewed as mixture outcomes generated by coarse transferred priors. Managerial policies that recognize this mixture structure are likely to produce more reliable counterfactuals than policies that assume no meta-learning or those that assume a fully integrated representative-consumer benchmark.

\section{Conclusion}
\label{sec:conclusion}

Many everyday purchase sequences contain opportunities to carry knowledge across contexts. We study this form of learning-to-learn in a hierarchical Bernoulli bandit setting and compare human behavior with dynamic-programming policies that differ in how they use cross-context information. Our framework nests a no-transfer DP benchmark that treats contexts independently, a fully integrated MetaDP benchmark that carries cross-context beliefs into dynamic planning, and a boundedly rational \(\mathrm{BRMDP}(D)\) family that approximates MetaDP by limiting the integration of hyper-posterior uncertainty.

Using data from a carefully designed multi-context experiment, we first present clear evidence of knowledge transfer. Pseudo-regret declines across routes, best-airline selection improves across routes, and later routes begin from a better choice baseline, indicating that people do not treat each route in isolation but instead carry structure forward. Next, we use the set of policy classes we developed and trial-by-trial likelihood comparisons to show that a boundedly rational meta-learning policy best accounts for behavior. Low-\(D\) \(\mathrm{BRMDP}(D)\) specifications, especially \(\mathrm{BRMDP}(1)\), fit the human choice data better than both the no-transfer DP benchmark and the fully integrated MetaDP benchmark. Thus, the likelihood results reject both extremes: participants do not ignore cross-route information, but they also do not appear to integrate prior uncertainty as fully as MetaDP requires. As \(D\) increases and BRMDP approaches MetaDP, the fit to the data worsens, suggesting that the best account is not simply a noisier version of full Bayesian integration. The main substantive conclusion is that consumers transfer knowledge across contexts using meta-learning, but they do so approximately, using a coarse representation of prior uncertainty. The paper considers a spectrum of meta-learning policies and identifies the one that best describes human choices: a policy that learns and reuses higher-order beliefs about brand performance across contexts through a structured, forward-looking process, while relying on a computationally parsimonious representation of uncertainty.

Methodologically, the paper introduces and empirically tests two new model classes for consumer meta-learning in sequential choice. The first is MetaDP, a fully integrated Bayesian meta-learning model that combines cross-context belief updating with finite-horizon dynamic programming. The second is \(\mathrm{BRMDP}(D)\), a boundedly rational approximation that limits the integration of cross-context uncertainty through a finite number of hyper-posterior draws. We also develop a route-level Monte Carlo likelihood estimator for BRMDP, which integrates over consumers' latent internal draws and allows DP, MetaDP, and BRMDP to be compared on identical information sets and observed histories.

For managers, this implies that early, high-signal experiences that shape brand priors can yield downstream gains beyond the initial context, but the behavioral model used for counterfactual analysis also matters. A no-transfer model can understate the effect of favorable prior experience or lead firms to overinvest in support that consumers no longer need. A fully integrated MetaDP model can also mislead because it collapses coarse heterogeneity in consumers' transferred priors into a single representative posterior. When behavior is closer to \(\mathrm{BRMDP}(1)\), demand in a new context is better viewed as arising from a mixture of consumers with coarse transferred priors. This has direct implications for pricing, promotion, featured placement, and trial-credit decisions, especially because these policies are nonlinear in consumer beliefs.

\begin{singlespace}
\bibliographystyle{jmr}
\bibliography{ref}
\end{singlespace}

\clearpage
\setcounter{section}{0}
\setcounter{subsection}{0}
\setcounter{subsubsection}{0}
\setcounter{table}{0}
\setcounter{figure}{0}
\setcounter{equation}{0}
\setcounter{algorithm}{0}
\renewcommand{\thesection}{\Alph{section}}
\renewcommand{\thesubsection}{\thesection.\arabic{subsection}}
\renewcommand{\thesubsubsection}{\thesubsection.\arabic{subsubsection}}
\renewcommand{\thetable}{W\arabic{table}}
\renewcommand{\thefigure}{W\arabic{figure}}
\renewcommand{\theequation}{W\arabic{equation}}
\renewcommand{\thealgorithm}{W\arabic{algorithm}}
\renewcommand{\theHsection}{web.\Alph{section}}
\renewcommand{\theHsubsection}{web.\Alph{section}.\arabic{subsection}}
\renewcommand{\theHsubsubsection}{web.\Alph{section}.\arabic{subsection}.\arabic{subsubsection}}
\renewcommand{\theHtable}{web.\arabic{table}}
\renewcommand{\theHfigure}{web.\arabic{figure}}
\renewcommand{\theHequation}{web.\arabic{equation}}
\providecommand{\theHalgorithm}{}
\renewcommand{\theHalgorithm}{web.\arabic{algorithm}}
\titleformat{\section}
  {\normalfont\bfseries\centering}
  {Web Appendix \thesection:}
  {.5em}
  {}

\makeatletter
\let\JMRmainaddcontentsline\addcontentsline
\renewcommand{\addcontentsline}[3]{%
  \def\JMRcontentsfile{#1}%
  \def\JMRtocfile{toc}%
  \ifx\JMRcontentsfile\JMRtocfile
    \JMRmainaddcontentsline{watoc}{#2}{#3}%
  \else
    \JMRmainaddcontentsline{#1}{#2}{#3}%
  \fi
}
\newcommand{\JMRwebtableofcontents}{%
  \section*{\contentsname}%
  \@starttoc{watoc}%
}
\makeatother

\pagenumbering{arabic}
\setcounter{page}{1}
\pagestyle{plain}

\begin{center}
{\Large\bfseries Web Appendix\par}
\vspace{1em}
{\large\bfseries Boundedly Rational Meta-Learning in Sequential Consumer Choice\par}
\end{center}

\vspace{1em}

\vspace{1em}
\JMRwebtableofcontents
\clearpage
\section{Experimental Instructions}
\label{web-sec:experiments_instructions}
In this section, we provide the experimental instructions we give to participants.

\begin{figure}[htp!]
\centering
\includegraphics[width=.6\textwidth]{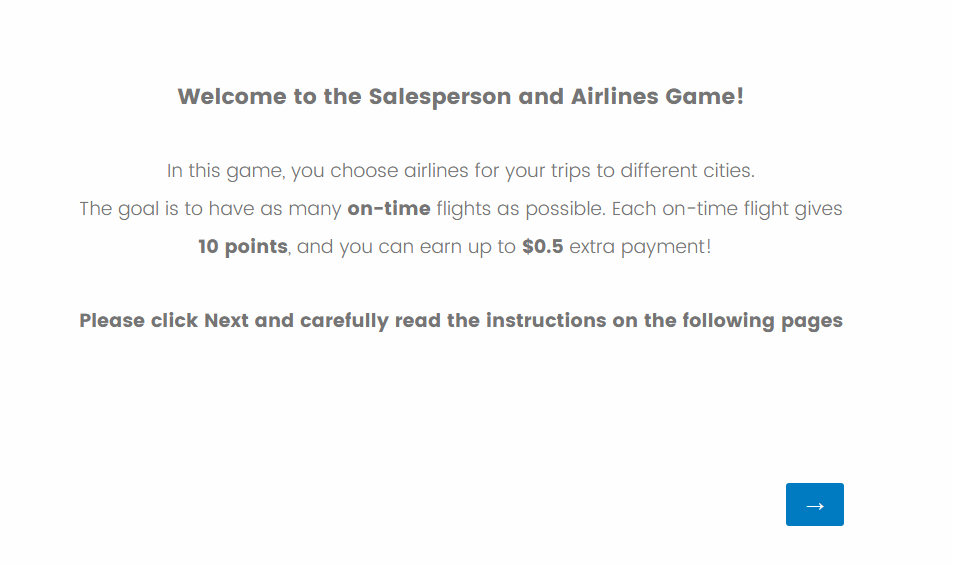}
\caption{Overview of the experimental task and participants' objective.}
\label{web-fig:experiment_overview_instructions}
\end{figure}

\begin{figure}[htp!]
\centering
\includegraphics[width=.6\textwidth]{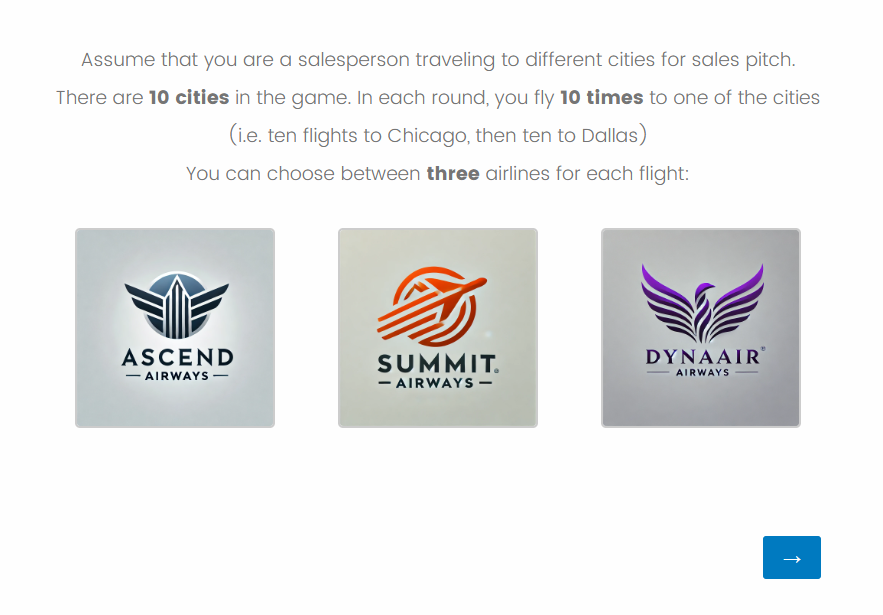}
\caption{Structure of the experimental task, including routes, flights, and airline options.}
\label{web-fig:game_structure_airlines_instructions}
\end{figure}

\begin{figure}[htp!]
\centering
\includegraphics[width=.6\textwidth]{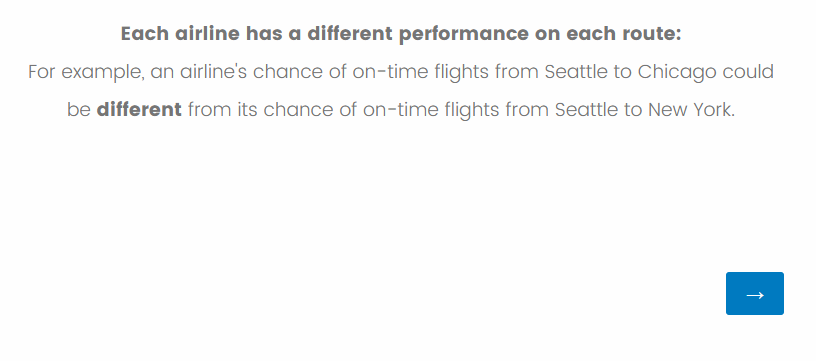}
\caption{Illustration that airline performance can vary across routes in the experimental environment.}
\label{web-fig:different_performances_instructions}
\end{figure}

\begin{figure}[htp!]
\centering
\includegraphics[width=.6\textwidth]{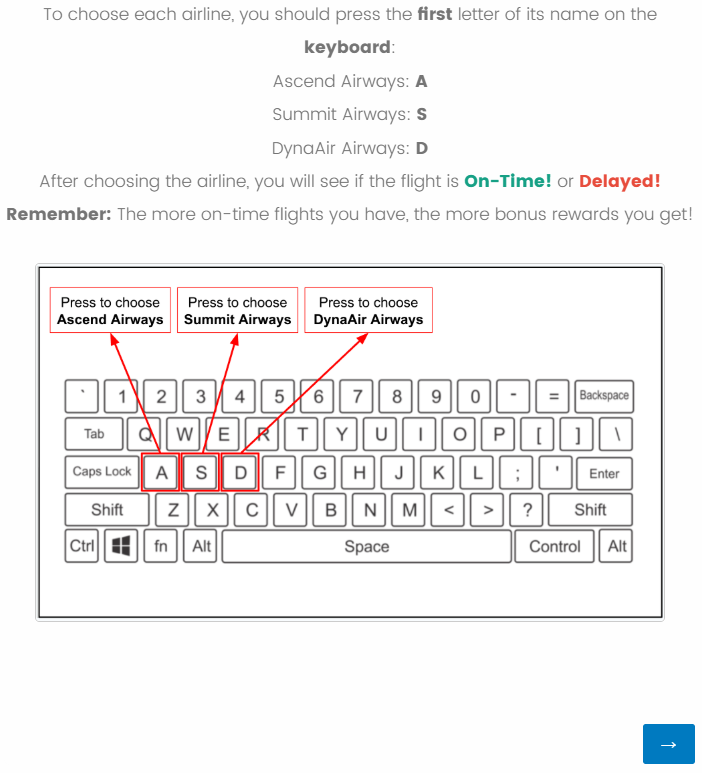}
\caption{Keyboard mappings used to select airlines in the experimental task.}
\label{web-fig:keyboard_instructions}
\end{figure}

\begin{figure}[htp!]
\centering
\includegraphics[width=.6\textwidth]{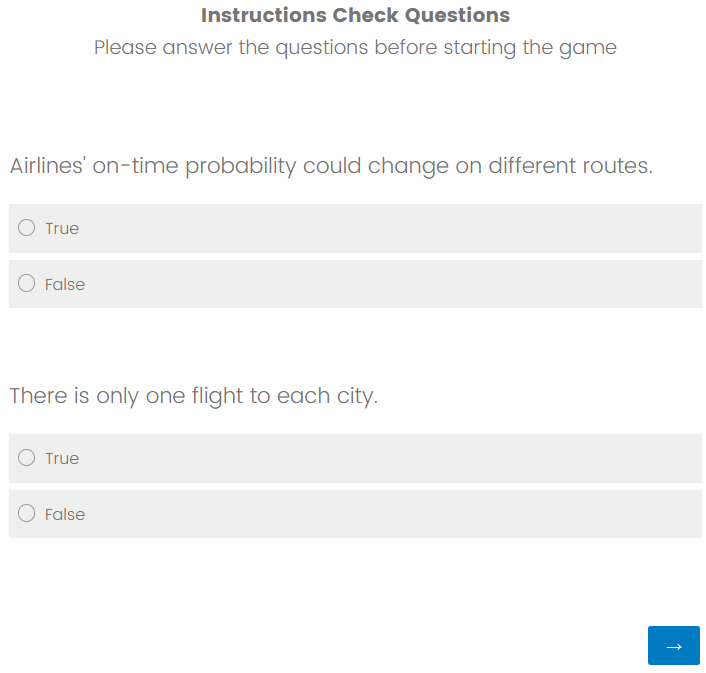}
\caption{Instruction-check questions used to verify participants' understanding of the task.}
\label{web-fig:comprehension_check_instructions}
\end{figure}

\clearpage
\section{Backward Induction Details}
\label{web-sec:backward_induction}

\subsection{Backward Induction Algorithm}
\label{web-ssec:backward_induction_algo}
The Dynamic Programming (DP) policy, as outlined in $\S$\ref{ssec:DP_reordered}, requires the computation of value functions for every possible state at each flight. In this section, we evaluate an \emph{$\epsilon$-greedy} policy: we compute the value of choosing a particular airline $k$, indexed by the noise level $\epsilon$ as
\(
V_{\epsilon}\!\bigl(k \mid \mathbf{N}^+,\mathbf{N}^-;\boldsymbol{\alpha},\boldsymbol{\beta}, t\bigr),
\)
and the corresponding \emph{policy value} of being in a given state under $\epsilon$-greedy, denoted
\(
W_{\epsilon}\!\bigl(\mathbf{N}^+,\mathbf{N}^-;\boldsymbol{\alpha},\boldsymbol{\beta}, t\bigr).
\)
Backward induction proceeds from the final decision point ($t=T$) recursively to the first flight ($t=1$), and considers the expectation implied by the $\epsilon$-greedy policy when calculating the optimal values.

A state at any given time $t$ in the decision process is fully characterized by the priors, accumulated history of choices, and outcomes. In our Bernoulli bandit setting with Beta priors, we can identify each state as
\[
s_t^{DP}=(\mathbf{N}^+,\mathbf{N}^-;\boldsymbol{\alpha},\boldsymbol{\beta}),
\]
where $\mathbf{N}^+=(N_1^+,\ldots,N_K^+)$ and $\mathbf{N}^-=(N_1^-,\ldots,N_K^-)$ are the on-time and delayed counts up to flight $t-1$, and $(\boldsymbol{\alpha},\boldsymbol{\beta})$ are the prior parameters. The posteriors satisfy
\(
\hat{\boldsymbol{\alpha}}=\boldsymbol{\alpha}+\mathbf{N}^+,
\hat{\boldsymbol{\beta}}=\boldsymbol{\beta}+\mathbf{N}^-,
\)
equivalently, componentwise
\begin{align*}
    \hat{\alpha}_{k,t} &= \alpha_{k} + N_{k,t-1}^+, \\
    \hat{\beta}_{k,t} &= \beta_{k} + N_{k,t-1}^- \,,
\end{align*}
and the flight index enters through $\|\mathbf{N}^+\|_1+\|\mathbf{N}^-\|_1=t-1$.

\paragraph{Base Case: Final Flight ($t=T$):}
At the final flight $T$, there are no subsequent flights to consider. Thus, the choice-specific value reduces to the immediate expected reward,
\begin{equation}
    V_{\epsilon}\!\bigl(k \mid \mathbf{N}^+,\mathbf{N}^-;\boldsymbol{\alpha},\boldsymbol{\beta}\bigr)
    \;=\; \hat{\theta}_{k,T}
    \;=\; \frac{\alpha_{k} + N^+_{k,T-1}}{\alpha_{k} + \beta_{k} + N^+_{k,T-1} + N^-_{k,T-1}}.
\end{equation}
The state value averages these one-step values according to the $\epsilon$-greedy rule. With a uniform choice-noise kernel $q(k)=1/K$,
\begin{equation}
\label{web-eq:We_T_uniform}
    W_{\epsilon}\!\bigl(\mathbf{N}^+,\mathbf{N}^-;\boldsymbol{\alpha},\boldsymbol{\beta}\bigr)
    \;=\; (1-\epsilon)\,\max_{k} \hat{\theta}_{k,T}
    \;+\; \frac{\epsilon}{K}\sum_{k=1}^{K} \hat{\theta}_{k,T},
\end{equation}
We calculate these values for all feasible DP states $s_T^{DP}$ that we can reach at time $T$ (i.e., for all $\mathbf{N}^+,\mathbf{N}^-$ with $\|\mathbf{N}^+\|_1+\|\mathbf{N}^-\|_1=T-1$).

\paragraph{Recursive Step for Flights $t < T$:}
For any flight $t < T$, we work backward, assuming we have computed $V_{\epsilon}(\cdot \mid \cdot)$ and $W_{\epsilon}(\cdot)$. The Bellman recursion with the state value $W_{\epsilon}$ gives:
\begin{equation}
\label{web-eq:Vt_eps_appendix}
\begin{aligned}
V_{\epsilon}\!\bigl(k \mid \mathbf{N}^+,\mathbf{N}^-;\boldsymbol{\alpha},\boldsymbol{\beta}\bigr)
&= \hat{\theta}_{k,t} \\
&\quad + \gamma\Bigl[
\hat{\theta}_{k,t}\,W_{\epsilon}\!\bigl(\mathbf{N}^++\mathbf{e}_k,\mathbf{N}^-;\boldsymbol{\alpha},\boldsymbol{\beta}\bigr) \\
&\qquad + \bigl(1-\hat{\theta}_{k,t}\bigr)\,
W_{\epsilon}\!\bigl(\mathbf{N}^+,\mathbf{N}^-+\mathbf{e}_k;\boldsymbol{\alpha},\boldsymbol{\beta}\bigr)
\Bigr].
\end{aligned}
\end{equation}
Given the set of choice-specific values $\{V_{\epsilon}(k \mid \mathbf{N}^+,\mathbf{N}^-;\boldsymbol{\alpha},\boldsymbol{\beta})\}_{k=1}^{K}$, the policy state value under a uniform choice-noise kernel is
\begin{equation}
\label{web-eq:We_uniform_appendix}
    W_{\epsilon}\!\bigl(\mathbf{N}^+,\mathbf{N}^-;\boldsymbol{\alpha},\boldsymbol{\beta}\bigr)
    \;=\;
    (1-\epsilon)\,\max_{k'} V_{\epsilon}\!\bigl(k' \mid \mathbf{N}^+,\mathbf{N}^-;\boldsymbol{\alpha},\boldsymbol{\beta}\bigr)
    \;+\;
    \frac{\epsilon}{K}\sum_{k'=1}^{K} V_{\epsilon}\!\bigl(k' \mid \mathbf{N}^+,\mathbf{N}^-;\boldsymbol{\alpha},\boldsymbol{\beta}\bigr),
\end{equation}
This recursive calculation is repeated by decrementing $t$ from $T-1$ down to $1$. After completing this process, we have the value functions $V_{\epsilon}\!\bigl(k \mid \mathbf{N}^+,\mathbf{N}^-;\boldsymbol{\alpha},\boldsymbol{\beta}, t\bigr)$ and $W_{\epsilon}\!\bigl(\mathbf{N}^+,\mathbf{N}^-;\boldsymbol{\alpha},\boldsymbol{\beta}, t\bigr)$ for all states. At any decision point \(t\), the \(\epsilon\)-greedy policy assigns total probability \(1-\epsilon\) to the value-maximizing airlines and distributes the choice-noise probability \(\epsilon\) uniformly across all \(K\) airlines; thus, \(q(k)=1/K\) for all \(k\).

\paragraph{Computational Considerations:}
The number of possible states $(\hat{\boldsymbol{\alpha}}, \hat{\boldsymbol{\beta}})$ can be very large, making the computation of value functions intensive. A state is defined by the $2K$ parameters $(\hat{\alpha}_1, \dots, \hat{\alpha}_K, \hat{\beta}_1, \dots, \hat{\beta}_K)$ or, equivalently, by the counts $(N_{1,t-1}^+, N_{1,t-1}^-, \dots, N_{K,t-1}^+, N_{K,t-1}^-)$ such that their sum equals $t-1$. As we mentioned before, even for $K=3$ airlines and $T=10$ flights, this can lead to a substantial number of states (e.g., 5005 states), underscoring the computational challenge of implementing full DP. In the next section, we explain how we count the number of possible states and action values for a given number of airlines and flights.

\subsection{State and Action Value Counts}
\label{web-ssec:state_action_count}
In this section, we explain how we can count the number of states and action values in the backward induction algorithm for DP. The main goal here is to show that the state space and, consequently, the amount of calculation growth, is polynomial in $T$
for fixed $K$, and combinatorial in $K$. Let $K$ denote the number of airlines and $T$ the finite horizon (number of flights). As in $\S$\ref{ssec:DP_reordered}, the state at flight $t\in\{1,\ldots,T\}$ is
    \begin{equation}
    (\mathbf{N}^+,\mathbf{N}^-;\boldsymbol{\alpha},\boldsymbol{\beta}),
    \end{equation}
where for each airline $k\in\{1,\ldots,K\}$, the Beta posterior parameters are obtained from the Beta--Bernoulli learning process via $\hat{\alpha}_{k}=\alpha_k+N_k^+$ and $\hat{\beta}_{k}=\beta_k+N_k^-$. We use uninformative priors, such as $(\boldsymbol{\alpha},\boldsymbol{\beta})$ like $\alpha_k = \beta_k = 1 \quad\forall~k \in\{1,\ldots,K\}$. After $t-1$ flights, let $N_k^+$ and $N_k^-$ be, respectively, the numbers of on-time and delayed outcomes observed for airline $k$ up to (but not including) flight $t$. Then

\begin{equation}
\sum_{k=1}^{K} (N_k^+ + N_k^-) \;=\; t-1.
\label{web-eq:sum_constraint}
\end{equation}

Thus, at flight $t$ the feasible DP states are in one-to-one correspondence with the \emph{increment} vectors
\[
\bigl( N_1^+ , N_1^-,\ldots,N_K^+ , N_K^- \bigr)\in\mathbb{Z}_{\ge 0}^{\,2K}
\quad\text{satisfying}\quad \eqref{web-eq:sum_constraint}.
\]
Importantly, the priors $(\boldsymbol{\alpha},\boldsymbol{\beta})$ only shift the posteriors and do not affect the number of distinct states; counting depends solely on the nonnegative-integer solutions to \eqref{web-eq:sum_constraint}.

The number of solutions to \eqref{web-eq:sum_constraint} is the number of weak compositions of $t-1$ into $2K$ parts. By the classical stars-and-bars argument,
\begin{equation}
S_t \;=\; \#\Bigl\{ s_t^{DP} \text{ feasible at flight } t \Bigr\}
\;=\; \binom{(t-1)+(2K-1)}{2K-1}
\;=\; \binom{t+2K-2}{2K-1}.
\label{web-eq:states_per_t}
\end{equation}
where $S_t$ is the total number of possible states at flight $t$. Each such state contributes one state value--$V^*(s_t^{DP})$ in the deterministic case, or $W_{\epsilon}(s_t^{DP})$ under the $\epsilon$-greedy policy, as defined in Equation~\eqref{web-eq:We_uniform_appendix}.
 Equation~\eqref{web-eq:states_per_t} is the exact count of the distinct state values that the DP policy must compute via backward induction for flights $t = 1, \ldots, T$.
\footnote{In a finite-horizon problem, $t$ is part of the state because the continuation value is non-stationary.}
Summing \eqref{web-eq:states_per_t} over flight $t=1,\ldots,T$ yields
\begin{align}
S_{\text{total}}
\;=\; \sum_{t=1}^{T} S_t
&= \sum_{t=1}^{T} \binom{t+2K-2}{\,2K-1\,}
= \sum_{n=0}^{T-1} \binom{n+2K-1}{\,2K-1\,}
= \binom{T+2K-1}{\,2K\,},
\label{web-eq:total_states}
\end{align}
where the last equality follows from the hockey-stick (or Christmas stocking) identity. Equation~\eqref{web-eq:total_states} is the exact count of the distinct values $V^{\ast}(\cdot)$ that the DP policy must compute via backward induction for flights $t=1,\ldots,T$.%
\footnote{If one explicitly includes the terminal boundary condition at $t=T+1$, it adds a single terminal state; \eqref{web-eq:total_states} counts only flight $t\le T$.}

If, instead of state-values, one tabulates all choice-specific values $V\!\bigl(k\mid \mathbf{N}^+,\mathbf{N}^-;\boldsymbol{\alpha},\boldsymbol{\beta}, t\bigr)$ for every airline $k$ at every state, the count is simply
\begin{equation}
K \times S_{\text{total}} \;=\; K \binom{T+2K-1}{\,2K\,}.
\label{web-eq:total_action_values}
\end{equation}
In our canonical example with $K=3$ airlines and $T=10$ flights, we have
\[
S_{\text{total}} \;=\; \binom{10+2\cdot 3 - 1}{\,2\cdot 3\,}
= \binom{15}{6}
= 5005.
\]
For reference, the number of states at the final decision epoch $t=10$ is, from \eqref{web-eq:states_per_t},
\[
S_{10} \;=\; \binom{10+2\cdot 3 - 2}{\,2\cdot 3 - 1\,}
= \binom{14}{5}
= 2002.
\]
If we enumerate all action-values, the total count is $K\times S_{\text{total}}=3\times 5005=15015$.

\clearpage
\section{Dynamic Programming-Based Learning Algorithms}
\label{web-sec:learning_algorithms}
In this section, we present the algorithms for the three policy classes that we formulate in $\S$\ref{sec:dp_models}---DP, MetaDP, and BRMDP. 
\subsection{Baseline DP Algorithm}
\label{web-ssec:baseline_dp_algo}
DP is the first algorithm that we explain as our baseline in $\S$\ref{ssec:DP_reordered}. This algorithm treats each route as a completely separate problem and begins solving it with uninformative priors. Accordingly, it has no meta-learning mechanism.
\begin{algorithm}[htp!]
\caption{Baseline Dynamic Programming Policies (DP)}
\label{web-alg:dynamic_programming_reordered}
\begin{algorithmic}[1]
\State \textbf{Inputs:} priors $(\boldsymbol{\alpha},\boldsymbol{\beta})$, policy noise $\epsilon$
\State \textbf{Precompute (backward induction):} for each $t=T,T-1,\ldots,1$ and each feasible state $(\mathbf{N}^+,\mathbf{N}^-)$ with $\|\mathbf{N}^+\|_1+\|\mathbf{N}^-\|_1=t-1$, compute $V_\epsilon(\cdot \mid \mathbf{N}^+,\mathbf{N}^-)$ and $W_\epsilon(\mathbf{N}^+,\mathbf{N}^-)$ using Equations~\eqref{eq:posterior_mean_reordered}, \eqref{eq:We_uniform_reordered}, and \eqref{eq:Ve_eps_single_reordered}.
\Statex
\State \textbf{Forward play (execute the policy within a route)}
\State Initialize counts $(\mathbf{N}^+,\mathbf{N}^-)\gets \mathbf{0}$
\For{$t=1$ \textbf{to} $T$}
  \State Sample $A_t \sim \pi^{DP}_{\epsilon}(\cdot \mid \mathbf{N}^+,\mathbf{N}^-;\boldsymbol{\alpha},\boldsymbol{\beta})$ \hfill (Equation~\eqref{eq:e_greedy_probability_func_reordered})
  \State Observe realized outcome $y_t \in \{0,1\}$ and update $(\mathbf{N}^+,\mathbf{N}^-)$ as in Algorithm~\ref{alg:bayesian_update_reordered}
\EndFor
\end{algorithmic}
\end{algorithm}

\subsection{MetaDP Algorithm}
\label{web-ssec:metadp_algo}
MetaDP is our algorithm described in $\S$\ref{ssec:metaDP_reordered} for transferring knowledge in an optimal, Bayesian manner. It carries a hyper-posterior across routes and uses it to the fullest extent.
\begin{algorithm}[htp!]
\caption{Meta Dynamic Programming (MetaDP)}
\label{web-alg:roundwise_metaDP_reordered}
\begin{algorithmic}[1]
\State \textbf{Inputs:} hypothesis set $\mathcal{P}=\{(\boldsymbol{\alpha}^{(j)},\boldsymbol{\beta}^{(j)})\}_{j=1}^{J}$, initial hyper-prior $Q_1$, policy noise $\epsilon$
\For{$m=1,\ldots,M$}
    \State \textbf{Backward induction within route $m$: treat $Q_m$ as fixed}
    \For{each $t=T,T-1,\ldots,1$ and each feasible state $(\mathbf{N}_m^+,\mathbf{N}_m^-)$ with $\|\mathbf{N}_m^+\|_1+\|\mathbf{N}_m^-\|_1=t-1$}
        \If{$t=T$}
            \State Compute $\bar{V}_{\epsilon}(\cdot \mid \mathbf{N}_m^+,\mathbf{N}_m^-,Q_m;\mathcal{P})$ using Equation~\eqref{eq:Vbar_meta_T_reordered}
        \Else
            \State For each hypothesis $j$ and airline $k$, compute $V_{\epsilon}^{(j)}(k \mid \mathbf{N}_m^+,\mathbf{N}_m^-,Q_m;\mathcal{P})$ using Equation~\eqref{eq:Ve_eps_category_reordered}
            \State Compute $\bar{V}_{\epsilon}(\cdot \mid \mathbf{N}_m^+,\mathbf{N}_m^-,Q_m;\mathcal{P})$ using Equation~\eqref{eq:Vbar_meta_reordered}
        \EndIf
        \State Compute $\bar{W}_{\epsilon}(\mathbf{N}_m^+,\mathbf{N}_m^-,Q_m;\mathcal{P})$ using Equation~\eqref{eq:Wbar_meta_reordered}
    \EndFor
    \State \textbf{Forward play within route $m$:} initialize $(\mathbf{N}_m^+,\mathbf{N}_m^-)\gets \mathbf{0}$
    \For{$t=1,\ldots,T$}
        \State Choose airline $A_{m,t}\sim \pi_{\epsilon}^{MetaDP}(\cdot \mid \mathbf{N}_m^+,\mathbf{N}_m^-,Q_m;\mathcal{P})$ \hfill (Equation~\eqref{eq:e_greedy_mixed_probability_func_reordered})
        \State Observe outcome $y_{m,t}\in\{0,1\}$ and update $(\mathbf{N}_m^+,\mathbf{N}_m^-)$ as in Algorithm~\ref{alg:bayesian_update_reordered}
    \EndFor
    \State Update the hyper-posterior from $Q_m$ to $Q_{m+1}$ using Algorithm~\ref{alg:cross_route_update_reordered}
\EndFor
\end{algorithmic}
\end{algorithm}

\subsection{BRMDP Algorithm}
\label{web-ssec:brmdp_algo}
We propose BRMDP as a boundedly rational policy in $\S$\ref{ssec:BRMDP_reordered}. In the core specification, this policy uses the number of hyper-posterior draws, \(D\), to control how finely prior uncertainty is represented during route-level planning.
\begin{algorithm}[htp!]
\caption{Boundedly Rational Meta Dynamic Programming (BRMDP)}
\label{web-alg:BRMDP_simple_reordered}
\begin{algorithmic}[1]
\State \textbf{Inputs:} $\mathcal{P}=\{(\boldsymbol{\alpha}^{(j)},\boldsymbol{\beta}^{(j)})\}_{j=1}^{J}$, initial hyper-prior $Q_1$, policy noise $\epsilon$, draws $D$
\For{$m=1,\ldots,M$}
  \State \textbf{Route weights:} draw $\mathbf{x}_m \sim \mathrm{Multinomial}(D,Q_m)$ and set $w_j(\mathbf{x}_m)\gets x_{j,m}/D$ for $j=1,\ldots,J$
  \Statex
  \State \textbf{Backward induction within route $m$}
  \For{$t=T$ \textbf{down to} $1$}
    \For{each feasible state $(\mathbf{N}^+,\mathbf{N}^-)$ such that $\|\mathbf{N}^+\|_1+\|\mathbf{N}^-\|_1=t-1$}
      \If{$t=T$}
        \State For all $k$: compute $\tilde{V}_{\epsilon}(k \mid \mathbf{N}^+,\mathbf{N}^-,\mathbf{x}_m;\mathcal{P})$ using Equation~\eqref{eq:Vbar_meta_resource_T_reordered}
      \Else
        \State For each $j,k$: compute $V^{(j)}_{\epsilon}(k \mid \mathbf{N}^+,\mathbf{N}^-,\mathbf{x}_m;\mathcal{P})$ using Equation~\eqref{eq:Ve_resource_category_reordered}
        \State For each $k$: compute $\tilde{V}_{\epsilon}(k \mid \mathbf{N}^+,\mathbf{N}^-,\mathbf{x}_m;\mathcal{P})$ using Equation~\eqref{eq:Vbar_meta_resource_reordered}
      \EndIf
      \State Compute $\tilde{W}_{\epsilon}(\mathbf{N}^+,\mathbf{N}^-,\mathbf{x}_m;\mathcal{P})$ using Equation~\eqref{eq:Wbar_meta_resource_reordered}
    \EndFor
  \EndFor
  \Statex
  \State \textbf{Forward play within route $m$}
  \State \textbf{Reset counts:} $(\mathbf{N}_m^+,\mathbf{N}_m^-)\gets \mathbf{0}$
  \For{$t=1,\ldots,T$}
    \State Choose $A_{m,t}\sim \pi^{\mathrm{BRMDP}}_{\epsilon,D}(\cdot \mid \mathbf{N}_m^+,\mathbf{N}_m^-,\mathbf{x}_m;\mathcal{P})$
    \State Observe $y_{m,t}\in\{0,1\}$ and update $(\mathbf{N}_m^+,\mathbf{N}_m^-)$ as in Algorithm~\ref{alg:bayesian_update_reordered}
  \EndFor
  \State \textbf{Cross-route update:} update $Q_m$ to $Q_{m+1}$ as in Algorithm~\ref{alg:cross_route_update_reordered}
\EndFor
\end{algorithmic}
\end{algorithm}

\clearpage
\section{Likelihood Estimation Details}
\label{web-sec:likelihood_estimation_details}

\subsection{DP and MetaDP Likelihood Estimation}
\label{web-ssec:single_round_likelihood_reordered}

\paragraph{Observed histories and likelihood logic.}
Recall that we defined \(\mathcal H_{1:m-1} = (\mathcal{H}_1,\ldots,\mathcal{H}_{m-1})\) as the concatenated history of the previous routes before route $m$ in $\S$\ref{ssec:multi_route_problem}.
Further, let the within-route history available before flight \(t\) on route \(m\) be $\mathcal{H}_{m,t-1} = \bigl((A_{m,1}, y_{m,1}), \ldots, (A_{m,t-1}, y_{m,t-1})\bigr)$. Then, the full information set before decision \(t\) on route \(m\) is denoted as $\mathcal{I}_{m,t-1} = \bigl(\mathcal{H}_{1:m-1},\mathcal{H}_{m,t-1}\bigr)$. Further, let \(\mathbf{N}_{m,t-1}^{+}\) and \(\mathbf{N}_{m,t-1}^{-}\) denote the vectors of on-time and delayed counts implied by \(\mathcal{H}_{m,t-1}\). The current flight index is pinned down by these counts through $
t = \|\mathbf{N}_{m,t-1}^{+}\|_1 + \|\mathbf{N}_{m,t-1}^{-}\|_1 + 1$.

The identification logic is simple. Each policy maps the observed history into a state, and each state implies a probability for the action actually chosen. Multiplying those trial-by-trial probabilities along the observed path gives the route-level likelihood; summing logs gives the participant-level log-likelihood.

\paragraph{DP likelihood.}
We begin with the baseline DP policy from $\S$\ref{ssec:DP_reordered}. Under baseline DP, previous routes are irrelevant because the policy does not transfer information across routes. Hence, although we write the likelihood as a function of \(\mathcal{I}_{m,t-1}\) for notational symmetry, only the current-route counts matter:
\begin{equation}
P\!\left(A_{m,t}=\tilde{k}\mid \mathcal{I}_{m,t-1}, \pi^{DP}, \epsilon\right)
=
\pi^{DP}_{\epsilon}\!\bigl(\tilde{k}\mid \mathbf{N}_{m,t-1}^{+},\mathbf{N}_{m,t-1}^{-};\boldsymbol{\alpha},\boldsymbol{\beta}\bigr)
\label{web-eq:DP_policy_likelihood_reordered}
\end{equation}
where $\pi^{DP}_{\epsilon}\!\bigl(\tilde{k}\mid \mathbf{N}_{m,t-1}^{+},\mathbf{N}_{m,t-1}^{-};\boldsymbol{\alpha},\boldsymbol{\beta}\bigr)$ is defined in Equation~\eqref{eq:e_greedy_probability_func_reordered}. Then, 
the participant-level log-likelihood is
\begin{equation}
   \ell_{DP}
   =
   \sum_{m=1}^{M} \sum_{t=1}^{T}
   \log
   \pi^{DP}_{\epsilon}\!\bigl(
   A_{m,t}\mid \mathbf{N}_{m,t-1}^{+},\mathbf{N}_{m,t-1}^{-};\boldsymbol{\alpha},\boldsymbol{\beta}
   \bigr).
   \label{web-eq:simple_strategy_likelihood_reordered}
\end{equation}
In practice, we set \(\boldsymbol{\alpha}=\mathbf{1}\) and \(\boldsymbol{\beta}=\mathbf{1}\), so DP starts each route from uninformative priors. This makes DP the natural no-transfer benchmark.

\paragraph{MetaDP likelihood.}
For MetaDP, the logic is the same, but the relevant state is richer. In addition to the current-route counts, the policy uses the hyper-posterior \(Q_m\), which summarizes what has been learned from routes \(1,\ldots,m-1\). Thus, the probability of the observed action at flight \(t\) on route \(m\) is
\begin{equation}
P\!\left(A_{m,t}=\tilde{k}\mid \mathcal{I}_{m,t-1}, \pi^{MetaDP}, \epsilon\right) = 
\pi^{MetaDP}_{\epsilon}\!\bigl(\tilde{k}\mid \mathbf{N}_{m,t-1}^{+},\mathbf{N}_{m,t-1}^{-},Q_m;\mathcal{P}\bigr)
\label{web-eq:MetaDP_policy_likelihood_reordered}
\end{equation}
where $\pi^{MetaDP}_{\epsilon}\!\bigl(\tilde{k}\mid \mathbf{N}_{m,t-1}^{+},\mathbf{N}_{m,t-1}^{-},Q_m;\mathcal{P}\bigr)$ is defined in Equation~\eqref{eq:e_greedy_mixed_probability_func_reordered}. Because \(Q_m\) is updated only between routes, it is fixed throughout route \(m\) and fully determined by \(\mathcal{H}_{1:m-1}\). The resulting participant-level log-likelihood is
\begin{equation}
    \ell_{MetaDP}
    =
    \sum_{m=1}^{M} \sum_{t=1}^{T}
    \log
    \pi_{\epsilon}^{MetaDP}\!\bigl(
    A_{m,t}\mid \mathbf{N}_{m,t-1}^{+},\mathbf{N}_{m,t-1}^{-},Q_m;\mathcal{P}
    \bigr).
    \label{web-eq:metadp_strategy_likelihood_reordered}
\end{equation}

Taken together, Equations~\eqref{web-eq:simple_strategy_likelihood_reordered} and \eqref{web-eq:metadp_strategy_likelihood_reordered} identify two benchmark extremes. DP treats every route as a fresh problem; MetaDP carries forward all cross-route information and uses it in fully forward-looking planning. Therefore, comparing these likelihoods reveals whether the data support no transfer or full transfer.

\subsection{BRMDP Likelihood Estimation}
\label{web-ssec:meta_dp_d_draws_reordered}

\paragraph{Why BRMDP is more difficult to estimate.}
Estimating the likelihood of BRMDP from $\S$\ref{ssec:BRMDP_reordered} is more challenging because the researcher does not observe consumers' route-level internal draws from the hyper-posterior. Recall that before route \(m\), a BRMDP consumer draws \(D\) hypothesis indices from the hyper-posterior \(Q_m\), forms the count vector \(\mathbf{x}_m\), converts it into weights \(w_j(\mathbf{x}_m)=x_{j,m}/D\), and then solves the route-level dynamic program conditional on that sampled mixture. Conditional on \(\mathbf{x}_m\), the policy generates well-defined trial-by-trial choice probabilities. Unconditionally, however, \(\mathbf{x}_m\) is latent, so the route likelihood must integrate over all possible count vectors.

This route-level integration is important. The latent draw \(\mathbf{x}_m\) is made once per route and then held fixed throughout that route. Therefore, the expectation is taken over route-level latent states, not independently at each flight. The exact route-\(m\) likelihood is
\begin{equation}
\label{web-eq:round_like_exact_meta_reordered}
P_m^{(D)}
=
\mathbb{E}_{\mathbf{x}_m\sim \mathrm{Multinomial}(D,Q_m)}
\left[
\prod_{t=1}^{T}
\pi_{\epsilon,D}^{BRMDP}\!\bigl(
A_{m,t}\mid \mathbf{N}_{m,t-1}^{+},\mathbf{N}_{m,t-1}^{-},\mathbf{x}_m;\mathcal{P}
\bigr)
\right].
\end{equation}

Because \(\mathbf{x}_m\) is a count vector from \(D\) i.i.d.\ categorical draws with probabilities \(Q_m\), its probability mass function is multinomial:
\[
\Pr(\mathbf{x}_m=\mathbf{x}\mid Q_m)
=
\frac{D!}{\prod_{j=1}^{J} x_j!}
\prod_{j=1}^{J} Q_m(j)^{x_j},
\qquad
\sum_{j=1}^{J} x_j = D.
\]
Substituting this into Equation~\eqref{web-eq:round_like_exact_meta_reordered} yields the exact finite-sum representation
\begin{equation}
\label{web-eq:round_like_counts_meta_reordered}
P_m^{(D)}
=
\sum_{\substack{\mathbf{x}\in\mathbb{N}^J\\ \sum_j x_j=D}}
\Pr(\mathbf{x}_m=\mathbf{x}\mid Q_m)
\prod_{t=1}^{T}
\pi_{\epsilon,D}^{BRMDP}\!\bigl(
A_{m,t}\mid \mathbf{N}_{m,t-1}^{+},\mathbf{N}_{m,t-1}^{-},\mathbf{x};\mathcal{P}
\bigr).
\end{equation}
The participant-level log-likelihood is then
\begin{equation}
\label{web-eq:ll_meta_reordered}
\ell_{BRMDP}^{(D)}
=
\sum_{m=1}^{M}\log P_m^{(D)}.
\end{equation}

This expression makes the identification problem transparent. The BRMDP likelihood differs from MetaDP likelihood not because the policy rule is unknown, but because the policy depends on latent internal randomness that must be integrated out. In other words, BRMDP is identified through the distribution of route-level sampled mixtures that it induces.

\paragraph{Monte Carlo estimator.}
Exact evaluation of Equation~\eqref{web-eq:round_like_counts_meta_reordered} is infeasible once \(D\) or \(J\) becomes even moderately large, because the number of possible count vectors grows combinatorially. We therefore approximate the route-level expectation by Monte Carlo (see Web Appendix~\ref{web-ssec:impossible_meta_likelihood} for additional computational details). For each route \(m\), we draw \(B\) i.i.d.\ samples \(\mathbf{x}_m^{(b)}\sim \mathrm{Multinomial}(D,Q_m)\). Conditional on each draw, we solve the induced BRMDP problem once, evaluate the observed route path, and then average the resulting route probabilities:
\begin{equation}
\label{web-eq:round_mc_meta_reordered}
\begin{aligned}
\widehat{P}_m^{(D)}
= \frac{1}{B}\sum_{b=1}^{B}\exp\!\left\{
\sum_{t=1}^{T}\log \right. \left.
\pi_{\epsilon,D}^{BRMDP}\!\bigl(
A_{m,t}\mid \mathbf{N}_{m,t-1}^{+},\mathbf{N}_{m,t-1}^{-},\mathbf{x}_m^{(b)};\mathcal{P}
\bigr)
\right\}, \quad
\widehat{\ell}_{BRMDP}^{(D)}
= \sum_{m=1}^{M}\log \widehat{P}_m^{(D)}.
\end{aligned}
\end{equation}
By construction, \(\widehat{P}_m^{(D)}\) is an unbiased and consistent estimator of the exact route likelihood \(P_m^{(D)}\). As \(B\) increases, this average concentrates on the true marginal likelihood \(P_m^{(D)}\), giving a consistent estimate of how likely the observed choices are under a boundedly rational sampler. The corresponding log-likelihood \(\widehat{\ell}_{BRMDP}^{(D)}\) is therefore a simulation-based approximation to the BRMDP fit. \footnote{For finite $B$, $\log \widehat P_m^{(D)}$ is not an unbiased estimator of $\log P_m^{(D)}$. Since the logarithm is concave, Jensen's inequality implies
\(
\mathbb{E}\!\left[\log \widehat P_m^{(D)}\right]
\leq 
\log \mathbb{E}\!\left[\widehat P_m^{(D)}\right]
=
\log P_m^{(D)}.
\)
Thus, the simulated BRMDP log-likelihood has a finite-simulation downward bias. This bias vanishes as $B \to \infty$.
}

Intuitively, the BRMDP likelihood integrates over the agent's unobserved internal randomness: instead of conditioning on a particular prior-sample \(\mathbf{x}_m\), we average the product of choice probabilities across all samples that the policy could draw from \(Q_m\). Because enumerating those samples is intractable for moderate \(D\) and \(J\), we approximate this integral by Monte Carlo---simulate \(\mathbf{x}_m\sim\text{Multinomial}(D,Q_m)\), solve the induced BRMDP problem once per draw, and average the resulting route-level probabilities. Please see Algorithm~\ref{web-alg:metaDP_lik_mc_reordered} in Appendix~\ref{web-ssec:algo_montecarlo} for a detailed summary of this Monte Carlo procedure. 

\subsection{Why is Exact Meta-Learning Likelihood Estimation Not Feasible?}
\label{web-ssec:impossible_meta_likelihood}
The sum in Equation~\eqref{web-eq:round_like_counts_meta_reordered} ranges over all count vectors $x \in \mathbb{N}^J$ with $\sum_{j=1}^J x_j = D$. By stars-and-bars, the number of terms is
\begin{equation}
    N_X(D,J) = \binom{D+J-1}{J-1} \asymp \frac{D^{J-1}}{(J-1)!}.
\end{equation}
For each $x$, we must run one integrated backward induction to obtain the corresponding route-level choice probabilities. Using the state convention from Appendix~\ref{web-ssec:state_action_count}, at flight $t$ there are
\[
\binom{t+2K-2}{2K-1}
\]
distinct within-route states. Summing over decision epochs $t = 1, \ldots, T$ gives
\begin{equation} 
    S(T,K) = \sum_{t=1}^{T} \binom{t+2K-2}{2K-1}
    = \sum_{n=0}^{T-1} \binom{n+2K-1}{2K-1}
    = \binom{T+2K-1}{2K}
    \asymp \frac{T^{2K}}{(2K)!}.
\end{equation}
Evaluating all choice-specific values over these states costs at least
\begin{equation}
    \mathrm{cost}_{DP}(T,K) = \Theta\!\bigl(K\,S(T,K)\bigr).
\end{equation}
Hence, a per-route lower bound on exact evaluation is
\begin{equation}  
    \mathrm{work}^{\mathrm{exact}}_m
    \gtrsim N_X(D,J)\times \mathrm{cost}_{DP}(T,K)
    =
    \Theta\!\left(
    \binom{D+J-1}{J-1}
    K
    \binom{T+2K-1}{2K}
    \right).
\end{equation}
Even moderate values are prohibitive:
\[
(J,D,K,T) = (8,50,4,40):\qquad
N_X = \binom{57}{7} = 2.64 \times 10^8,\qquad
S = \binom{47}{8} = 3.14 \times 10^8,
\]
\[
\Rightarrow \mathrm{work}^{\mathrm{exact}}_m \gtrsim N_X \cdot K S \approx 3.33 \times 10^{17};
\]
\[
(J,D,K,T) = (10,100,5,40):\qquad
N_X = \binom{109}{9} = 4.26 \times 10^{12},\qquad
S = \binom{49}{10} = 8.22 \times 10^9,
\]
\[
\Rightarrow \mathrm{work}^{\mathrm{exact}}_m \gtrsim 1.75 \times 10^{23}.
\]

Memory also binds: one double per state needs $8S(T,K)$ bytes, so $S = 3.14 \times 10^8$ already implies roughly $2.5$ GB for a single value array (and several arrays are required). Because the integrated values $\bar V_{\epsilon}$ depend on $w = x/D$, distinct $x$ prevent reuse and preserve the $\binom{D+J-1}{J-1}$ factor.

Conclusion. Exact evaluation scales as
\begin{equation}
\mathrm{work}^{\mathrm{exact}}_m
=
\Theta\!\left(
\frac{K}{(J-1)!(2K)!} D^{J-1} T^{2K}
\right),
\end{equation}
which grows combinatorially in $(J,D)$ and as a high-degree polynomial in $T$, rendering Equation~\eqref{web-eq:round_like_counts_meta_reordered} intractable beyond small cases. This motivates the route-level Monte Carlo estimator in Equation~\eqref{web-eq:round_mc_meta_reordered}.

\subsection{Algorithm for Monte Carlo Procedure}
\label{web-ssec:algo_montecarlo}
We provide the pseudocode for the algorithm we use to compute the BRMDP log-likelihood using the Monte Carlo method.
\begin{algorithm}[htp!]
\caption{BRMDP(\(D\)) Monte Carlo Log-Likelihood}
\label{web-alg:metaDP_lik_mc_reordered}
\begin{algorithmic}[1]
\State \textbf{Input:} observed data; prior sets $\mathcal{P}$; initial hyper-prior $Q_1$; policy noise $\epsilon$; hyper-draws $D$; Monte Carlo samples $B$
\State \textbf{Initialize:} $\widehat{\ell}_{BRMDP}^{(D)} \gets 0$
\For{$m=1$ to $M$}
  \State $a_m \gets -\infty$;\quad $s_m^{(b)} \gets 0$ for all $b=1,\ldots,B$
  \For{$b=1$ to $B$} \Comment{route-level Monte Carlo over latent count vectors}
     \State Draw $\mathbf{x}_m^{(b)}\sim \mathrm{Multinomial}(D,Q_m)$ and set $w_j(\mathbf{x}_m^{(b)}) \gets x_{j,m}^{(b)}/D$
     \State Precompute $\tilde{V}_{\epsilon}(\cdot \mid \cdot,\mathbf{x}_m^{(b)};\mathcal{P})$ and $\tilde{W}_{\epsilon}(\cdot,\mathbf{x}_m^{(b)};\mathcal{P})$ by backward induction using Equations~\eqref{eq:Ve_resource_category_reordered}--\eqref{eq:Wbar_meta_resource_reordered}
     \State Initialize $(\mathbf{N}^{+},\mathbf{N}^{-})\gets \mathbf{0}$
     \For{$t=1$ to $T$} \Comment{evaluate the policy on the observed route history}
        \State $s_m^{(b)} \gets s_m^{(b)} + \log \!\left[
        \pi_{\epsilon,D}^{BRMDP}\!\bigl(
        A_{m,t}\mid \mathbf{N}^{+},\mathbf{N}^{-},\mathbf{x}_m^{(b)};\mathcal{P}
        \bigr)
        \right]$
        \State Update $(\mathbf{N}^{+},\mathbf{N}^{-})$ using the observed outcome $y_{m,t}$ as in Algorithm~\ref{alg:bayesian_update_reordered}
     \EndFor
     \State $a_m \gets \max(a_m, s_m^{(b)})$
  \EndFor
  \State $\log \widehat{P}_m^{(D)}
  \gets
  a_m + \log\!\Bigl(\frac{1}{B}\sum_{b=1}^{B}\exp\{s_m^{(b)}-a_m\}\Bigr)$
  \State $\widehat{\ell}_{BRMDP}^{(D)} \gets \widehat{\ell}_{BRMDP}^{(D)} + \log \widehat{P}_m^{(D)}$
  \State Compute $\{N_{k,m}^{+},N_{k,m}^{-},T_{k,m}\}_{k=1}^{K}$ from $\{(A_{m,t},y_{m,t})\}_{t=1}^{T}$
  \State Update the hyper-posterior from $Q_m$ to $Q_{m+1}$ using Algorithm~\ref{alg:cross_route_update_reordered}
\EndFor
\State \textbf{return} $\widehat{\ell}_{BRMDP}^{(D)}$
\end{algorithmic}
\end{algorithm}

\clearpage
\section{Robustness Checks and Extensions}
\label{web-ssec:checks_extensions}

\subsection{Choice-Path Evidence of Knowledge Transfer}
\label{web-ssec:choice_path_transfer}

The best-airline selection and pseudo-regret measures in
$\S$\ref{sec:behavioral_evidence_meta_learning} evaluate participants'
choices relative to the route-specific on-time rates
\(\theta_{m,k}^{\ast}\). These environmental properties provide natural
measures of decision quality, but participants do not observe them. We therefore
complement those measures with a choice-path analysis that relates the first
choice on each new route to a posterior summary of the participant's own earlier
outcomes. This analysis shows whether, before observing any feedback on the current route, a participant is more likely to choose an airline that performed well in their earlier experience. A positive correlation between each airline's observed performance and the participants' chance of selecting that airline at the start of each route corroborates the reduced-form results we already have. 

To construct the posterior summaries, we add the participant index \(i\) to the
choice and outcome notation from
$\S$\ref{ssec:experimental_task}. Thus, \(A_{i,m,t}\) is participant \(i\)'s
choice on flight \(t\) of route \(m\), \(N^{+}_{i,k,m}\) and
\(N^{-}_{i,k,m}\) count the on-time and delayed outcomes that participant \(i\)
observes after choosing airline \(k\) on that route, and
\(T_{i,k,m}=N^{+}_{i,k,m}+N^{-}_{i,k,m}\). Let
\(\mathcal{H}_{i,1:m-1}\) denote the corresponding participant-specific history
available at the beginning of route \(m\).

We summarize this history using the same independent
\(\operatorname{Beta}(1,1)\) prior that the baseline updating model employs.
First, the cumulative posterior mean for airline \(k\) before route \(m\) is
\begin{equation}
    \widehat{\theta}^{\mathrm{cum}}_{i,k,m-1}
    =
    \frac{
        1+\sum_{\widetilde{m}=1}^{m-1}N^{+}_{i,k,\widetilde{m}}
    }{
        2+\sum_{\widetilde{m}=1}^{m-1}T_{i,k,\widetilde{m}}
    },
    \qquad m=2,\ldots,M.
    \label{web-eq:choice_path_cumulative_posterior}
\end{equation}
This posterior uses every outcome that the participant observed on routes
\(1,\ldots,m-1\). To
distinguish accumulated experience from one-route carryover, we also construct a
posterior using only the immediately preceding route:
\begin{equation}
    \widehat{\theta}^{\mathrm{last}}_{i,k,m-1}
    =
    \frac{
        1+N^{+}_{i,k,m-1}
    }{
        2+T_{i,k,m-1}
    }.
    \label{web-eq:choice_path_previous_posterior}
\end{equation}
If the participant does not choose an airline in the relevant history, its
posterior mean remains \(1/2\). These auxiliary posteriors provide descriptive
summaries of the outcomes that participants observed. In particular, they differ
from the hyper-posterior \(Q_m\) and do not impose the finite hypothesis class
that MetaDP and BRMDP use.

For each history definition, we also identify the airline or tied airlines with
the largest posterior mean:
\begin{align}
    \mathcal{B}^{\mathrm{cum}}_{i,m-1}
    &=
    \argmax_{k\in\{1,\ldots,K\}}
    \widehat{\theta}^{\mathrm{cum}}_{i,k,m-1},
    \nonumber\\
    \mathcal{B}^{\mathrm{last}}_{i,m-1}
    &=
    \argmax_{k\in\{1,\ldots,K\}}
    \widehat{\theta}^{\mathrm{last}}_{i,k,m-1},
    \label{web-eq:choice_path_posterior_best_sets}
\end{align}
and retain all ties. The cumulative set reflects the participant's full earlier history, whereas the previous-route set isolates the information from route \(m-1\).

We estimate alternative-level conditional logits for the first choice on routes
\(m=2,\ldots,M\). Let
\(\zeta\in\{\mathrm{cum},\mathrm{last}\}\) index the history definition, let
\(e(i)\) denote participant \(i\)'s experimental condition, and let
\(\delta_{e(i),k}\) denote condition-specific airline fixed effects. Following
the condition-interaction structure in
Equation~\eqref{eq:regret_reward_regression_new}, let
\(\mathbf{1}_{\mathrm{experiment},i}\) denote the three-vector of indicators for
the non-baseline experimental conditions. We omit Far Means--Low Var. The
posterior-mean specifications are
\begin{equation}
\resizebox{.9\linewidth}{!}{$\displaystyle
\Pr\!\left(
        A_{i,m,1}=k
        \,\middle|\,
        \mathcal{H}_{i,1:m-1}
    \right)
    =
    \frac{
        \exp\!\left\{
            \psi_{\zeta}
            \widehat{\theta}^{\zeta}_{i,k,m-1}
            +
            (\boldsymbol{\psi}_{\zeta}^{\mathrm{int}})^{\top}
            \!\left(
                \widehat{\theta}^{\zeta}_{i,k,m-1}
                \times \mathbf{1}_{\mathrm{experiment},i}
            \right)
            +
            \delta_{e(i),k}
        \right\}
    }{
        \sum_{k'=1}^{K}
        \exp\!\left\{
            \psi_{\zeta}
            \widehat{\theta}^{\zeta}_{i,k',m-1}
            +
            (\boldsymbol{\psi}_{\zeta}^{\mathrm{int}})^{\top}
            \!\left(
                \widehat{\theta}^{\zeta}_{i,k',m-1}
                \times \mathbf{1}_{\mathrm{experiment},i}
            \right)
            +
            \delta_{e(i),k'}
        \right\}
    }.
$}
\label{web-eq:choice_path_posterior_mean_logit}
\end{equation}
For the posterior-best specifications, let
\(\iota^{\zeta}_{i,k,m-1}=\mathbf{1}\{k\in\mathcal{B}^{\zeta}_{i,m-1}\}\). We then
estimate
\begin{equation}
\resizebox{.9\linewidth}{!}{$\displaystyle
\Pr\!\left(
        A_{i,m,1}=k
        \,\middle|\,
        \mathcal{H}_{i,1:m-1}
    \right)
    =
    \frac{
        \exp\!\left\{
            \varphi_{\zeta}
            \iota^{\zeta}_{i,k,m-1}
            +
            (\boldsymbol{\varphi}_{\zeta}^{\mathrm{int}})^{\top}
            \!\left(
                \iota^{\zeta}_{i,k,m-1}
                \times \mathbf{1}_{\mathrm{experiment},i}
            \right)
            +
            \delta_{e(i),k}
        \right\}
    }{
        \sum_{k'=1}^{K}
        \exp\!\left\{
            \varphi_{\zeta}
            \iota^{\zeta}_{i,k',m-1}
            +
            (\boldsymbol{\varphi}_{\zeta}^{\mathrm{int}})^{\top}
            \!\left(
                \iota^{\zeta}_{i,k',m-1}
                \times \mathbf{1}_{\mathrm{experiment},i}
            \right)
            +
            \delta_{e(i),k'}
        \right\}
    }.
$}
\label{web-eq:choice_path_posterior_best_logit}
\end{equation}
The coefficients \(\psi_{\zeta}\) and \(\varphi_{\zeta}\) capture the
posterior relationships in the Far Means--Low Var. condition. The elements of
\(\boldsymbol{\psi}_{\zeta}^{\mathrm{int}}\) and
\(\boldsymbol{\varphi}_{\zeta}^{\mathrm{int}}\) measure how those relationships
differ in the other three conditions.
We treat each participant-route as one choice set containing all three airlines. The
choice-set likelihood conditions out the participant-route intercept. A separate
participant fixed effect would enter all three alternatives in a choice set
equally and would therefore cancel from Equations~\eqref{web-eq:choice_path_posterior_mean_logit} and
\eqref{web-eq:choice_path_posterior_best_logit}. We cluster standard errors at the
participant level to account for repeated route-entry choices.

\begin{table}[htp!]
\centering
\small
\begingroup
\setlength{\tabcolsep}{2.5pt}
\begin{tabular}{lcccccccc}
\toprule
& \multicolumn{4}{c}{\textbf{Posterior Mean}}
& \multicolumn{4}{c}{\textbf{Posterior-Best Indicator}}\\
\cmidrule(lr){2-5}\cmidrule(lr){6-9}
& \multicolumn{2}{c}{All Earlier Routes}
& \multicolumn{2}{c}{Previous Route}
& \multicolumn{2}{c}{All Earlier Routes}
& \multicolumn{2}{c}{Previous Route}\\
\cmidrule(lr){2-3}\cmidrule(lr){4-5}
\cmidrule(lr){6-7}\cmidrule(lr){8-9}
& Coef. & \(\mathit{p}\)-value
& Coef. & \(\mathit{p}\)-value
& Coef. & \(\mathit{p}\)-value
& Coef. & \(\mathit{p}\)-value\\
\midrule
Posterior coefficient
& \(1.978\) & \(<.001\)
& \(1.879\) & \(<.001\)
& \(.847\) & \(<.001\)
& \(.710\) & \(<.001\)\\
& \((.340)\) &
& \((.375)\) &
& \((.133)\) &
& \((.131)\) &\\
\addlinespace
Far Means--High Var.
& \(-.151\) & \(.739\)
& \(-.152\) & \(.758\)
& \(-.043\) & \(.821\)
& \(-.024\) & \(.890\)\\
& \((.453)\) &
& \((.491)\) &
& \((.188)\) &
& \((.172)\) &\\
\addlinespace
Close Means--Low Var.
& \(-.401\) & \(.438\)
& \(-.893\) & \(.082\)
& \(-.416\) & \(.025\)
& \(-.397\) & \(.027\)\\
& \((.518)\) &
& \((.514)\) &
& \((.186)\) &
& \((.180)\) &\\
\addlinespace
Close Means--High Var.
& \(.148\) & \(.791\)
& \(.105\) & \(.844\)
& \(-.012\) & \(.952\)
& \(-.030\) & \(.869\)\\
& \((.557)\) &
& \((.533)\) &
& \((.191)\) &
& \((.180)\) &\\
\midrule
Condition--airline fixed effects
& \multicolumn{2}{c}{\checkmark}
& \multicolumn{2}{c}{\checkmark}
& \multicolumn{2}{c}{\checkmark}
& \multicolumn{2}{c}{\checkmark}\\
Participant--route choice sets
& \multicolumn{2}{c}{\checkmark}
& \multicolumn{2}{c}{\checkmark}
& \multicolumn{2}{c}{\checkmark}
& \multicolumn{2}{c}{\checkmark}\\
Choice sets
& \multicolumn{2}{c}{\(2{,}745\)}
& \multicolumn{2}{c}{\(2{,}745\)}
& \multicolumn{2}{c}{\(2{,}745\)}
& \multicolumn{2}{c}{\(2{,}745\)}\\
Alternative-level observations
& \multicolumn{2}{c}{\(8{,}235\)}
& \multicolumn{2}{c}{\(8{,}235\)}
& \multicolumn{2}{c}{\(8{,}235\)}
& \multicolumn{2}{c}{\(8{,}235\)}\\
Participants (clusters)
& \multicolumn{2}{c}{\(305\)}
& \multicolumn{2}{c}{\(305\)}
& \multicolumn{2}{c}{\(305\)}
& \multicolumn{2}{c}{\(305\)}\\
\bottomrule
\end{tabular}
\endgroup
\caption{Conditional-logit regression results for the first airline choice
on routes \(2,\ldots,10\). The first two specifications estimate Equation~\ref{web-eq:choice_path_posterior_mean_logit}, and the final two estimate
Equation~\eqref{web-eq:choice_path_posterior_best_logit}. Participant-clustered
standard errors are in parentheses. The \(\mathit{p}\)-values on the coefficient rows are from two-sided
Wald tests. The posterior
coefficient represents the relationship in the baseline Far Means--Low
Var.\ condition; the remaining rows report interactions. All specifications
include condition-specific airline fixed effects and condition on the
participant-route choice set.}
\label{web-tab:choice_path_posterior_logit}
\end{table}
\FloatBarrier

Table~\ref{web-tab:choice_path_posterior_logit} shows that, in the baseline Far Means--Low Var.\ condition, the
cumulative posterior mean is positively associated with the participant's first choice on a new route (\(1.978\), \(\mathrm{SE}=.340\), \(\mathit{p}<.001\)). The previous route posterior mean yields a similar result (\(1.879\), \(\mathrm{SE}=.375\), \(\mathit{p}<.001\)). Relative to the baseline condition, the cumulative-posterior interaction estimates are \(-.151\) for Far Means--High Var.\ (\(\mathit{p}=.739\)), \(-.401\) for Close Means--Low Var.\ (\(\mathit{p}=.438\)),
and \(.148\) for Close Means--High Var.\
(\(\mathit{p}=.791\)); the corresponding previous-route interaction estimates are \(-.152\) (\(\mathit{p}=.758\)),
\(-.893\) (\(\mathit{p}=.082\)), and
\(.105\) (\(\mathit{p}=.844\)). These estimates indicate that the relationship between posterior means and route entry choices is broadly similar across the four experimental environments.

The posterior-best specifications show a positive baseline relationship, too. When an airline belongs to the cumulative posterior-best set, the odds that the
participant chooses it first are \(2.332\) times the odds for an airline outside that set (\(.847\), \(\mathrm{SE}=.133\), \(\mathit{p}<.001\)). When the posterior uses only the immediately preceding route, the corresponding odds ratio is \(2.034\) (\(.710\), \(\mathrm{SE}=.131\),
\(\mathit{p}<.001\)). Relative to the baseline condition, the Close Means--Low Var.\ interaction estimates are \(-.416\) (\(\mathit{p}=.025\)) in the cumulative specification and \(-.397\)
(\(\mathit{p}=.027\)) in the previous-route specification, indicating a weaker posterior-best relationship in that condition. The cumulative interaction estimates for Far Means--High Var.\ and Close Means--High Var.\ are \(-.043\) (\(\mathit{p}=.821\)) and \(-.012\) (\(\mathit{p}=.952\)), respectively; the corresponding previous-route
interaction estimates are \(-.024\) (\(\mathit{p}=.890\)) and
\(-.030\) (\(\mathit{p}=.869\)).

These estimates complement the performance results in the main text. The main analysis shows that choices improve relative to the properties of the experimental environment. The choice-path analysis instead links the first choice on a new route to posterior summaries that use the participant's own earlier outcomes. The estimates reveal this relationship when the posterior uses
only the previous route and when it uses all earlier routes. The condition
interactions reveal no statistically detectable variation in the continuous
posterior-mean relationship across the experimental environments, whereas the discrete posterior-best relationship weakens in one condition.

\subsection{Qualitative Human-Algorithms Comparison}
\label{web-ssec:qual_human_vs_algo_comparison}
In $\S$\ref{ssec:qualitative_regret_patterns}, we present the pseudo-regret plot and the best airline selection plot of the Far Means--Low Variance experiment. In this section, we provide the plots for all experiments. Beyond the trends we mentioned in the main text, we also see that experiments with close means yield tighter pseudo-regret and best-airline selection lines. This is expected because these environments represent airlines that are more similar to one another, making the distinction more difficult.
\begin{figure}[htp!]
  \includegraphics[width=\textwidth]{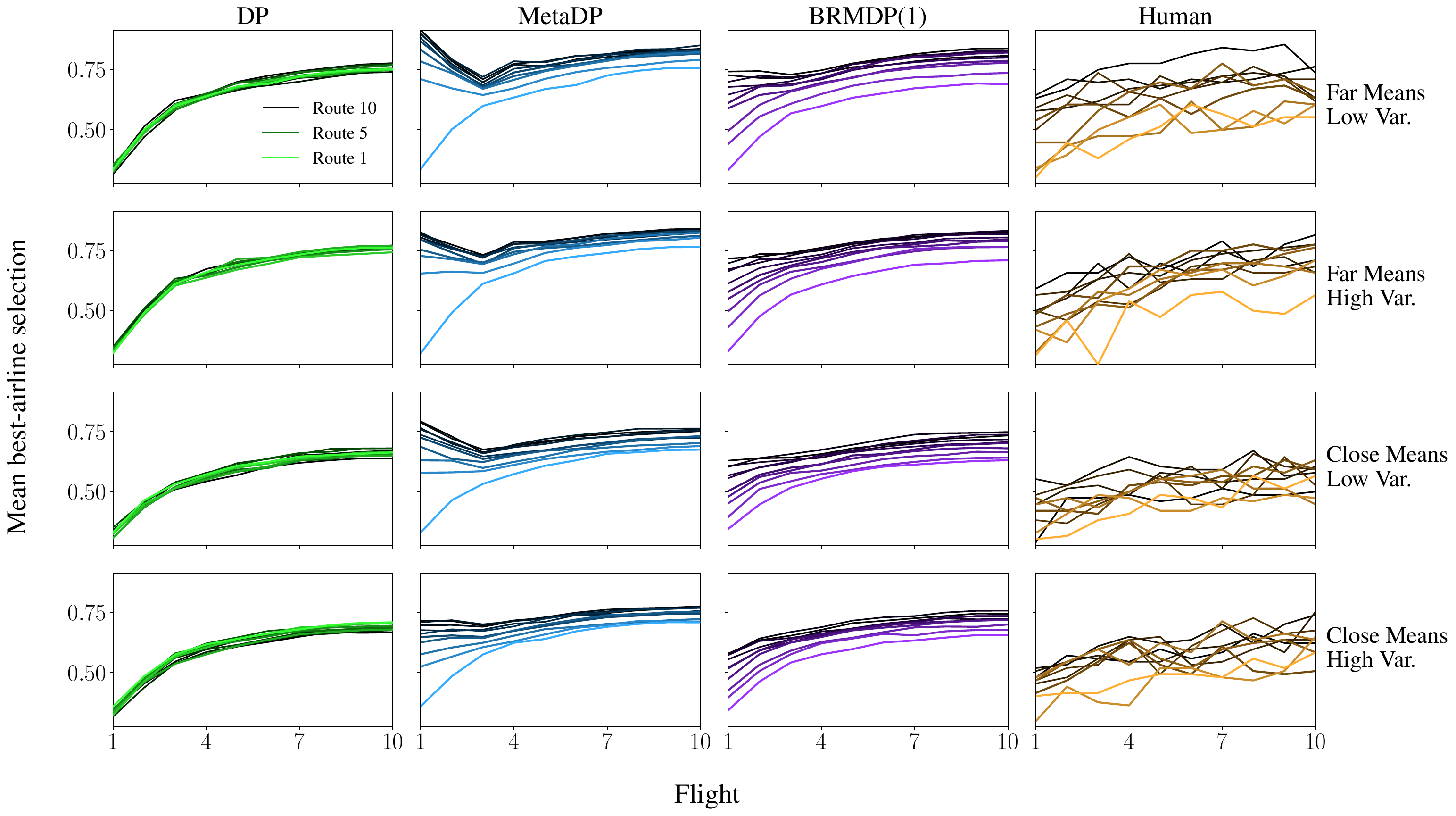}
    \caption{Mean best-airline selection over flights for different routes. Each line represents a distinct route, and earlier routes are shown in lighter shades.}
    \label{web-fig:3_qualitative_bestarm_new}
\end{figure}

\begin{figure}[htp!]
  \includegraphics[width=\textwidth]{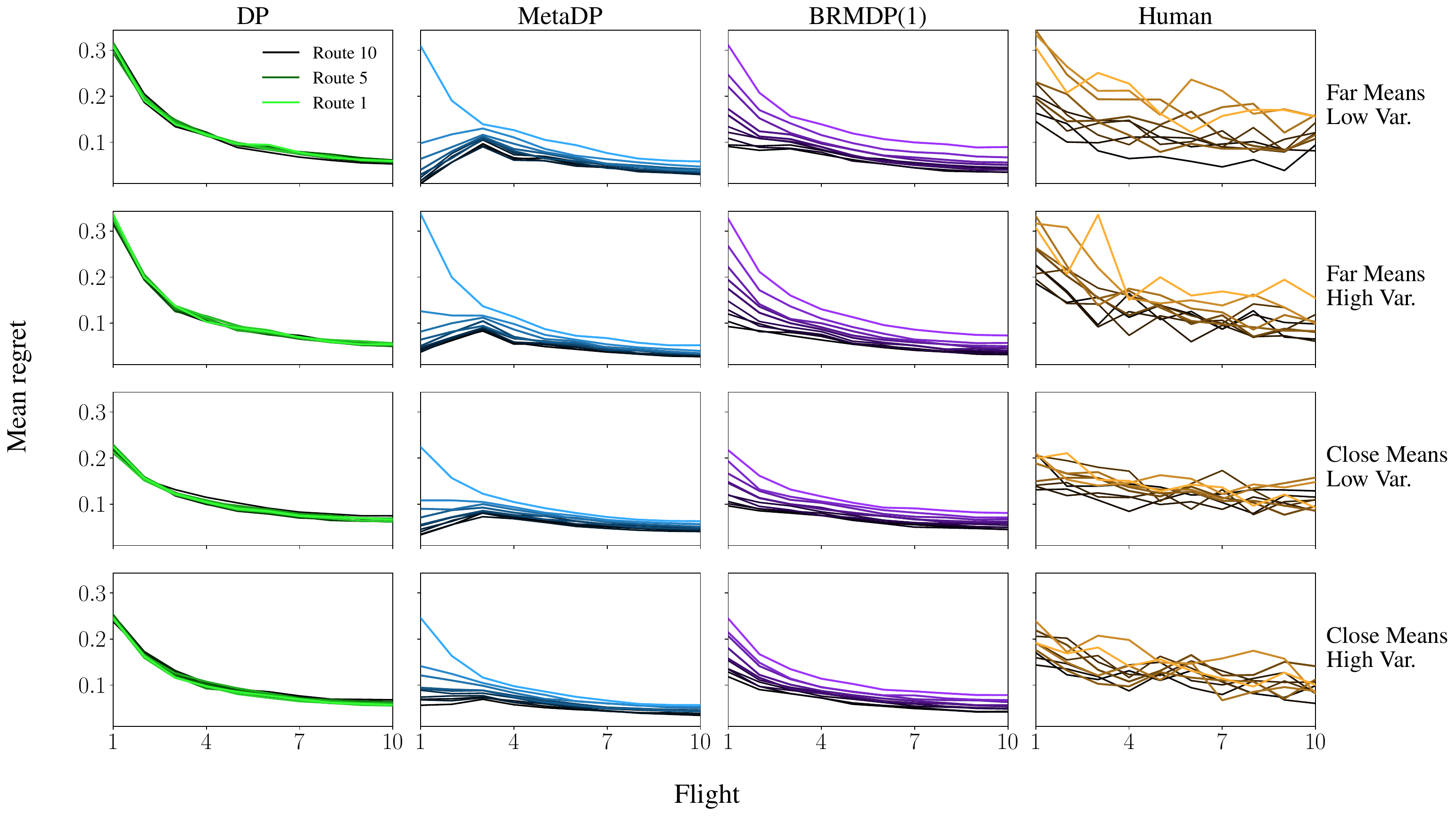}
    \caption{Mean pseudo-regret over flights for different routes. Each line represents a distinct route, and earlier routes are shown in lighter shades.}
    \label{web-fig:3_qualitative_regret_new}
\end{figure}

\subsection{Log-Likelihood Results for All Experiments}
\label{web-ssec:all_experiments_loglike}
\begin{figure}[htp!]
\centering
\includegraphics[width=.6\textwidth]{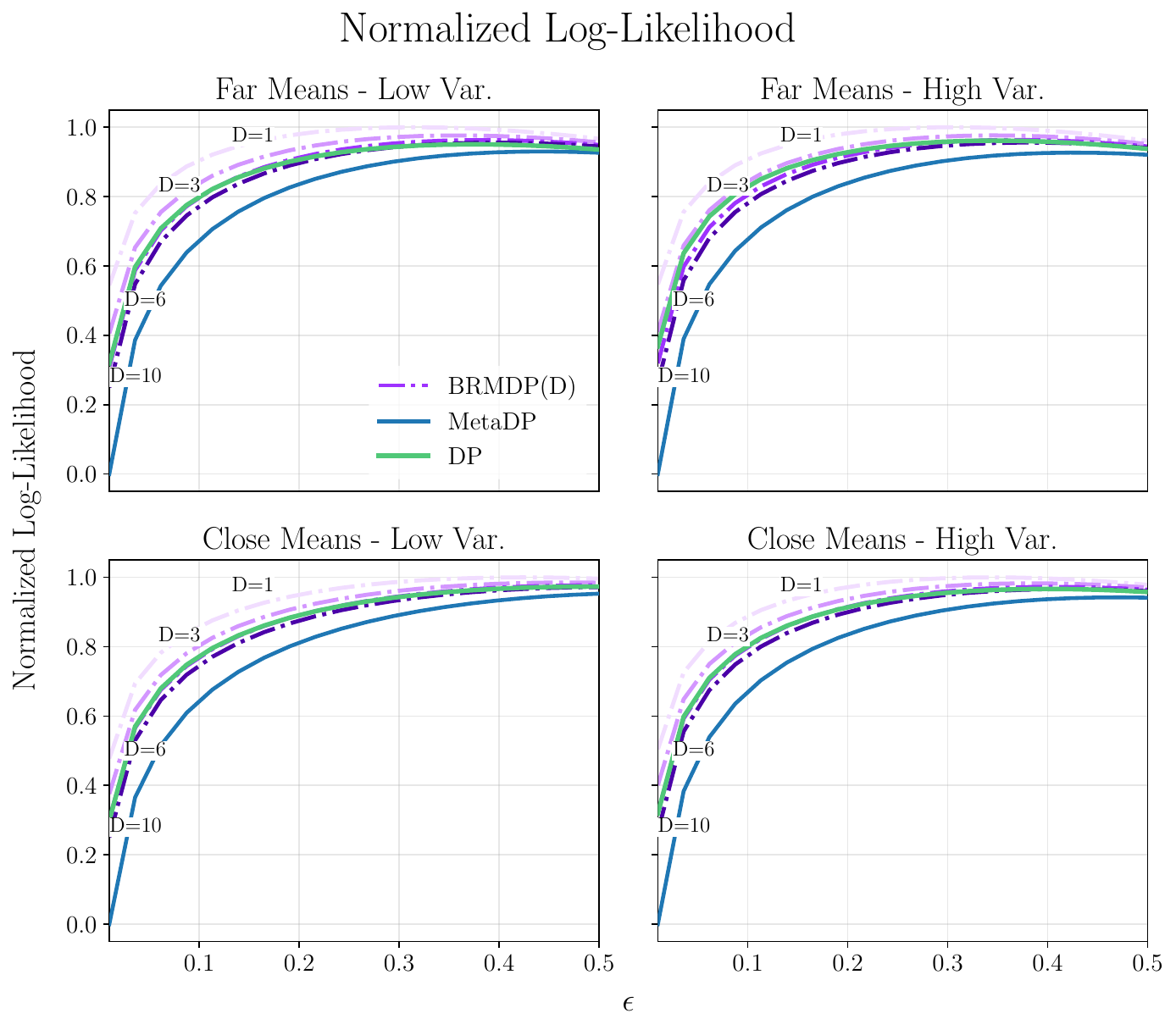}
\caption{Log-likelihood comparison across policies for \(20\) values of the noise parameter \(\epsilon\). We set \(\gamma=1\) for all policies and report BRMDP(\(D\)) fits for \(D\in\{1,3,6,10\}\). Monte Carlo likelihood calculations use \(B=2{,}000\) samples. On each chart, we normalized the log-likelihood values via $(l_. - l_{min})/(l_{max} - l_{min})$ for better visualizations and unified plots. Higher values indicate a better fit.}
\label{web-fig:dp_meta_brmdp_log_like_reordered}
\end{figure}

In $\S$\ref{ssec:dgp_identification_reordered}, we present the likelihood estimation results for one of the experiments (Far Means--Low Variance). Figure~\ref{web-fig:dp_meta_brmdp_log_like_reordered} reports results for all the experiments. We can see from the plot that the conclusions are the same as those we described in the main text.

\subsection{Robustness to Prior Means Misspecification}
\label{web-ssec:prior_misspecification}
\begin{figure}[htp!]
\centering
\includegraphics[width=.6\textwidth]{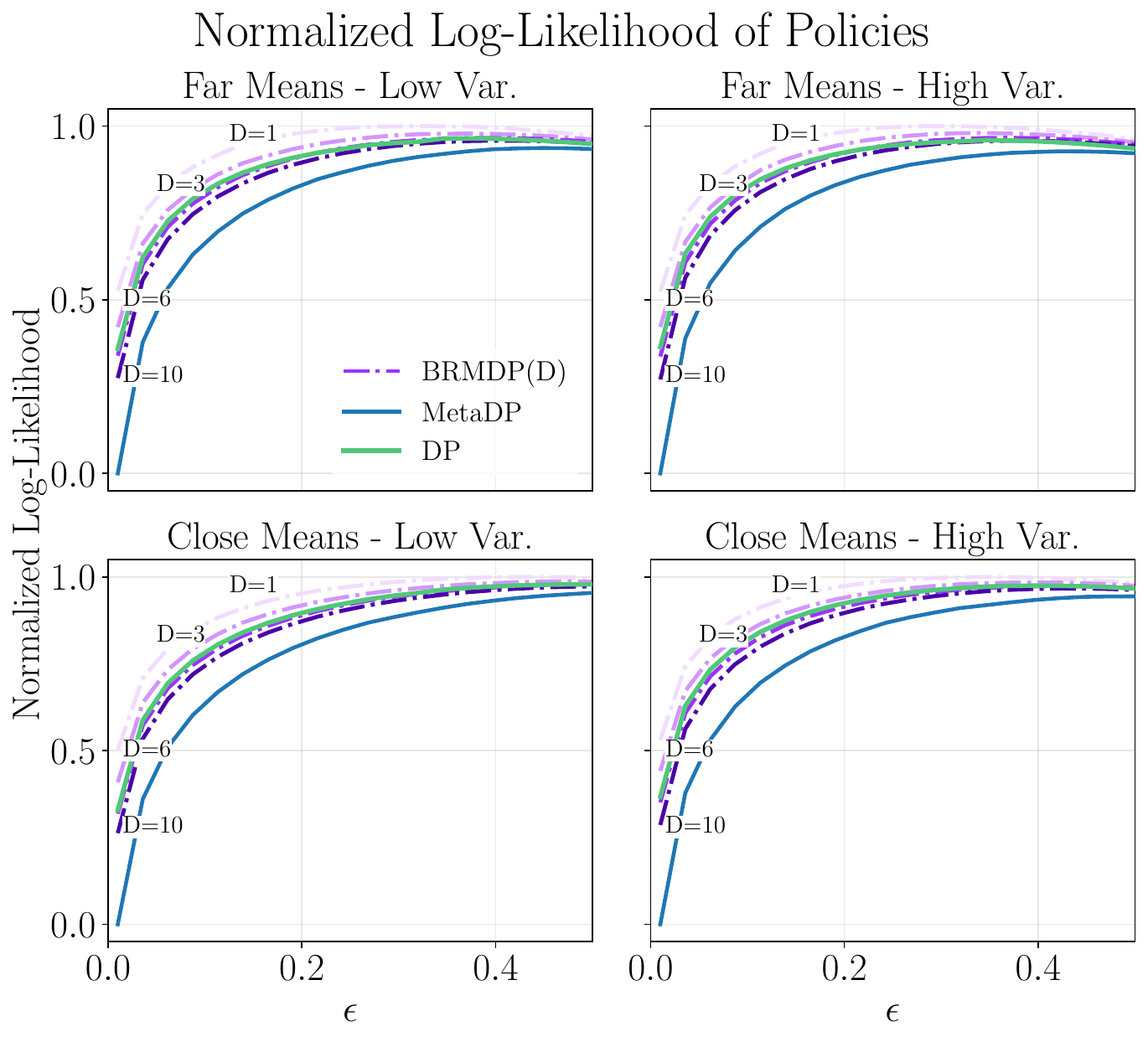}
\caption{Log-likelihood comparison across policies for \(20\) values of the noise parameter \(\epsilon\), with misspecified prior means. We set \(\gamma=1\) for all policies and report BRMDP(\(D\)) fits for \(D\in\{1,3,6,10\}\). Monte Carlo likelihood calculations use \(B=2{,}000\) samples. On each chart, we normalized the log-likelihood values via $(l_. - l_{min})/(l_{max} - l_{min})$ for better visualizations and unified plots. Higher values indicate a better fit.}
\label{web-fig:dp_meta_brmdp_log_like_reordered_mis}
\end{figure}

In $\S$\ref{ssec:dgp_identification_reordered}, we assumed that the prior class contains the true environment's prior means by including all the prior means from our experiments. Accordingly, we use \(
\{.2,.4,.5,.6,.8\},
\) as the set of participants' prior means. Thus, the candidate classes always include the true prior means. In this section, we show the robustness of the results by using misspecified prior means in our estimations. We acknowledge that there are infinitely many misspecified prior classes we could use here; nevertheless, we present results for one class, in which we simply swap the prior means between experiments. In this way, we use \{.4,.6,.8\} for the Far Means - Low/High variance experiments where the environment's means are \(\{.2,.5,.8\}\), and vice versa. Figure \ref{web-fig:dp_meta_brmdp_log_like_reordered_mis} reports the estimation results for these new prior classes. As we can see, the results are directionally in line with our main results, where BRMDP(1) achieves the highest log-likelihood values, MetaDP shows the lowest fit, and the other algorithms fall between them. Additionally, among the BRMDP class of algorithms, the more a policy performs optimally in solving the airline choice problem, the poorer it fits the behavioral data.

\subsection{Planning-Horizon Robustness}
\label{web-ssec:planning_horizon_robustness}
In the main text, all policies are solved over the full route horizon. As a supplementary robustness check, we also report the results from four experiments under the alternative truncation cases \(h=2\) and \(h=5\).
\begin{figure}[htp!]
\centering
\begin{subfigure}[t]{.49\textwidth}
\centering
\includegraphics[width=\textwidth]{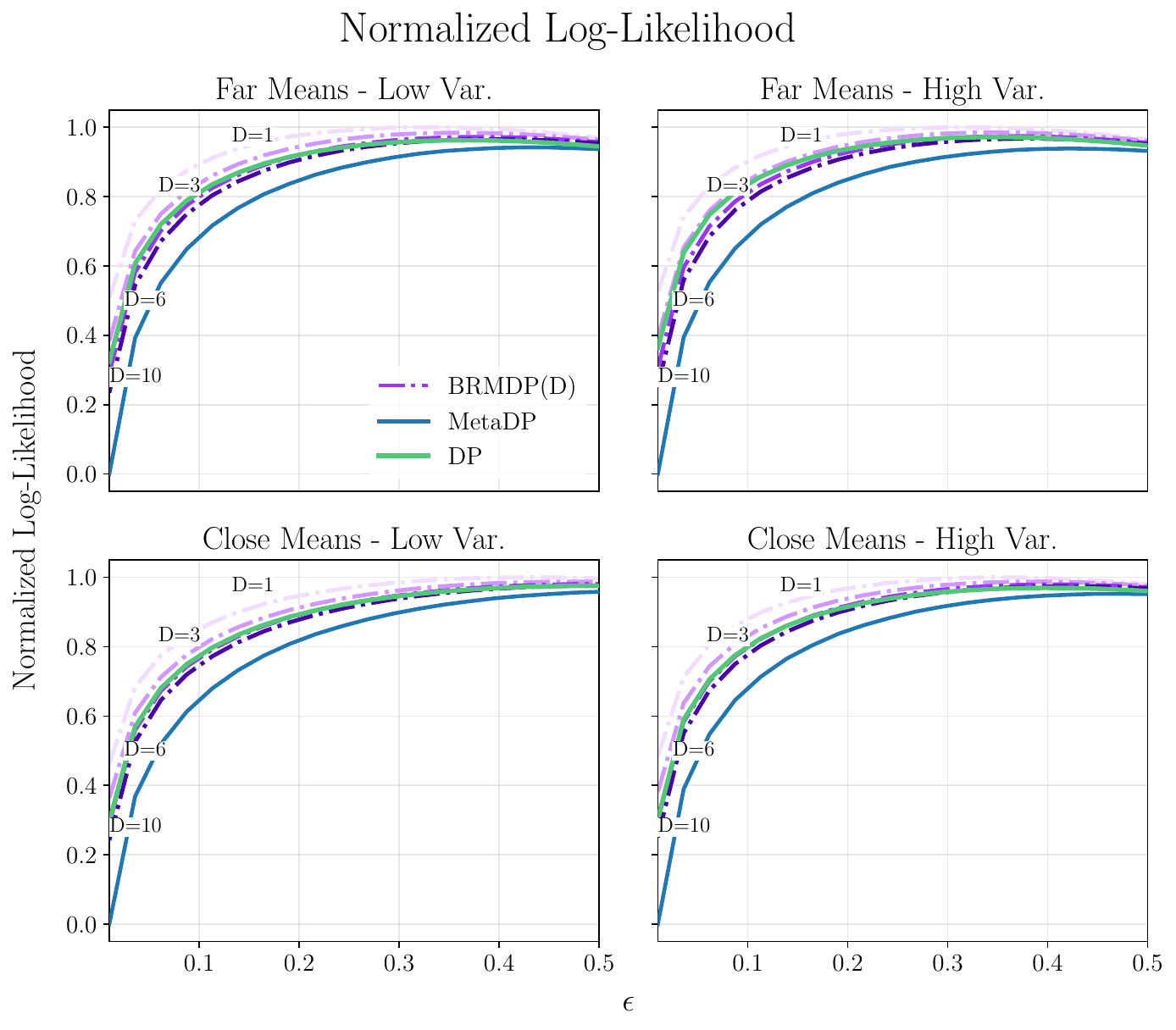}
\caption{Robustness check with \(h=2\).}
\label{web-fig:dp_meta_log_like_h2_reordered}
\end{subfigure}
\hfill
\begin{subfigure}[t]{.49\textwidth}
\centering
\includegraphics[width=\textwidth]{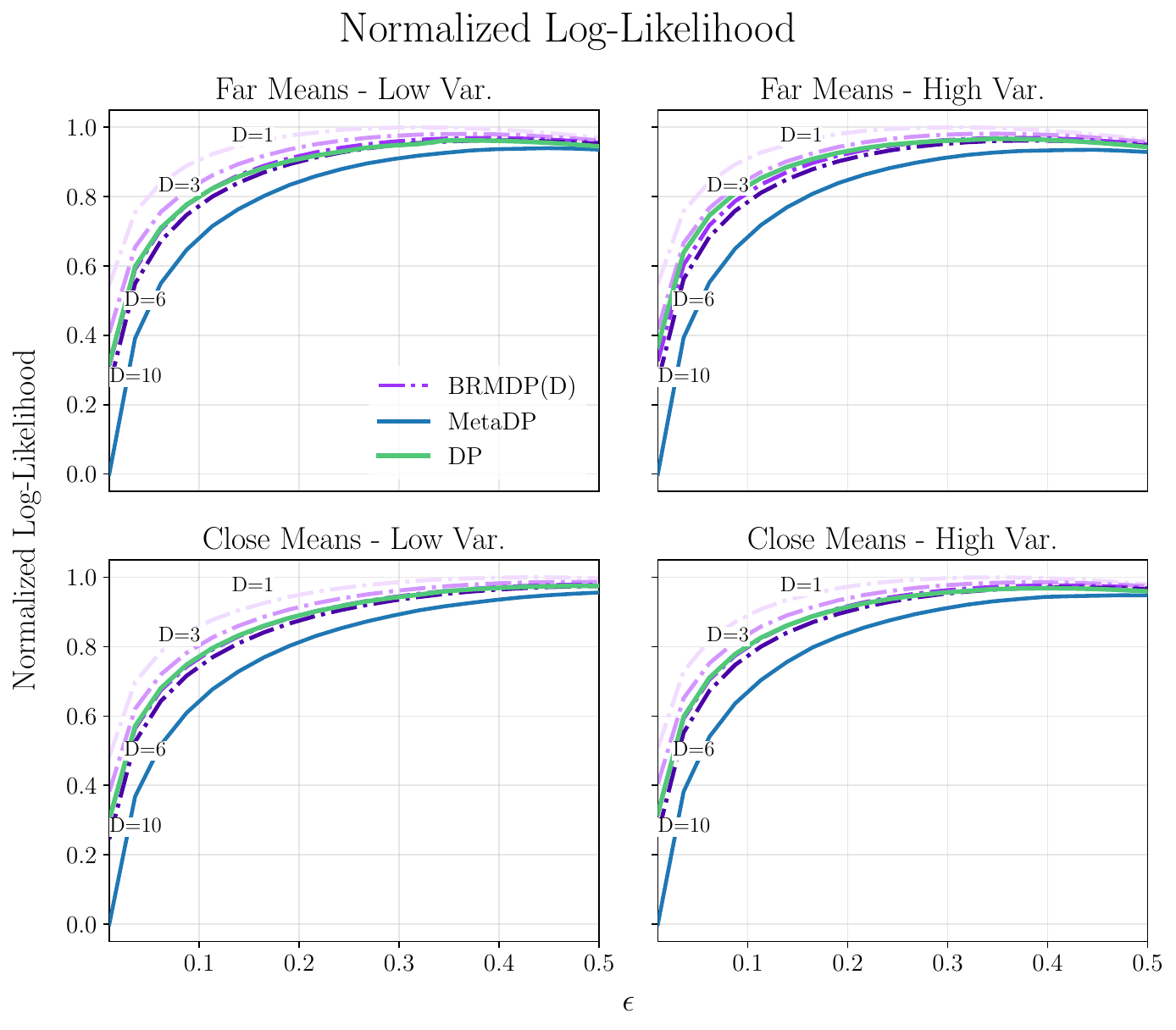}
\caption{Robustness check with \(h=5\).}
\label{web-fig:dp_meta_log_like_h5_reordered}
\end{subfigure}
\caption{Four-experiment grids of normalized log-likelihood over \(\epsilon\). The backward induction truncated at \(h=2\) and \(h=5\), respectively.}
\label{web-fig:log_like_h_robustness_reordered}
\end{figure}

\subsection{Softmax Specification}
\label{web-ssec:softmax_spec}

Throughout the paper, we use the \(\epsilon\)-greedy rule to map action values into choice probabilities. As a robustness check, we also estimate the same policies under a softmax rule. The immediate payoff, posterior updating, and backward-induction logic remain the same. The only change is that, instead of using the \(\epsilon\)-greedy state value, the continuation value is computed using a softmax-weighted average of the action values.

For example, in MetaDP, once we compute the integrated action values \(\bar{V}_{\tau}\!\bigl(k \mid \mathbf{N}_m^+,\mathbf{N}_m^-,Q_m;\mathcal{P}\bigr)\), the softmax rule assigns choice probabilities according to
\[
\pi^{MetaDP}_{\tau}\!\bigl(\tilde{k} \mid \mathbf{N}_m^+,\mathbf{N}_m^-,Q_m;\mathcal{P}\bigr)
=
\frac{\exp\!\left(\bar{V}_{\tau}\!\bigl(\tilde{k} \mid \mathbf{N}_m^+,\mathbf{N}_m^-,Q_m;\mathcal{P}\bigr)/\tau\right)}
{\sum_{k'=1}^{K}\exp\!\left(\bar{V}_{\tau}\!\bigl(k' \mid \mathbf{N}_m^+,\mathbf{N}_m^-,Q_m;\mathcal{P}\bigr)/\tau\right)},
\]
where $\tau$ is the temperature parameter. In this way, the induced state value is
\[
\bar{W}_{\tau}\!\bigl(\mathbf{N}_m^+,\mathbf{N}_m^-,Q_m;\mathcal{P}\bigr)
=
\sum_{k=1}^{K}
\pi^{MetaDP}_{\tau}\!\bigl(k \mid \mathbf{N}_m^+,\mathbf{N}_m^-,Q_m;\mathcal{P}\bigr)\,
\bar{V}_{\tau}\!\bigl(k \mid \mathbf{N}_m^+,\mathbf{N}_m^-,Q_m;\mathcal{P}\bigr).
\]
Thus, compared with the \(\epsilon\)-greedy specification, the action-value recursion changes only through this continuation term: the policy averages the future action values using softmax probabilities rather than the uniform choice-noise probabilities of the \(\epsilon\)-greedy rule. The same substitution applies to DP and BRMDP, replacing \(V_{\epsilon}\), \(\bar{V}_{\epsilon}\), and \(\tilde{V}_{\epsilon}\) by their \(\tau\)-indexed counterparts.

\begin{figure}[htp!]
\centering
\includegraphics[width=.8\textwidth]{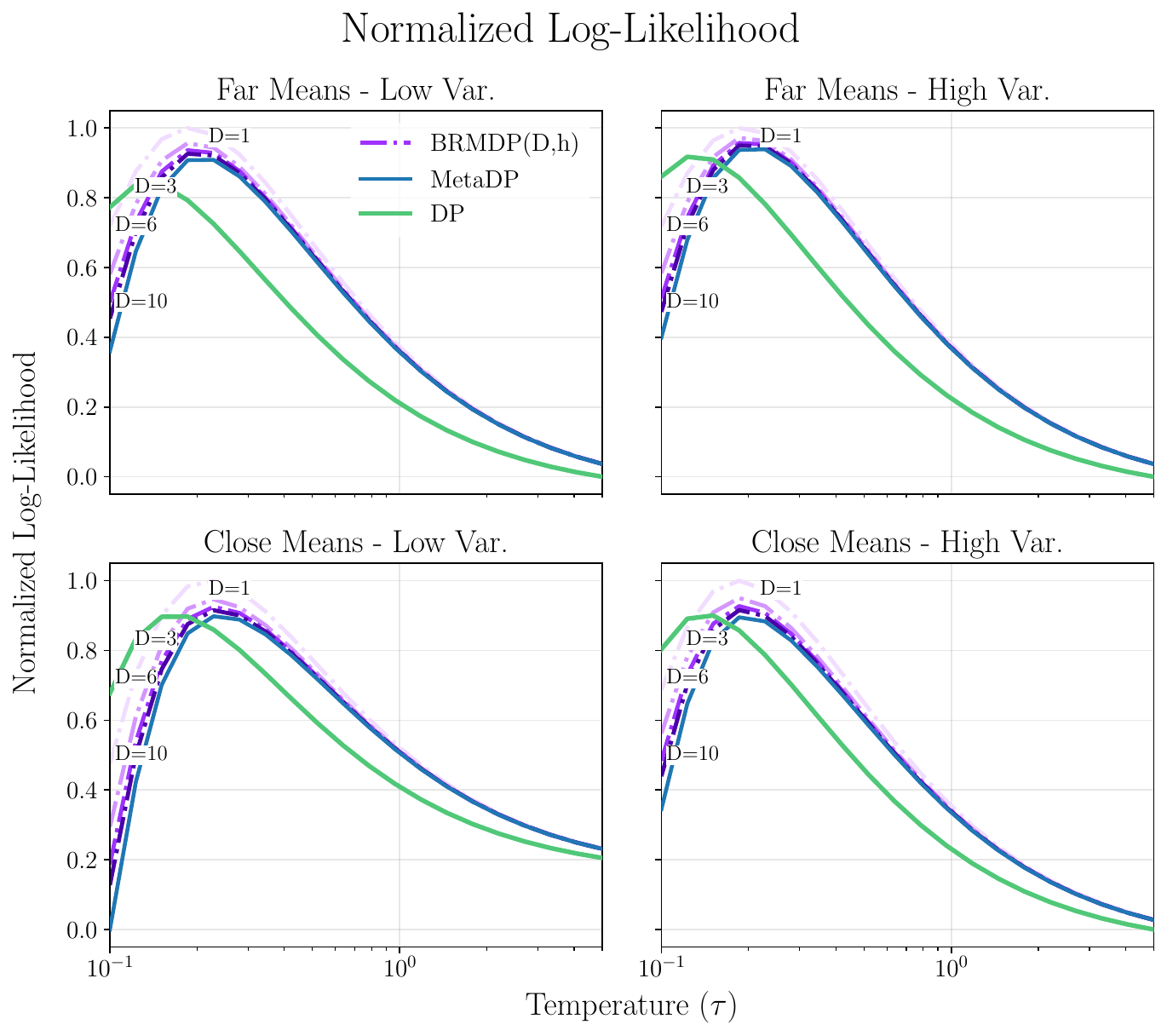}
\caption{Log-likelihood comparison across policies for \(20\) values of the temperature parameter \(\tau\). We set \(\gamma=1\) for all policies and report BRMDP(\(D\)) fits for \(D\in\{1,3,6,10\}\). Monte Carlo likelihood calculations use \(B=2{,}000\) samples. On each chart, we normalized the log-likelihood values via $(l_. - l_{min})/(l_{max} - l_{min})$ for better visualizations and unified plots. Higher values indicate a better fit.}
\label{web-fig:dp_meta_log_like_temperature_reordered}
\end{figure}

In the softmax specification, \(\tau\) governs the noise in the policy. Accordingly, as \(\tau \rightarrow 0\), the decision rule becomes deterministic, choosing the highest-value airline. In contrast, as \(\tau \rightarrow \infty\), the decision rule becomes a uniform one choosing airlines with equal probability. Accordingly, instead of plotting log-likelihood over a grid of \(\epsilon\) values, Figure~\ref{web-fig:dp_meta_log_like_temperature_reordered} plots the results over a grid of temperature values between $.1$ and $5$. The qualitative conclusion is unchanged: BRMDP\((1)\) remains the best-fitting specification, and BRMDP agents with higher $D$ values perform worse. As $D$ increases, the BRMDP performance approaches that of our most optimal policy, MetaDP. The only difference between the results here and the ones in \ref{ssec:dgp_identification_reordered} is that MetaDP outperforms DP for a wide range of $\tau$ values.

\subsection{Out-of-Sample Validation}
\label{web-ssec:out_of_sample_predictive_validation}

We next assess whether BRMDP(1)'s superior fit generalizes out of sample, both to held-out participants and to choices on future routes. We focus on the three policies central to our analysis: DP, BRMDP(1), and MetaDP. In both validation exercises, we estimate the exact likelihood of BRMDP(1). When \(D=1\), the finite sum in Equation~\eqref{web-eq:round_like_counts_meta_reordered} contains only the \(J=125\) possible hypothesis draws, making full enumeration feasible. For each validation method and experimental condition, we first report the mean within-participant log-likelihood difference, with participant-bootstrap 95\% confidence intervals. We then examine the participant-level distribution using normalized per-choice likelihood shares.

\subsubsection{Leave-One-Participant-Out Prediction}
\label{web-sssec:leave_one_participant_out_validation}

Our first validation method compares the policies’ ability to describe the choices of held-out participants within each experimental condition. For each policy and each leave-one-out participant, we estimate the \(\epsilon\) that maximizes the aggregate log-likelihood among all remaining participants. We then evaluate the selected policy on the held-out participant's complete observed choice sequence. Repeating this procedure for every participant ensures that each participant's predictive log-likelihood is obtained using a value of \(\epsilon\) without incorporating that participant's data.

\begin{figure}[htp!]
\centering
\includegraphics[width=\textwidth]{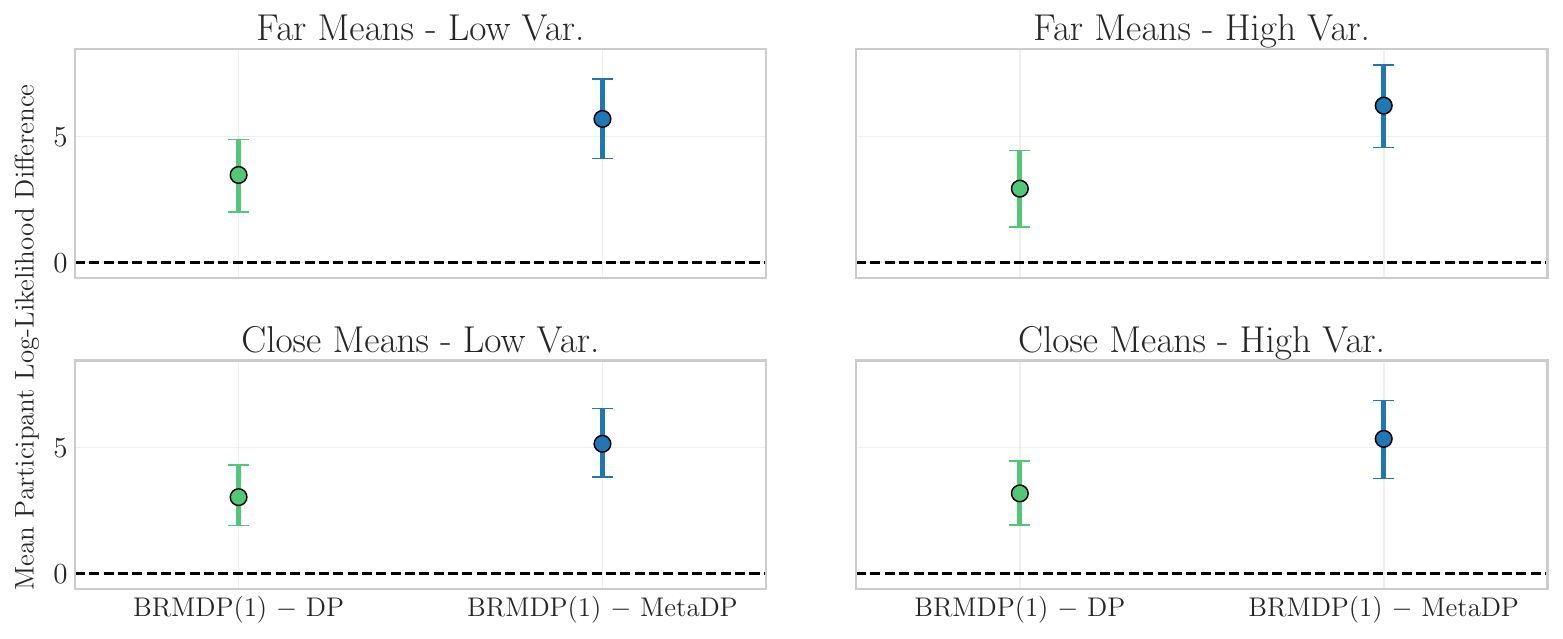}
\caption{Leave-one-participant-out predictive log-likelihood differences. Points show the mean within-participant log-likelihood difference between BRMDP(1) and the indicated benchmark. Error bars are paired participant-bootstrap 95\% confidence intervals based on \(10{,}000\) resamples. The dashed line denotes equal predictive fit; positive values favor BRMDP(1).}
\label{web-fig:leave_one_participant_out_predictive_fit}
\end{figure}

Figure~\ref{web-fig:leave_one_participant_out_predictive_fit} reports the resulting within-participant paired log-likelihood differences. In every experimental condition, the mean differences between BRMDP(1) and both DP and MetaDP are positive, and the corresponding 95\% confidence intervals lie entirely above zero (\(\mathit{p}<.001\) for all mean differences). This result shows the overall superiority of BRMDP(1) over both DP and MetaDP in predicting participants' choices using an out-of-sample approach.

\paragraph{Participant-level distribution of predictive fit.}
The paired mean differences in Figure~\ref{web-fig:leave_one_participant_out_predictive_fit} establish whether BRMDP(1) predicts held-out participants better {\it on average}. However, an average difference alone does not reveal whether this advantage is broadly distributed across participants or is generated by a relatively small number of participants with large likelihood improvements. Thus, we also examine the participant-level distribution of predictive fit.

For participant \(i\), let \(\ell_{DP,i}\), \(\ell_{MetaDP,i}\), and \(\ell_{BRMDP,i}^{(1)}\) denote the leave-one-participant-out predictive log-likelihoods under DP, MetaDP, and BRMDP(1), respectively. Because each held-out participant makes \(T\) choices on each of the \(M\) routes, we can calculate the corresponding per-choice log-likelihoods by dividing these quantities by \(MT\). We then calculate the normalized BRMDP(1) likelihood shares separately for the two benchmark comparisons (DP and MetaDP). Let \(\rho_{i,\mathrm{Benchmark}}\) denote BRMDP(1)'s normalized per-choice predictive-likelihood share for participant \(i\) relative to the indicated benchmark. We define
\begin{equation}
\label{web-eq:normalized_predictive_likelihood_share}
\begin{aligned}
\rho_{i,\mathrm{Benchmark}}
&\equiv
\frac{\exp\!\left(\ell_{BRMDP,i}^{(1)}/(MT)\right)}
{\exp\!\left(\ell_{BRMDP,i}^{(1)}/(MT)\right)
+\exp\!\left(\ell_{\mathrm{Benchmark},i}/(MT)\right)}, \\
&\qquad \mathrm{Benchmark}\in\{\mathrm{DP},\mathrm{MetaDP}\}.
\end{aligned}
\end{equation}

Each expression exponentiates the two per-choice log-likelihoods and normalizes them to sum to one. Consequently, a value of \(.5\) indicates that BRMDP(1) and the relevant benchmark assign the same per-choice predictive likelihood to participant \(i\). Values above \(.5\) favor BRMDP(1), whereas values below \(.5\) favor the benchmark. The per-choice normalization prevents the resulting shares from being driven mechanically toward zero or one by the number of evaluated choices.

\begin{figure}[htp!]
\centering
\includegraphics[width=\textwidth]{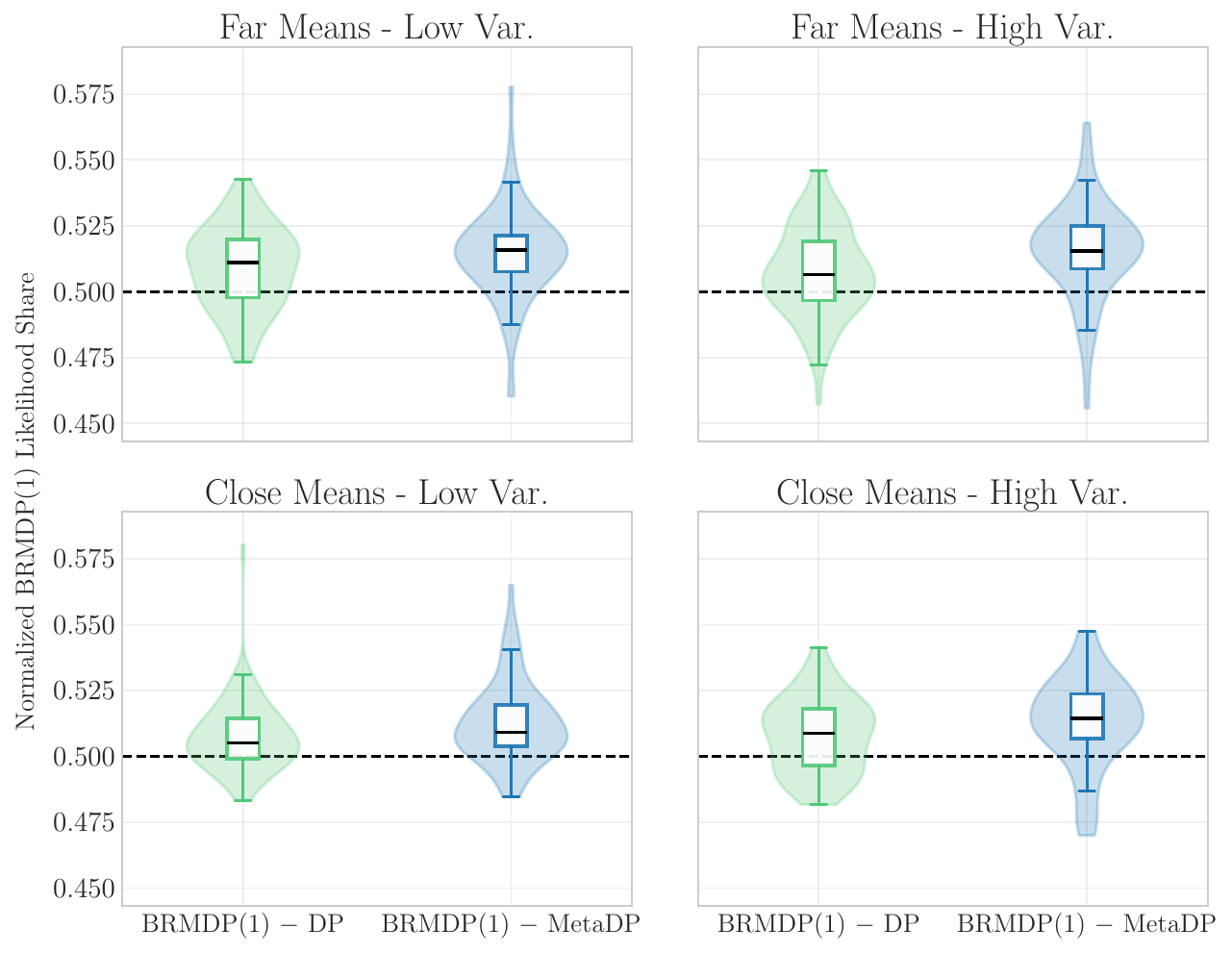}
\caption{Participant-level distributions of normalized predictive-likelihood shares under leave-one-participant-out validation. Each violin represents the distribution across held-out participants within one experimental condition. The embedded boxplot reports the median, interquartile range, and conventional \(1.5\)-IQR whiskers. The dashed line at \(.5\) denotes equal predictive likelihood; values above \(.5\) favor BRMDP(1) over the indicated benchmark.}
\label{web-fig:leave_one_participant_out_participant_likelihood}
\end{figure}

Figure~\ref{web-fig:leave_one_participant_out_participant_likelihood} shows that the median normalized likelihood share lies above \(.5\) for both comparisons in every experimental condition. The distributions relative to MetaDP are shifted more clearly above \(.5\), while those relative to DP are generally closer to, but still above, the equal-fit line. Nevertheless, the median of each distribution favors BRMDP(1). Thus, the positive mean differences in Figure~\ref{web-fig:leave_one_participant_out_predictive_fit} do not result solely from a small number of participants with unusually large likelihood gains. Instead, BRMDP(1)'s predictive advantage is present across a broad portion of the participant distribution. At the same time, the mass below \(.5\) indicates meaningful heterogeneity among participants.

\subsubsection{Validation on Future Routes}
\label{web-sssec:future_route_validation}

Our second out-of-sample validation method compares the policies’ ability to describe choices on future routes. Within each experimental condition and for each participant, we estimate \(\epsilon\) separately for DP, BRMDP(1), and MetaDP using choices from the early routes \(1\)--\(5\). We then hold the estimated values fixed and evaluate the policies on choices from later routes \(6\)--\(10\). The likelihoods for these later routes remain conditional on each participant's realized earlier-route history, consistent with the cross-route updating process defined in the main analysis.

\begin{figure}[htp!]
\centering
\includegraphics[width=\textwidth]{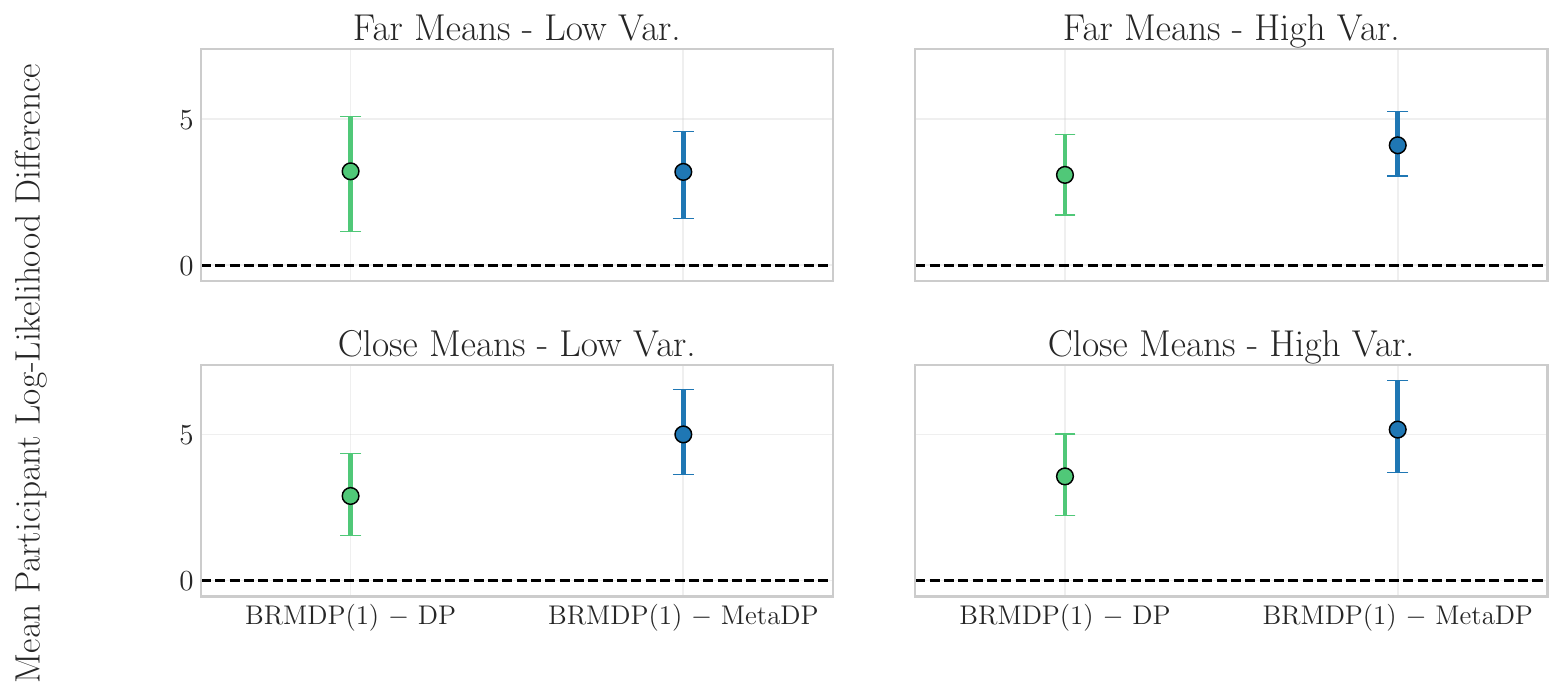}
\caption{Predictive log-likelihood differences on future routes. Points show the mean within-participant log-likelihood difference between BRMDP(1) and the indicated benchmark. Error bars are paired participant-bootstrap 95\% confidence intervals based on \(10{,}000\) resamples. The dashed line denotes equal predictive fit; positive values favor BRMDP(1).}
\label{web-fig:future_routes_predictive_fit}
\end{figure}

Figure~\ref{web-fig:future_routes_predictive_fit} shows that BRMDP(1) predicts choices on routes \(6\)--\(10\) better than both DP and MetaDP in all four experimental conditions. Similar to the previous results, the mean differences between BRMDP(1) and both DP and MetaDP are positive in every experimental condition, and the corresponding 95\% confidence intervals lie entirely above zero (\(\mathit{p}<.01\) for all mean differences). Therefore, the superior fit of BRMDP(1) persists when we estimate the policy-noise parameter \(\epsilon\) exclusively from earlier routes. This result is particularly informative about the transfer mechanism: the coarse integration of prior uncertainty embodied by BRMDP(1) better predicts how participants use their accumulated experience in subsequent routes than either no transfer or fully integrated meta-learning.

\begin{figure}[htp!]
\centering
\includegraphics[width=\textwidth]{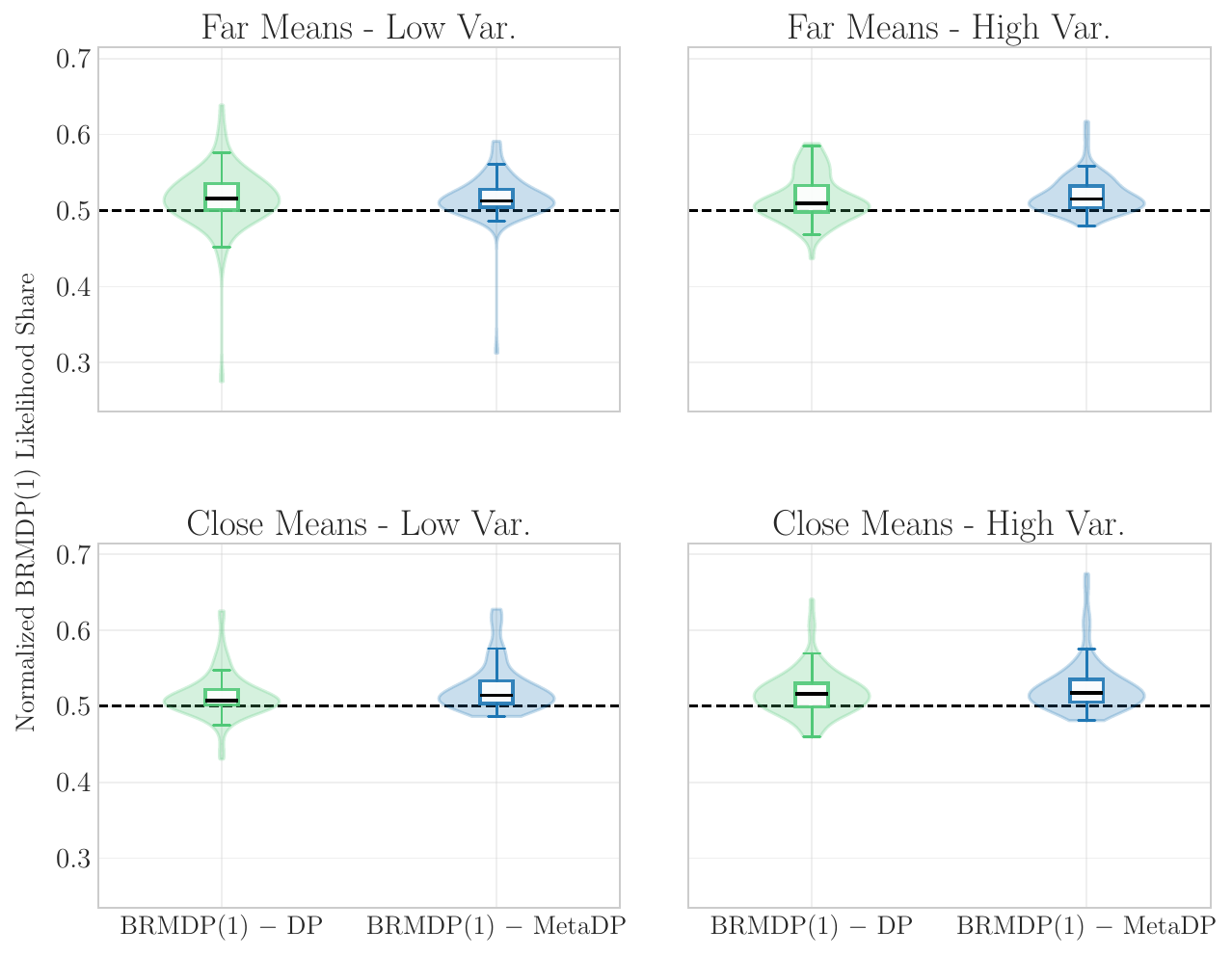}
\caption{Participant-level distributions of normalized predictive-likelihood shares on future routes. Each violin represents the distribution across participants, and the embedded boxplot reports the median, interquartile range, and conventional \(1.5\)-IQR whiskers. The dashed line at \(.5\) denotes equal predictive likelihood; values above \(.5\) favor BRMDP(1) over the indicated benchmark.}
\label{web-fig:future_routes_participant_likelihood}
\end{figure}

\paragraph{Participant-level distribution of predictive fit.}
Figure~\ref{web-fig:future_routes_participant_likelihood} presents the participant-level distributions corresponding to the future-route mean differences in Figure~\ref{web-fig:future_routes_predictive_fit}. We use the same normalization defined above, but calculate the predictive log-likelihoods using routes \(6\)--\(10\) and divide them by \(MT/2\), the number of choices evaluated for each participant.

The participant-level distributions reinforce the conclusions from the paired mean differences. In every experimental condition, the median normalized likelihood share is above \(.5\) relative to both DP and MetaDP. The distributions relative to both DP and MetaDP are again clearly shifted above the equal-fit line. Therefore, BRMDP(1)'s advantage in predicting future-route choices is not driven solely by a small set of participants with large improvements. It appears across a broad portion of the participant distribution, although the density below \(.5\) again shows that the preferred policy varies for some participants. Taken together with the positive paired mean differences, these results show that the predictive advantage of boundedly rational meta-learning persists both on average and across participants' later-route behavior.

\subsection{No-Transfer DP with Improving Execution (DPIE)}
\label{web-ssec:improving_execution_robustness}

Across-route improvement could arise without belief transfer if participants
become more familiar with the task or execute their choices more consistently.
The baseline DP uses the same choice-noise parameter on every route, so it cannot
represent this form of generic improvement. We therefore consider a no-transfer
DP whose execution improves across routes.

Let \(\epsilon_m\) denote the choice-noise parameter on route \(m\), and denote
the resulting policy by DPIE. The policy uses the same DP action values and the
same choice rule as Equation~\eqref{eq:e_greedy_probability_func_reordered}, but
allows choice noise to decline across routes:
\begin{equation}
    \pi_m^{\mathrm{DP\text{-}IE}}\!\left(\,\cdot\mid s_t^{DP}\right)
    \equiv
    \pi_{\epsilon_m}^{DP}\!\left(\,\cdot\mid s_t^{DP}\right),
    \qquad
    \epsilon_1 \geq \epsilon_2 \geq \cdots \geq \epsilon_M.
    \label{web-eq:dp_improving_execution_policy}
\end{equation}
Every route still begins from the independent \(\operatorname{Beta}(1,1)\)
priors of the baseline DP, and belief updating uses only outcomes from the
current route. Only \(\epsilon_m\) changes across routes.

We index each noise profile by its endpoints \((\epsilon_1,\epsilon_M)\). For
each candidate pair, we determine the intermediate values rather than selecting
a separate choice-noise parameter for every route. Specifically, we first set
\begin{equation}
    \operatorname{logit}(\widetilde{\epsilon}_m)
    = \operatorname{logit}(\epsilon_1)
    + \frac{\log m}{\log M}
    \left[
        \operatorname{logit}(\epsilon_M)
        - \operatorname{logit}(\epsilon_1)
    \right],
    \qquad
    \operatorname{logit}(x)=\log\!\left(\frac{x}{1-x}\right),
    \label{web-eq:dp_improving_execution_interpolation}
\end{equation}
and assign \(\epsilon_m\) to the evaluated grid value closest to
\(\widetilde{\epsilon}_m\). Thus, the values between the two endpoints follow a
fixed path that is linear in log-odds against the log of the route number. The
grid contains 20 equally spaced values from \(.01\) to \(.50\) and 20 equally
spaced values from \(.525\) to \(.99\). The expanded range avoids restricting
the no-transfer alternative at the main grid's upper bound.

We use leave-one-participant-out (LOO) validation separately within each
experimental condition. For each omitted participant, we select the DPIE
endpoint pair that maximizes the summed log-likelihood of all other participants
in that condition. We separately select the constant \(\epsilon\) for exact
BRMDP(1) on those same training participants, and then evaluate both selected
policies on the omitted participant.

\begin{table}[htp!]
\centering
\small
\begingroup
\setlength{\tabcolsep}{5pt}
\begin{tabular}{lcccc}
\toprule
& \multicolumn{2}{c}{\textbf{DPIE Endpoints}}
& \multicolumn{2}{c}{\textbf{BRMDP(1) minus DPIE}}\\
\cmidrule(lr){2-3}\cmidrule(lr){4-5}
\textbf{Experimental Condition}
& \textbf{Mean \(\epsilon_1\)}
& \textbf{Mean \(\epsilon_M\)}
& \textbf{Mean Difference}
& \textbf{\(\mathit{p}\)-value}\\
\midrule
Far Means--Low Var.
& .507 & .340 & \(3.59\) & \(<.001\)\\
& & & \((.81)\) &\\
\addlinespace
Far Means--High Var.
& .500 & .281 & \(3.02\) & \(<.001\)\\
& & & \((.86)\) &\\
\addlinespace
Close Means--Low Var.
& .629 & .420 & \(2.39\) & \(<.001\)\\
& & & \((.67)\) &\\
\addlinespace
Close Means--High Var.
& .535 & .345 & \(3.21\) & \(<.001\)\\
& & & \((.70)\) &\\
\bottomrule
\end{tabular}
\endgroup
\caption{Leave-one-participant-out predictive comparison of BRMDP(1) and
DPIE. Bootstrap standard errors are in parentheses. Positive differences favor BRMDP(1). The endpoint columns average the LOO-selected values of
\(\epsilon_1\) and \(\epsilon_M\) across held-out-participant folds within
each condition. Standard errors and two-sided \(\mathit{p}\)-values use \(10{,}000\)
participant-level bootstrap resamples.}
\label{web-tab:dp_improving_execution_comparison}
\end{table}
\FloatBarrier

Table~\ref{web-tab:dp_improving_execution_comparison} reports the pairwise
comparisons. BRMDP(1) has a higher mean predictive log-likelihood in every
condition
(\(\mathit{p} < .001\)). The selected DPIE profile declines strictly in every fold,
while no fold selects the expanded upper boundary of \(.99\). Thus, improving
execution without belief transfer does not account for BRMDP(1)'s predictive
advantage within this policy class.

\subsection{Non-Meta-Learning Knowledge-Transfer Benchmarks}
\label{web-ssec:non_meta_transfer_robustness}
\label{web-ssec:ktmu_robustness}

Our main model comparison builds on the consumer-learning literature by using forward-looking dynamic programming and spans different degrees of rationality in cross-route knowledge transfer. Accordingly, we use meta-learning policies as a narrower set of algorithms that transfer knowledge through a bilevel approach that learns a higher-level distribution from which each route's parameters are generated. A natural alternative could transfer airline-specific information across routes without maintaining a hyper-posterior. Such a policy could pool all past outcomes or carry forward a lower-dimensional summary of them, and it could use the resulting beliefs either myopically or in a forward-looking dynamic program. If one of these simpler policies fits as well as BRMDP, the comparative-fit result would concern cross-route information transfer more generally, rather than the hyper-posterior updating and sampled integration embodied in BRMDP.

To examine these possibilities, we introduce three non-meta-learning knowledge-transfer benchmarks. The Knowledge Transfer Mean-Updating policy, denoted KTMU, pools all observations for an airline across routes and chooses according to the resulting posterior mean. This represents the simplest form of a knowledge transfer policy, which ignores the route changes and future choices. The \(\nu\)-KTMU policy separates the historical mean from the strength with which it is transferred, using \(\nu\) as an effective prior sample size, but remains myopic. Consequently, larger values of \(\nu\) give earlier-route experience more influence over current beliefs, whereas smaller values allow outcomes from the current route to receive more weight. The \(\nu\)-KTDP policy uses the same transferred prior as \(\nu\)-KTMU and introduces forward-looking choice through the finite-horizon Bellman recursion. None of the three policies maintains the finite hypothesis class \(\mathcal P\), updates the hyper-posterior \(Q_m\), or draws candidate prior hypotheses. Together, they allow us to distinguish the role of simple outcome pooling, the strength of transferred experience, and forward-looking planning from the higher-order learning mechanism in MetaDP and BRMDP.

\subsubsection{Knowledge Transfer Mean-Updating (KTMU)}
KTMU is the simplest of the three benchmarks. It treats all observations for a given airline as evidence about a single airline-specific on-time rate, pools those observations across routes, and chooses according to the resulting posterior means. KTMU therefore transfers knowledge because earlier routes affect later choices, but it is not a meta-learning policy under our narrower definition because it does not update a higher-order belief about the distribution of route-specific parameters. It is also myopic in the sense of limited-horizon models of sequential choice \citep{gabaix2006costly,tehrani2024heuristic}, current choices depend on current posterior means alone, without a continuation value.

\paragraph{Pooled mean updating.}
For each airline \(k\), KTMU begins with a uniform, uninformative
\(\mathrm{Beta}(1,1)\) prior. We define the pooled counts available before flight \(t\) on route \(m\) as
\begin{align}
\bar{N}_{k,m,t-1}^{+}
&=
\sum_{r=1}^{m-1}N_{k,r}^{+}
+
N_{k,m,t-1}^{+},
\nonumber\\
\bar{N}_{k,m,t-1}^{-}
&=
\sum_{r=1}^{m-1}N_{k,r}^{-}
+
N_{k,m,t-1}^{-}.
\label{web-eq:ktmu_pooled_counts}
\end{align}
Here, the bar distinguishes the counts pooled across routes from the
within-route counts \(N_{k,m,t-1}^{+}\) and \(N_{k,m,t-1}^{-}\). Under KTMU, the consumer behaves as if each airline has a single
route-invariant on-time rate, denoted by
\(\theta_k^{\mathrm{KTMU}}\). This perceived rate is distinct from the
environment's true route-specific rate \(\theta_{m,k}^{\ast}\). The KTMU
posterior is
\begin{equation}
\theta_k^{\mathrm{KTMU}}
\mid
\mathcal{I}_{m,t-1}
\sim
\operatorname{Beta}
\left(
1+\bar{N}_{k,m,t-1}^{+},
1+\bar{N}_{k,m,t-1}^{-}
\right),
\label{web-eq:ktmu_posterior}
\end{equation}
with posterior mean
\begin{equation}
\hat{\theta}_{k,t,m}^{\mathrm{KTMU}}
=
\frac{1+\bar{N}_{k,m,t-1}^{+}}
     {2+\bar{N}_{k,m,t-1}^{+}
        +\bar{N}_{k,m,t-1}^{-}}.
\label{web-eq:ktmu_posterior_mean}
\end{equation}

Let
\begin{equation}
\mathcal{G}_{m,t}^{\mathrm{KTMU}}
=
\arg\max_{k\in\{1,\ldots,K\}}
\hat{\theta}_{k,t,m}^{\mathrm{KTMU}}
\label{web-eq:ktmu_greedy_set}
\end{equation}
denote the set of airlines with the highest pooled posterior mean. We apply the same \(\epsilon\)-greedy choice-noise rule used in the main specification:
\begin{equation}
\pi_{\epsilon}^{KTMU}
\!\left(
\widetilde{k}
\mid
\overline{\mathbf N}_{m,t-1}^{+},
\overline{\mathbf N}_{m,t-1}^{-}
\right)
=
(1-\epsilon)
\frac{
\mathbf{1}
\left\{
\widetilde{k}\in\mathcal G_{m,t}^{KTMU}
\right\}
}{
\left|\mathcal G_{m,t}^{KTMU}\right|
}
+
\frac{\epsilon}{K}.
\label{web-eq:ktmu_choice_probability}
\end{equation}
Thus, total probability \(1-\epsilon\) is assigned to the airlines with the highest posterior mean, with ties resolved uniformly, while the remaining probability \(\epsilon\) is distributed uniformly across all \(K\) airlines. In contrast to DP, MetaDP, and BRMDP, KTMU contains no information-seeking mechanism that utilizes a continuation value. Algorithm~\ref{web-alg:ktmu_policy} summarizes the policy.

\begin{algorithm}[H]
\small
\begin{singlespace}
\caption{Knowledge Transfer Mean-Updating Policy (KTMU)}
\label{web-alg:ktmu_policy}
\begin{algorithmic}[1]
\State \textbf{Inputs:} number of airlines \(K\), routes \(M\), flights \(T\), and choice noise \(\epsilon\)
\State \textbf{Initialize once:}
\(\overline{\mathbf N}_{1,0}^{+}\gets\mathbf 0\) and
\(\overline{\mathbf N}_{1,0}^{-}\gets\mathbf 0\)
\For{\(m=1,\ldots,M\)}
    \For{\(t=1,\ldots,T\)}
        \State For every airline \(k\), compute
        \[
        \hat{\theta}_{k,t,m}^{\mathrm{KTMU}}
        \gets
        \frac{1+\bar N_{k,m,t-1}^{+}}
             {2+\bar N_{k,m,t-1}^{+}
                +\bar N_{k,m,t-1}^{-}}
        \]
        \State Set
        \[
        \mathcal G_{m,t}^{\mathrm{KTMU}}
        \gets
        \arg\max_k
        \hat{\theta}_{k,t,m}^{\mathrm{KTMU}}
        \]
        \State Compute
        \[
        \pi_{\epsilon}^{\mathrm{KTMU}}
        \!\left(
        \cdot
        \mid
        \overline{\mathbf N}_{m,t-1}^{+},
        \overline{\mathbf N}_{m,t-1}^{-}
        \right)
        \]
        using Equation~\eqref{web-eq:ktmu_choice_probability}
        \State Choose
        \[
        A_{m,t}
        \sim
        \pi_{\epsilon}^{\mathrm{KTMU}}
        \!\left(
        \cdot
        \mid
        \overline{\mathbf N}_{m,t-1}^{+},
        \overline{\mathbf N}_{m,t-1}^{-}
        \right)
        \]
        \State Observe \(y_{m,t}\in\{0,1\}\)
        \State Update
        \[
        \overline{\mathbf N}_{m,t}^{+}
        \gets
        \overline{\mathbf N}_{m,t-1}^{+}
        +
        y_{m,t}\mathbf e_{A_{m,t}}
        \]
        \State Update
        \[
        \overline{\mathbf N}_{m,t}^{-}
        \gets
        \overline{\mathbf N}_{m,t-1}^{-}
        +
        (1-y_{m,t})\mathbf e_{A_{m,t}}
        \]
    \EndFor
    \If{\(m<M\)}
        \State Carry the pooled counts into the next route:
        \[
        \overline{\mathbf N}_{m+1,0}^{+}
        \gets
        \overline{\mathbf N}_{m,T}^{+},
        \qquad
        \overline{\mathbf N}_{m+1,0}^{-}
        \gets
        \overline{\mathbf N}_{m,T}^{-}
        \]
    \EndIf
\EndFor
\end{algorithmic}
\end{singlespace}
\end{algorithm}

\begin{figure}[htp!]
\centering
\includegraphics[width=.6\textwidth]
{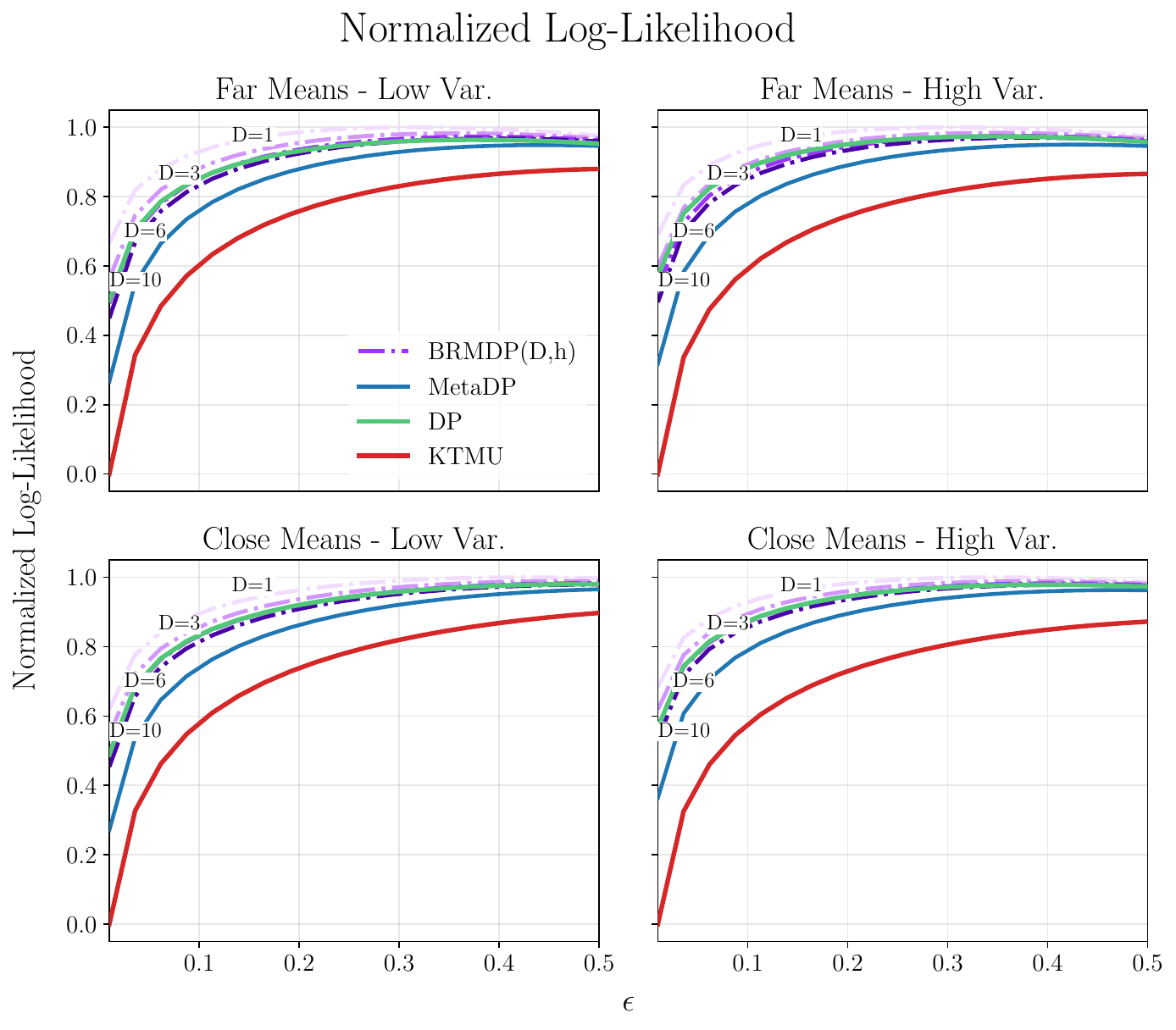}
\caption{Normalized log-likelihood comparison including the Knowledge Transfer Mean-Updating policy. Each panel corresponds to one experimental condition and reports the likelihood over \(20\) values of the choice-noise parameter \(\epsilon\). We set \(\gamma=1\) and \(h=10\) for the dynamic-programming policies and report BRMDP(\(D\)) for \(D\in\{1,3,6,10\}\). BRMDP likelihood calculations use \(B=2{,}000\) Monte Carlo samples. Within each panel, we normalize the log-likelihood as \((\ell-\ell_{\min})/(\ell_{\max}-\ell_{\min})\). Higher values indicate a better fit.}
\label{web-fig:dp_metadp_tl_ll_epsilon_normalized_h_10}
\end{figure}

\paragraph{Likelihood comparison.}
For estimation, we evaluate KTMU along each participant's observed sequence of choices and outcomes. The participant-level log-likelihood is
\begin{equation}
\ell_{\mathrm{KTMU}}(\epsilon)
=
\sum_{m=1}^{M}\sum_{t=1}^{T}
\log
\pi_{\epsilon}^{\mathrm{KTMU}}
\!\left(
A_{m,t}
\mid
\overline{\mathbf N}_{m,t-1}^{+},
\overline{\mathbf N}_{m,t-1}^{-}
\right).
\label{web-eq:ktmu_log_likelihood}
\end{equation}
We update the pooled counts using the participant's realized history. Figure~\ref{web-fig:dp_metadp_tl_ll_epsilon_normalized_h_10} adds KTMU to the likelihood comparison for all four experimental conditions. Across every condition and throughout the displayed \(\epsilon\) grid, KTMU has a lower likelihood than every other policy. It even fits worse than the no-transfer DP benchmark, despite carrying all past outcomes forward.

\FloatBarrier

\subsubsection{Shrinkage Knowledge Transfer Mean-Updating (\(\nu\)-KTMU)}
KTMU gives every earlier outcome the same weight as a current-route outcome. Consequently, the amount of information transferred to a new route grows with the number of times an airline was chosen on previous routes. This form of transfer could severely impair performance on a route where an airline becomes the best option after many routes of poor results. For example, if Ascend performs poorly on the first eight routes but becomes the best airline on route nine, KTMU will react very slowly to this environmental change because its posterior mean remains heavily weighted against Ascend. The \(\nu\)-KTMU policy relaxes this restriction by separating the historical mean for each airline from the strength with which that mean is carried into the new route. For compactness, define the historical counts available before route \(m\) as
\begin{equation}
N_{k,<m}^{+}=\sum_{r=1}^{m-1}N_{k,r}^{+},
\qquad
N_{k,<m}^{-}=\sum_{r=1}^{m-1}N_{k,r}^{-}.
\label{web-eq:nu_ktmu_historical_counts}
\end{equation}
For \(m\geq 2\), the smoothed historical mean for airline \(k\) is
\begin{equation}
\hat{\theta}_{k,m}^{\mathrm{past}}
=
\frac{1+N_{k,<m}^{+}}
     {2+N_{k,<m}^{+}+N_{k,<m}^{-}}.
\label{web-eq:nu_ktmu_historical_mean}
\end{equation}
The policy uses this mean to initialize the airline's route-\(m\) Beta prior:
\begin{equation}
\theta_{m,k}^{(\nu)}
\mid \mathcal{H}_{1:m-1}
\sim
\operatorname{Beta}\!\left(
\alpha_{k,m}^{(\nu)},\beta_{k,m}^{(\nu)}
\right),
\qquad
\left(\alpha_{k,m}^{(\nu)},\beta_{k,m}^{(\nu)}\right)
=
\begin{cases}
(1,1), & m=1,\\[2pt]
\left(\nu\hat{\theta}_{k,m}^{\mathrm{past}},
\nu\bigl(1-\hat{\theta}_{k,m}^{\mathrm{past}}\bigr)\right),
& m\geq 2.
\end{cases}
\label{web-eq:nu_ktmu_prior}
\end{equation}
On the first route, before a historical mean is available, the policy therefore retains the uninformative \(\mathrm{Beta}(1,1)\) prior. On every later route, \(\alpha_{k,m}^{(\nu)}+\beta_{k,m}^{(\nu)}=\nu\), so the parameter \(\nu>0\) is the effective prior sample size assigned to earlier-route experience. After observing the current-route counts before flight \(t\), the posterior mean is
\begin{equation}
\hat{\theta}_{k,t,m}^{(\nu)}
=
\frac{
\alpha_{k,m}^{(\nu)}+N_{k,m,t-1}^{+}
}{
\alpha_{k,m}^{(\nu)}+\beta_{k,m}^{(\nu)}
+N_{k,m,t-1}^{+}+N_{k,m,t-1}^{-}
}
=
\frac{
\nu\hat{\theta}_{k,m}^{\mathrm{past}}+N_{k,m,t-1}^{+}
}{
\nu+N_{k,m,t-1}^{+}+N_{k,m,t-1}^{-}
},\qquad m\geq 2.
\label{web-eq:nu_ktmu_posterior_mean}
\end{equation}
Thus, a smaller \(\nu\) allows current-route outcomes to move beliefs more quickly, while a larger \(\nu\) preserves the influence of earlier routes for more flights. The policy remains myopic. It defines
\begin{equation}
\mathcal G_{m,t}^{\nu\text{-}\mathrm{KTMU}}
=
\arg\max_{k\in\{1,\ldots,K\}}
\hat{\theta}_{k,t,m}^{(\nu)}
\label{web-eq:nu_ktmu_greedy_set}
\end{equation}
and applies the same \(\epsilon\)-greedy rule as KTMU:
\begin{equation}
\pi_{\epsilon}^{\nu\text{-}\mathrm{KTMU}}
\!\left(\widetilde{k}\mid\mathcal I_{m,t-1};\nu\right)
=
(1-\epsilon)
\frac{
\mathbf{1}\!\left\{
\widetilde{k}\in\mathcal G_{m,t}^{\nu\text{-}\mathrm{KTMU}}
\right\}
}{
\left|\mathcal G_{m,t}^{\nu\text{-}\mathrm{KTMU}}\right|
}
+\frac{\epsilon}{K}.
\label{web-eq:nu_ktmu_choice_probability}
\end{equation}
Thus, the \(\nu\)-KTMU policy is a controlled-shrinkage generalization of KTMU, but KTMU is not nested by a single fixed value of \(\nu\). In particular, Equation~\eqref{web-eq:ktmu_posterior_mean} is recovered if the effective prior sample size is allowed to vary by airline and route according to
\begin{equation}
\nu_{k,m}^{\mathrm{KTMU}}
=2+N_{k,<m}^{+}+N_{k,<m}^{-}.
\label{web-eq:ktmu_adaptive_nu}
\end{equation}
Fixing \(\nu\) instead prevents the transferred prior from becoming mechanically more concentrated merely because more routes have been completed.

\begin{figure}[htp!]
\centering
\includegraphics[width=.8\textwidth]
{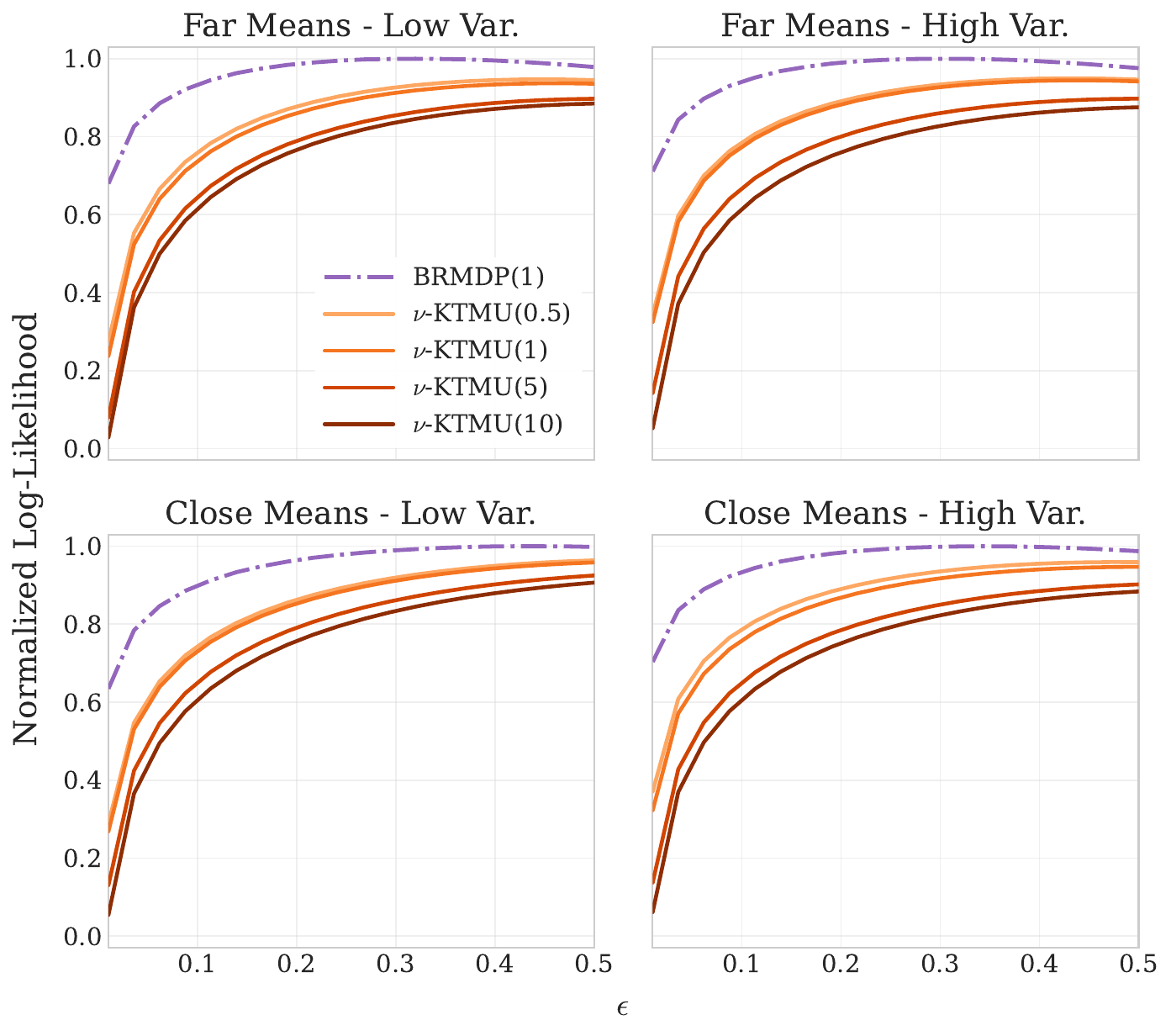}
\caption{Normalized log-likelihood profiles for BRMDP(1) and the myopic shrinkage benchmark. Each panel corresponds to one experimental condition and reports the sample log-likelihood over \(20\) values of \(\epsilon\) for fixed \(\nu\in\{.5,1,5,10\}\). Within each condition, the normalization range is shared across BRMDP(1), both shrinkage-policy families, and the full evaluated grid \(\nu\in\{.5,1,2,5,10,20\}\). Higher values indicate a better fit.}
\label{web-fig:nu_ktmu_epsilon_profiles}
\end{figure}

\paragraph{Likelihood comparison.}
For either shrinkage benchmark \(g\in\{\nu\text{-}\mathrm{KTMU},\nu\text{-}\mathrm{KTDP}\}\), the participant-level log-likelihood is
\begin{equation}
\ell_g(\epsilon,\nu)
=
\sum_{m=1}^{M}\sum_{t=1}^{T}
\log
\pi_{\epsilon}^{g}
\!\left(A_{m,t}\mid\mathcal I_{m,t-1};\nu\right).
\label{web-eq:nu_kt_log_likelihood}
\end{equation}
Thus, \(\epsilon\) governs choice noise, while \(\nu\) governs the strength of the prior transferred from earlier routes. The likelihood is exact because \(\nu\)-KTMU contains no latent route-level draw and requires only sequential posterior updating. Figure~\ref{web-fig:nu_ktmu_epsilon_profiles} compares fixed values \(\nu\in\{.5,1,5,10\}\) with BRMDP(1) over the same \(20\)-point \(\epsilon\) grid. We denote a $\nu$-KTMU policy with $\nu = \dot{\nu}$ as $\nu$-KTMU($\dot{\nu}$). As in our previous results, BRMDP(1) lies above every displayed \(\nu\)-KTMU profile throughout the \(\epsilon\) grid in all four conditions. Allowing the strength of the transferred historical mean to vary therefore does not close the fit gap; within the shrinkage family, the weaker-transfer specifications generally fit better than the stronger-transfer specifications.

\FloatBarrier

\subsubsection{Shrinkage Knowledge Transfer Dynamic Programming (\(\nu\)-KTDP)}
Both KTMU and \(\nu\)-KTMU choose directly from posterior means, so their likelihood disadvantage relative to BRMDP could reflect the absence of forward-looking planning rather than the absence of a hyper-posterior. The \(\nu\)-KTDP policy isolates this distinction. It initializes every route with the same shrinkage prior in Equation~\eqref{web-eq:nu_ktmu_prior}, but uses that prior in the finite-horizon dynamic program introduced in $\S$\ref{ssec:DP_reordered}.

At flight \(t\) on route \(m\), the state is
\begin{equation}
s_{t,m}^{\nu\text{-}\mathrm{KTDP}}
=
\left(
\mathbf N_{m,t-1}^{+},\mathbf N_{m,t-1}^{-};
\boldsymbol{\alpha}_{m}^{(\nu)},\boldsymbol{\beta}_{m}^{(\nu)}
\right),
\label{web-eq:nu_ktdp_state}
\end{equation}
where the prior vectors collect the airline-specific parameters in Equation~\eqref{web-eq:nu_ktmu_prior}. The posterior mean \(\hat{\theta}_{k,t,m}^{(\nu)}\) is given by Equation~\eqref{web-eq:nu_ktmu_posterior_mean}. The action value satisfies
\begin{align}
V_{\epsilon}^{\nu\text{-}\mathrm{KTDP}}
&\!\left(
k\mid
\mathbf N_{m,t-1}^{+},\mathbf N_{m,t-1}^{-};
\boldsymbol{\alpha}_{m}^{(\nu)},\boldsymbol{\beta}_{m}^{(\nu)}
\right)
=
\hat{\theta}_{k,t,m}^{(\nu)}
\nonumber\\
&\quad+
\gamma\Bigl[
\hat{\theta}_{k,t,m}^{(\nu)}
W_{\epsilon}^{\nu\text{-}\mathrm{KTDP}}
\!\left(
\mathbf N_{m,t-1}^{+}+\mathbf e_k,\mathbf N_{m,t-1}^{-};
\boldsymbol{\alpha}_{m}^{(\nu)},\boldsymbol{\beta}_{m}^{(\nu)}
\right)
\nonumber\\
&\qquad\quad+
\bigl(1-\hat{\theta}_{k,t,m}^{(\nu)}\bigr)
W_{\epsilon}^{\nu\text{-}\mathrm{KTDP}}
\!\left(
\mathbf N_{m,t-1}^{+},\mathbf N_{m,t-1}^{-}+\mathbf e_k;
\boldsymbol{\alpha}_{m}^{(\nu)},\boldsymbol{\beta}_{m}^{(\nu)}
\right)
\Bigr].
\label{web-eq:nu_ktdp_action_value}
\end{align}
As in the baseline DP, the corresponding state value averages over future choices induced by the same \(\epsilon\)-greedy rule:
\begin{align}
W_{\epsilon}^{\nu\text{-}\mathrm{KTDP}}
\!\left(s_{t,m}^{\nu\text{-}\mathrm{KTDP}}\right)
&=
(1-\epsilon)\max_k
V_{\epsilon}^{\nu\text{-}\mathrm{KTDP}}
\!\left(k\mid s_{t,m}^{\nu\text{-}\mathrm{KTDP}}\right)
\nonumber\\
&\quad+
\frac{\epsilon}{K}\sum_{k=1}^{K}
V_{\epsilon}^{\nu\text{-}\mathrm{KTDP}}
\!\left(k\mid s_{t,m}^{\nu\text{-}\mathrm{KTDP}}\right).
\label{web-eq:nu_ktdp_state_value}
\end{align}
The choice probability is therefore
\begin{equation}
\pi_{\epsilon}^{\nu\text{-}\mathrm{KTDP}}
\!\left(\widetilde{k}\mid s_{t,m}^{\nu\text{-}\mathrm{KTDP}}\right)
=
(1-\epsilon)
\frac{
\mathbf{1}\!\left\{
\widetilde{k}\in\arg\max_k
V_{\epsilon}^{\nu\text{-}\mathrm{KTDP}}
\!\left(k\mid s_{t,m}^{\nu\text{-}\mathrm{KTDP}}\right)
\right\}
}{
\left|
\arg\max_k
V_{\epsilon}^{\nu\text{-}\mathrm{KTDP}}
\!\left(k\mid s_{t,m}^{\nu\text{-}\mathrm{KTDP}}\right)
\right|
}
+\frac{\epsilon}{K}.
\label{web-eq:nu_ktdp_choice_probability}
\end{equation}
At the final flight, Equation~\eqref{web-eq:nu_ktdp_action_value} reduces to the posterior mean because there is no continuation value. At the start of the next route, the policy recomputes the historical mean and shrinkage prior from the completed-route counts. Thus, \(\nu\)-KTDP transfers airline-specific experience and values information within a route, but it never constructs \(\mathcal P\), updates \(Q_m\), or integrates over higher-order uncertainty.

\begin{figure}[htp!]
\centering
\includegraphics[width=.8\textwidth]
{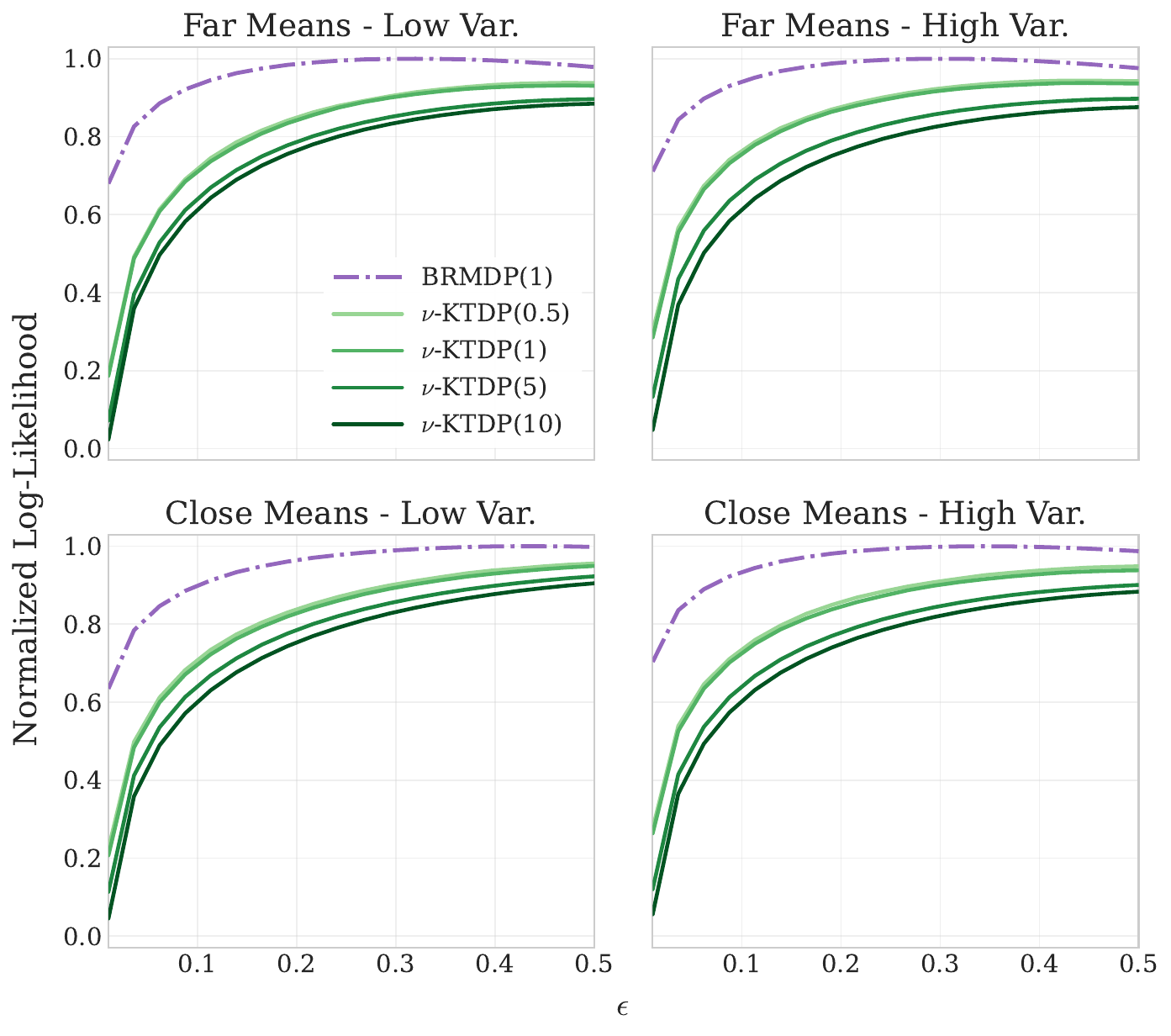}
\caption{Normalized log-likelihood profiles for BRMDP(1) and the forward-looking shrinkage benchmark. Each panel corresponds to one experimental condition and reports the sample log-likelihood over \(20\) values of \(\epsilon\) for fixed \(\nu\in\{.5,1,5,10\}\). We set \(\gamma=1\) and \(h=10\) for both dynamic-programming policies. The normalization follows Figure~\ref{web-fig:nu_ktmu_epsilon_profiles}, so the vertical scale is common across the two figures within each condition. Higher values indicate a better fit.}
\label{web-fig:nu_ktdp_epsilon_profiles}
\end{figure}

\paragraph{Likelihood comparison.}
We evaluate \(\nu\)-KTDP with the same likelihood in Equation~\eqref{web-eq:nu_kt_log_likelihood}. The likelihood remains exact, but the policy additionally requires backward induction for each transferred prior. Figure~\ref{web-fig:nu_ktdp_epsilon_profiles} compares the forward-looking shrinkage specifications with BRMDP(1) over the same \(\epsilon\) and \(\nu\) grid used for \(\nu\)-KTMU. BRMDP(1) again lies above every displayed profile throughout the \(\epsilon\) grid in every condition. Adding a continuation value to the transferred historical mean therefore does not eliminate BRMDP(1)'s likelihood advantage. As with \(\nu\)-KTMU, weaker-transfer specifications generally fit better than stronger-transfer specifications.

\FloatBarrier

Taken together, the three benchmarks narrow the interpretation of the main result. BRMDP(1)'s advantage is not explained by cross-route transfer alone, by calibrating the strength of a transferred historical mean, or by combining that mean with forward-looking planning. The data instead favor the boundedly rational representation of higher-order uncertainty embodied in BRMDP(1) over these lower-dimensional forms of knowledge transfer.

\clearpage
\section{Stylized Managerial Counterfactuals}
\label{web-sec:web_managerial_counterfactuals}

This appendix provides stylized counterfactuals that illustrate the managerial implications discussed in $\S$\ref{sec:managerial_implications}. These examples are not estimated market outcomes from our laboratory task. Instead, they are simple numerical illustrations of how the paper's preferred behavioral model can imply different firm policies than DP or MetaDP when consumer behavior is closer to BRMDP(1).

\subsection{Stylized Counterfactuals for Pricing and Promotion}
\label{web-ssec:web_pricing_promotion_counterfactuals}

Consider a first-purchase decision on a new route or in a new category for a focal brand. Normalize the outside option and marginal cost to zero, and suppose the consumer buys if expected quality exceeds price:
\begin{equation}
u = \hat{\theta} - p,
\qquad
\text{buy if } \hat{\theta}\ge p.
\label{web-eq:web_managerial_buy_rule}
\end{equation}
Interpret market demand as the share of otherwise identical consumers who face the same prior history.

For transparency, suppose the consumer's hyper-posterior is formed on two hypotheses:
\begin{equation}
H_H:\theta \sim \mathrm{Beta}(8,2)
\quad (\mu_H=.8),
\qquad
H_L:\theta \sim \mathrm{Beta}(4,6)
\quad (\mu_L=.4),
\label{web-eq:web_managerial_hypotheses}
\end{equation}
with initial hyper-prior \(Q_1(H_H)=Q_1(H_L)=.5\). Under DP, consumers treat the new route as unrelated and start from the uninformative prior \(\mathrm{Beta}(1,1)\), so expected quality is \(.5\). Under MetaDP, all consumers use the integrated posterior mean \(.8Q_H+.4Q_L\). Under BRMDP(1), a fraction \(Q_H\) of consumers behave as if they drew \(H_H\), while a fraction \(Q_L\) behave as if they drew \(H_L\). Therefore, launch demand is
\begin{align}
q^{DP}(p) &= \mathbf{1}\{p\le .5\}, \\
q^{MetaDP}(p) &= \mathbf{1}\{p\le .8Q_H+.4Q_L\}, \\
q^{BRMDP(1)}(p) &= Q_H\,\mathbf{1}\{p\le .8\} + Q_L\,\mathbf{1}\{p\le .4\}.
\label{web-eq:web_managerial_demands}
\end{align}
The key point is that
\begin{equation}
\mathbb{E}\!\left[\mathbf{1}\{\hat{\theta}\ge p\}\right]
\neq
\mathbf{1}\!\left\{\mathbb{E}[\hat{\theta}] \ge p\right\},
\label{web-eq:web_managerial_nonlinearity}
\end{equation}
so a firm that collapses heterogeneous coarse priors into one representative posterior can choose the wrong price or promotion.

\paragraph{Example 1: after strong favorable experience, DP underprices.}
Suppose the focal brand performed very well on the previous route, generating two on-time flights out of two. This history shifts posterior weight toward the high-quality hypothesis:
\[
Q_H = .783,
\qquad
Q_L = .217.
\]
The firm is now choosing a launch price for the new route. DP ignores cross-route transfer and therefore recommends a conservative introductory price, while MetaDP and BRMDP(1) both imply stronger demand. Under true BRMDP(1) behavior, however, the profit-maximizing price is even higher because the market contains a large high-belief segment rather than a single representative consumer.

\begin{table}[htp!]
\centering
\small
\caption{Pricing after strong favorable experience}
\label{web-tab:web_managerial_pricing_example1}
\begin{tabular}{p{.18\textwidth} p{.21\textwidth} p{.17\textwidth} p{.19\textwidth} p{.17\textwidth}}
\toprule
\textbf{Observed history} & \textbf{Posterior weights} & \textbf{Firm action} & \textbf{True BRMDP(1) outcome} & \textbf{Interpretation} \\
\midrule
Previous route yields \(2/2\) successes
&
\(Q_H=.783,\;Q_L=.217\)
&
DP sets \(\mathit{p} = .50\)

MetaDP sets \(\mathit{p} = .713\)

BRMDP(1)-optimal price is \(\mathit{p} = .80\)
&
At \(\mathit{p} = .50\), true profit is \(.392\)

At \(\mathit{p} = .713\), true profit is \(.558\)

At \(\mathit{p} = .80\), true profit is \(.626\)
&
DP underprices because it ignores transferable learning. Even MetaDP underprices because it smooths the high-belief segment into an average posterior.
\\
\bottomrule
\end{tabular}
\end{table}

To see the BRMDP(1) logic directly, note that demand on the new route is
\[
q^{BRMDP(1)}(p)=
\begin{cases}
1, & p\le .4,\\
.783, & .4 < p \le .8,\\
0, & p>.8.
\end{cases}
\]
Hence true profit, \(\pi(p)=p\,q^{BRMDP(1)}(p)\), is maximized at \(\mathit{p} = .8\). The managerial implication is that a no-transfer model can make the firm too conservative precisely when prior experience should justify more aggressive monetization.

\paragraph{Example 2: after mixed experience, MetaDP rejects a profitable coupon.}
Now suppose the prior route generated mixed evidence: one on-time and one delayed. This history leaves substantial posterior weight on both hypotheses:
\[
Q_H = .4,
\qquad
Q_L = .6.
\]
The firm starts from a list price of \(\mathit{p} = .56\), which is the price suggested by MetaDP because the integrated posterior mean is \(.8(.4)+.4(.6)=.56\). The managerial question is whether to offer a launch coupon of \(.16\), reducing the effective price to \(.40\). Under MetaDP, the coupon appears unnecessary because the representative consumer already buys at \(\mathit{p} = .56\). Under BRMDP(1), by contrast, the coupon is profitable because it brings the low-prior segment into the market.

\begin{table}[htp!]
\centering
\small
\caption{Promotion decision after mixed experience}
\label{web-tab:web_managerial_promotion_example2}
\begin{tabular}{p{.18\textwidth} p{.20\textwidth} p{.18\textwidth} p{.18\textwidth} p{.18\textwidth}}
\toprule
\textbf{Observed history} & \textbf{Posterior weights} & \textbf{MetaDP recommendation} & \textbf{True BRMDP(1) outcome} & \textbf{Interpretation} \\
\midrule
Previous route yields \(1\) on-time and \(1\) delayed
&
\(Q_H=.4,\;Q_L=.6\)
&
Set list price \(\mathit{p} = .56\) and reject the coupon
&
At \(\mathit{p} = .56\), true demand is \(.4\) and revenue is \(.224\)

With a coupon, effective price becomes \(.40\), true demand rises to \(1\), and revenue rises to \(.40\)
&
MetaDP rejects a profitable promotion because it averages away the low-prior segment that BRMDP(1) preserves.
\\
\bottomrule
\end{tabular}
\end{table}

Under BRMDP(1), demand in this case is
\[
q^{BRMDP(1)}(p)=
\begin{cases}
1, & p\le .4,\\
.4, & .4 < p \le .8,\\
0, & p>.8.
\end{cases}
\]
So at the MetaDP list price \(\mathit{p} = .56\), only the high-prior segment buys. A coupon that reduces the effective price to \(.40\) activates the full market. The managerial implication is that full integration can make the firm too aggressive on price and too reluctant to use targeted launch promotions.

These examples are stylized, but the broader implication is general. When consumers transfer beliefs across contexts using coarse sampled priors, managers should expect launch demand to look more like a mixture of segments than a single representative-consumer curve. As a result, the direction of the policy error need not be constant: DP can under-react after strong favorable experience, while MetaDP can over- or under-react depending on the decision threshold. This cautions against using either no-transfer DP or fully integrated MetaDP as the default basis for pricing and promotion counterfactuals.

\subsection{Stylized Counterfactuals for Route-Entry Interventions}
\label{web-ssec:web_route_entry_counterfactuals}

We next illustrate the managerial implications of the model comparison with two stylized counterfactuals that stay close to our experimental task. Both focus on the beginning of a new route, when consumers have little route-specific information but may transfer beliefs from prior routes. Specifically, we consider two route-entry decisions: whether to feature a focal airline as the recommended option on the first flight, and whether to encourage an early trial of that airline through a one-time intervention.

This subsection is explicitly a misspecified-forecasting exercise. Consistent with the likelihood results, we take consumers' true behavior to be BRMDP(1). The manager, however, does not know this and instead chooses route-entry interventions as if consumers followed either DP or MetaDP. The purpose of the examples below is therefore to show how a manager can make the wrong decision when the behavioral model used for forecasting is misspecified.

We focus on the beginning of a new route, which is the part of the task where cross-route priors matter most. Let airline \(A\) denote the focal airline and \(B\) the best competing airline. Let \(z\in\{0,1\}\) denote a route-entry intervention chosen by the manager. When \(z=1\), the intervention adds utility \(b>0\) to choosing airline \(A\) on the first flight of the new route. Depending on the example, this intervention can be interpreted as featured placement or a one-time trial credit. Let \(r_B\) denote the route-entry utility of the best competing airline \(B\). A consumer chooses airline \(A\) on the first flight iff
\begin{equation}
\hat{\theta}_A + zb \ge r_B.
\label{web-eq:web_route_entry_choice_rule}
\end{equation}

If airline \(A\) earns margin \(m\) from a first-flight booking and the intervention costs \(c\) per customer, then the manager's payoff is
\begin{equation}
\Pi(z)=m\,q(z)-cz,
\label{web-eq:web_route_entry_profit}
\end{equation}
where \(q(z)\) is the share of consumers who choose airline \(A\) on the first flight. A manager who assumes model \(M\in\{\text{DP},\text{MetaDP}\}\) chooses
\begin{equation}
z^{M}\in \arg\max_{z\in\{0,1\}} \Pi^{M}(z),
\label{web-eq:web_manager_misspecified_choice}
\end{equation}
whereas the correct decision is
\begin{equation}
z^{BRMDP(1)}\in \arg\max_{z\in\{0,1\}} \Pi^{BRMDP(1)}(z).
\label{web-eq:web_manager_correct_choice}
\end{equation}
For simplicity, we focus on first-flight acquisition value. Adding continuation value from later flights would strengthen the same logic, because early-route choices affect subsequent learning within the route.

To keep the exposition transparent, we again use the two-hypothesis representation of transferred beliefs about airline \(A\):
\begin{equation}
H_H:\theta_A \sim \mathrm{Beta}(8,2)
\quad (\mu_H=.8),
\qquad
H_L:\theta_A \sim \mathrm{Beta}(4,6)
\quad (\mu_L=.4).
\label{web-eq:web_managerial_hypotheses_route}
\end{equation}
Let \(Q_H\) and \(Q_L\) denote the current hyper-posterior weights on \(H_H\) and \(H_L\). Then route-entry demand for airline \(A\) is
\begin{align}
q^{DP}(z) &= \mathbf{1}\{.5+zb \ge r_B\}, \\
q^{MetaDP}(z) &= \mathbf{1}\{(.8Q_H+.4Q_L)+zb \ge r_B\}, \\
q^{BRMDP(1)}(z) &= Q_H\,\mathbf{1}\{.8+zb \ge r_B\}
+ Q_L\,\mathbf{1}\{.4+zb \ge r_B\}.
\label{web-eq:web_route_entry_demands}
\end{align}
The key nonlinearity is that
\begin{equation}
\mathbb{E}\!\left[\mathbf{1}\{\hat{\theta}_A+zb\ge r_B\}\right]
\neq
\mathbf{1}\!\left\{\mathbb{E}[\hat{\theta}_A]+zb\ge r_B\right\}.
\label{web-eq:web_route_entry_nonlinearity}
\end{equation}
This is exactly why a manager using MetaDP can choose differently from a manager who correctly recognizes that consumers behave according to BRMDP(1).

The two examples below isolate different misspecifications. Example 1 shows how a manager who assumes DP can overinvest in route-entry support because DP ignores transferred beliefs that consumers already carry into the new route. Example 2 shows how a manager who assumes MetaDP can underinvest in a trial incentive because MetaDP averages away a responsive high-prior segment.

\paragraph{Example 1: assuming DP leads the airline to overinvest in featured placement.}
Suppose airline \(A\) generated two on-time flights out of two on the previous route. This shifts posterior weight toward the favorable hypothesis:
\[
Q_H=.783,
\qquad
Q_L=.217.
\]
Consider a featured-placement intervention with bonus \(b=.25\), competing-airline utility \(r_B=.60\), first-flight margin \(m=.10\), and per-customer cost \(c=.05\). A manager who assumes DP predicts that without featured placement no consumer chooses airline \(A\), while with featured placement all consumers do. Under true BRMDP(1) behavior, however, a large share of consumers already chooses airline \(A\) even without the intervention because favorable experience from the earlier route carries over.

\begin{table}[htp!]
\centering
\small
\caption{Manager assumes DP, but consumers follow BRMDP(1)}
\label{web-tab:web_managerial_route_example1}
\begin{tabular}{p{.12\textwidth} p{.18\textwidth} p{.18\textwidth} p{.20\textwidth} p{.20\textwidth}}
\toprule
\textbf{Intervention} & \textbf{Predicted share under DP} & \textbf{Actual share under BRMDP(1)} & \textbf{Predicted payoff under DP} & \textbf{Actual payoff under BRMDP(1)} \\
\midrule
\(z=0\): no featured placement
&
\(q^{DP}(0)=0\)
&
\(q^{BRMDP(1)}(0)=.783\)
&
\(.10\times 0 = 0\)
&
\(.10\times .783 = .078\)
\\[1.0ex]
\(z=1\): featured placement
&
\(q^{DP}(1)=1\)
&
\(q^{BRMDP(1)}(1)=1\)
&
\(.10\times 1 - .05 = .05\)
&
\(.10\times 1 - .05 = .05\)
\\
\bottomrule
\end{tabular}
\end{table}

Thus, a manager using DP chooses \(z^{DP}=1\), while the correct decision under BRMDP(1) is \(z^{BRMDP(1)}=0\). The error arises because DP ignores the favorable beliefs consumers already transfer into the new route and therefore overstates the value of paying for route-entry support.

\paragraph{Example 2: assuming MetaDP leads the airline to underinvest in a trial credit.}
Now suppose airline \(A\) generated a mixed prior-route experience: one on-time and one delayed. This yields
\[
Q_H=.4,
\qquad
Q_L=.6.
\]
Consider a one-time trial credit with bonus \(b=.05\), competing-airline utility \(r_B=.82\), first-flight margin \(m=.10\), and per-customer cost \(c=.02\). Under MetaDP, the integrated route-entry belief about airline \(A\) is
\[
.8(.4)+.4(.6)=.56,
\]
so even with the credit the representative consumer still does not choose airline \(A\). A manager using MetaDP therefore rejects the intervention. Under true BRMDP(1) behavior, however, the credit is enough to move the high-prior segment, even though it does not move the low-prior segment.

\begin{table}[htp!]
\centering
\small
\caption{Manager assumes MetaDP, but consumers follow BRMDP(1)}
\label{web-tab:web_managerial_route_example2}
\begin{tabular}{p{.12\textwidth} p{.20\textwidth} p{.20\textwidth} p{.18\textwidth} p{.18\textwidth}}
\toprule
\textbf{Intervention} & \textbf{Predicted share under MetaDP} & \textbf{Actual share under BRMDP(1)} & \textbf{Predicted payoff under MetaDP} & \textbf{Actual payoff under BRMDP(1)} \\
\midrule
\(z=0\): no trial credit
&
\(q^{MetaDP}(0)=0\)
&
\(q^{BRMDP(1)}(0)=0\)
&
\(.10\times 0 = 0\)
&
\(.10\times 0 = 0\)
\\[1.0ex]
\(z=1\): trial credit
&
\(q^{MetaDP}(1)=0\)
&
\(q^{BRMDP(1)}(1)=.4\)
&
\(.10\times 0 - .02 = -.02\)
&
\(.10\times .4 - .02 = .02\)
\\
\bottomrule
\end{tabular}
\end{table}

Hence, a manager using MetaDP chooses \(z^{MetaDP}=0\), while the correct decision under BRMDP(1) is \(z^{BRMDP(1)}=1\). The error arises because MetaDP compresses the market into a single representative consumer and therefore misses a responsive high-prior segment that is profitable to target with a small trial incentive.

Taken together, these examples show that the managerial problem is one of behavioral misspecification. The underlying consumer behavior is BRMDP(1), but a manager who assumes either DP or MetaDP can choose the wrong route-entry intervention. The direction of the mistake depends on the source of misspecification. Assuming DP can make the manager overinvest in route-entry support because it ignores transferred beliefs consumers already have. Assuming MetaDP can make the manager underinvest in targeted interventions because it averages away coarse segment structure. More broadly, when behavior is closer to BRMDP(1), managers should evaluate route-entry policies using a mixture of coarse transferred priors rather than a no-transfer benchmark or a fully integrated representative-consumer benchmark.

\clearpage
\section{Environment Check}
\label{web-sec:environment_check}
We calculate the means and variances of the generated data and report them in Table~\ref{web-tab:generated_vs_theory_3} to verify that the environment produces distributions close to the theoretical targets. The empirical statistics show that participants faced conditions that are close to the theoretical ones we intended.
\begin{table}[htp!]
\centering
\small
{\setlength{\tabcolsep}{4pt}
\renewcommand{\arraystretch}{1.08}
\begin{tabular}{l rr rr rr}
\toprule
\multirow{2}{*}{\shortstack[l]{\textbf{Condition}\\\textbf{(target means; variance)}}}
 & \multicolumn{2}{c}{\textbf{Lower Airline}}
 & \multicolumn{2}{c}{\textbf{Middle Airline}}
 & \multicolumn{2}{c}{\textbf{Higher Airline}} \\
\cmidrule(lr){2-3} \cmidrule(lr){4-5} \cmidrule(lr){6-7}
 & \textbf{Mean} & \textbf{Var.}
 & \textbf{Mean} & \textbf{Var.}
 & \textbf{Mean} & \textbf{Var.} \\
\midrule
\shortstack[l]{Far Means--Low Var. \\ $(.2,.5,.8);\,.02$}
 & .201 & .020
 & .499 & .020
 & .799 & .019 \\
\shortstack[l]{Far Means--High Var. \\ $(.2,.5,.8);\,.04$}
 & .191 & .038
 & .506 & .038
 & .799 & .039 \\
\shortstack[l]{Close Means--Low Var. \\ $(.4,.6,.8);\,.02$}
 & .412 & .019
 & .606 & .019
 & .800 & .021 \\
\shortstack[l]{Close Means--High Var. \\ $(.4,.6,.8);\,.04$}
 & .404 & .041
 & .605 & .039
 & .810 & .035 \\
\bottomrule
\end{tabular}
}
\caption{Comparison of generated and theoretical means and variances across airlines under different parameters.}
\label{web-tab:generated_vs_theory_3}
\end{table}

\end{document}